\documentclass{article}

\usepackage{microtype}
\usepackage{graphicx}
\usepackage{subcaption}
\usepackage{booktabs} % for professional tables

\usepackage{hyperref}

\usepackage[preprint]{icml2026}

\usepackage{amsmath}
\usepackage{amssymb}
\usepackage{mathtools}
\usepackage{amsthm}
\usepackage{caption}
\usepackage{subcaption}

\usepackage{xcolor}
\usepackage{tcolorbox}
\newtcolorbox{takeaway}{
  colback=gray!10,
  colframe=gray!70,
  arc=2pt,
  boxrule=0.8pt,
  left=5pt, right=5pt,
  fonttitle=\bfseries,
  top=2pt,              % ← extra top padding
  bottom=2pt,           % ← extra bottom padding
  before skip=1em,      % ← spacing before the box
  after skip=1em,       % ← spacing after the box
}

\usepackage[capitalize,noabbrev]{cleveref}
\usepackage{xspace}
\usepackage{multirow}

\theoremstyle{plain}

\theoremstyle{definition}

\theoremstyle{remark}

\usepackage[textsize=tiny]{todonotes}

\definecolor{cvprblue}{rgb}{0.21,0.49,0.74}
\usepackage{tikz}
\usetikzlibrary{positioning, arrows.meta}
\usepackage{pgfplots}

\usepackage{multibib}

\usepackage[table,x11names,dvipsnames,table]{xcolor}

\usepackage{array}
\newcolumntype{H}{>{\setbox0=\hbox\bgroup}c<{\egroup}@{}}

\usepackage{quoting}
\quotingsetup{vskip=0pt}

\makeatletter
\DeclareRobustCommand\onedot{\futurelet\@let@token\@onedot}
\def\@onedot{\ifx\@let@token.\else.\null\fi\xspace}

\def\eg{\emph{e.g}\onedot}

\def\etal{\emph{et al}\onedot}
\makeatother

\definecolor{rowblue}{RGB}{230,244,255}
\definecolor{rowgreen}{RGB}{232,246,232}
\definecolor{verylightgray}{RGB}{245, 245, 245}

\makeatletter
\renewcommand\paragraph{\@startsection{paragraph}{4}{0pt}%
{0pt}% space before heading
{-0.8em}% run-in spacing
{\normalfont\normalsize\bfseries}%
}
\makeatother

\def\maketitlesupplementary
   {
   \newpage
   }

\icmltitlerunning{When Positional Embeddings Help and Hurt Spatial Reasoning}

\begin{document}

\twocolumn[
  % \icmltitle{Geometry without Position? On the Geometric Role of \\ Positional Embeddings in Vision Transformers}
  % \icmltitle{Geometry Without Position?\\ The Role of Positional Embeddings in Spatial Correspondence}
  \icmltitle{Geometry without Position?  \\When Positional Embeddings Help and Hurt Spatial Reasoning
  }



  \icmlsetsymbol{equal}{*}

  \begin{icmlauthorlist}
    \icmlauthor{Jian Shi}{yyy}
    \icmlauthor{Michael Birsak}{yyy}
    \icmlauthor{Wenqing Cui}{yyy}
    \icmlauthor{Zhenyu Li}{yyy}
    \icmlauthor{Peter Wonka}{yyy}
  \end{icmlauthorlist}

  \icmlaffiliation{yyy}{King Abdullah University of Science and Technology, Thuwal, Saudi Arabia}

  \icmlcorrespondingauthor{Jian Shi}{jian.shi@kaust.edu.sa}
  \icmlcorrespondingauthor{Peter Wonka}{peter.wonka@kaust.edu.sa}

  \icmlkeywords{Machine Learning, ICML}

  \vskip 0.3in
]



\printAffiliationsAndNotice{}  % no special notice (required even if empty)

\begin{abstract}
This paper revisits the role of positional embeddings (PEs) within vision transformers (ViTs) from a geometric perspective.
We show that PEs are not mere token indices but effectively function as geometric priors that shape the spatial structure of the representation.
We introduce token-level diagnostics that measure how multi-view geometric consistency in ViT representation depends on consitent PEs.
Through extensive experiments on 14 foundation ViT models, we reveal how PEs influence multi-view geometry and spatial reasoning.
Our findings clarify the role of PEs as a causal mechanism that governs spatial structure in ViT representations.
Our code is provided in \url{https://github.com/shijianjian/vit-geometry-probes}
\end{abstract}

% Through feature-level analysis of 14 foundation ViTs, we demonstrate that multi-view geometric coherence depends on the consistency of PEs.
% Strikingly, geometric correspondence heavily degrades with unchanged visual content, solely due to inconsistent PEs. 
    
\section{Introduction} % First paragraph written by Jian

Vision Transformers (ViTs)~\cite{dosovitskiy2020image} have revolutionized visual representation learning yet their ability to reason about geometry remains insufficiently understood.
While the self-attention mechanism enables flexible and long-range dependencies, it is inherently permutation-invariant and lacks any explicit spatial structure.
Consequently, ViTs rely on positional embeddings (PEs) to encode the “where” of image patches to introduce coordinates for each token.

Early analyses have offered mixed insights.
Raghu \etal~\cite{raghu2021vision} reported that standard ViTs (trained with a CLS token) are likely to retain strong local spatial representations within their deep layers, suggesting that geometric cues can emerge even without convolution.
However, later studies such as Probe3D~\cite{el2024probing} revealed that large-scale foundation models still struggle with multi-view correspondence and suffer under large viewpoint changes, indicating a lack of geometric generalization.
Notably, while single-view approaches such as DepthAnything~\cite{yang2024depth} benefit directly from pretrained DINOv2~\cite{oquab2023dinov2} representations, multi-view methods like VGGT~\cite{wang2025vggt} and $\pi^3$~\cite{wang2025pi3} typically retrain DINO-based backbones to achieve cross-view consistency.
This contrast suggests that ViTs might encode geometry in a view-dependent manner, rather than being intrinsically tied to visual content alone.

A natural hypothesis is that such behavior arises from the design of PEs.
Because attention lacks intrinsic spatial bias, ViTs depend on PEs to impose spatial order.
Various forms of PEs have been proposed for both 2D and 3D understanding, including absolute coordinate embeddings~\cite{carion2020end, chu2021conditional, touvron21deit}, relative positional biases~\cite{li2021revisiting}, and rotary encodings for angular consistency~\cite{su2024roformer}.
While PEs are known to stabilize training and improve recognition performance~\cite{carion2020end, chu2021conditional, touvron21deit}, they have also been observed to degrade geometric alignment and correspondence accuracy~\cite{sun2021loftr, li2021revisiting}.
For example, DeiT~\cite{touvron21deit} demonstrated that inappropriate PE schemes tend to overfit to spatial layouts rather than learn transferable geometric features.
LoFTR~\cite{sun2021loftr} observed that adopting a DETR-style~\cite{carion2020end} architecture with PEs at every layer led to a decline in feature matching performance, implying that excessive positional bias can hinder correspondence learning.
Li \textit{et al.}~\cite{li2021revisiting} observed that the choice between absolute and relative PEs can drastically affect convergence and localization accuracy.
Recent works like PE-Fields~\cite{bai2025positional} further reveal that positional fields can even govern volumetric reasoning in generative settings, suggesting that PEs play a deeper role in geometric reasoning than previously assumed.
\begin{table*}
\centering
\scriptsize
\setlength{\tabcolsep}{3.5pt}
\renewcommand{\arraystretch}{1.15}
\setlength{\abovecaptionskip}{0em}
\setlength{\belowcaptionskip}{0em}
\caption{\textbf{Comparison of representative foundation ViT architectures and vision encoders}. We categorize by training paradigm (SUP = supervised, SSL = self-supervised learning, VLM = vision–language modeling, MIM = masked image modeling), dataset scale, and positional encoding (PE) strategy.}
\begin{tabular}{l|l|l|l|l}
\toprule
\textbf{Model} & \textbf{Training Objective} & \textbf{Data} & \textbf{Scale} & \textbf{PE Strategy} \\
\midrule
\textbf{ViT-B/16}~\cite{dosovitskiy2020image} & Classification (SUP) & JFT-300M & 303M & Absolute \\
\textbf{DeiT}~\cite{touvron21deit} & Classification + distillation (SUP, SSL) & ImageNet-1k & 1.28M & Absolute \\
\textbf{DINO}~\cite{zhang2022dino} & Self-distillation (SSL) & ImageNet-1k & 1.28M & Absolute \\
\textbf{DINOv2}~\cite{oquab2023dinov2} & Self-distill. + register tokens (SSL) & LVD (Curated multi-source images) & 142M & Absolute \\
\textbf{DINOv3}~\cite{simeoni2025dinov3} & Self-distillation (SSL) & Web-scale curated images & $\sim$1.7B & Rotary \\
\textbf{BEiT}~\cite{bao2021beit} & Masked image modeling (MIM) & ImageNet-21k & 14M & Relative \\
\textbf{Data2Vec-Vision}~\cite{baevski2022data2vec} & Latent prediction (SSL) & ImageNet-1k & 1.28M & Relative \\
\textbf{CLIP}~\cite{radford2021learning} & Contrastive (VLM) & Web image–text pairs & 400M & Absolute \\
% \textbf{MetaCLIP}~\cite{xu2023demystifying} & Contrastive (VLM) & CommonCrawl (curated pairs) & 0.4B–2.5B & Absolute \\
\textbf{I-JEPA}~\cite{assran2023self} & Context prediction (SSL) & ImageNet-1k & 1.28M & Absolute \\
\textbf{SAM}~\cite{kirillov2023segment} & Segmentation (SUP) & SA-1B & 11M imgs / 1.1B masks & Absolute \\
\textbf{SigLIP2}~\cite{tschannen2025siglip} & Multi-Stage (SUP, SSL, MIM, VLM) & WebLI & $\sim$10B & Absolute \\ % checked
\textbf{MLCD}~\cite{anxiang_2024_mlcd} & Contrastive distillation (SSL) & LAION-400M & $\sim$400M & Rotary \\
\textbf{Swin Transformer}~\cite{liu2021swin} & Classification (SUP) & ImageNet-1k & 1.28M & Relative \\
\textbf{SwinV2}~\cite{liu2022swinv2} & Detection / Segmentation (SUP) & ImageNet-22k + COCO & 22M + 118k & Relative \\
\bottomrule
\end{tabular}
\label{tab:foundation-models}
\end{table*}

Prior studies establish that PEs are essential for spatial awareness, yet their precise role in shaping geometric structure remains unclear. Thus, we ask:
\begin{quoting}
\textit{To what extent do positional embeddings determine the geometric structure of visual representations, and under what conditions do they facilitate or hinder spatial reasoning?}
\end{quoting}
We investigate this through a direct, token-level analysis of 14 foundation ViTs spanning diverse PE strategies (\cref{tab:foundation-models}).
In this work, we define geometric structure as the preservation of relative spatial relations between tokens under changes in viewpoint.
Unlike most works~\cite{el2024probing,chen2025feat2gs} indirectly evaluated PEs through downstream tasks, our approach explicitly decouples visual content from positional structure at the token level.

This work provides a systematic analysis of the geometric role of PEs in ViTs.
Rather than treating PEs as auxiliary coordinates, our key perspective is to view PEs as inducing an \emph{implicit spatial kernel} that governs how tokens interact with visual content as a function of position.
Notably, ViT representation encodes spatial relations relative to the model's own viewpoint defined by PEs.
Through controlled experiments that decouple visual content and PEs, we demonstrate that geometric structure across views depends primarily on the consistency of PEs, rather than on visual content alone.
We further show that correspondence degradation under viewpoint change can be largely reversed by restoring the PE alignment.
These findings reframe PEs as a causal mechanism underlying spatial reasoning in ViTs.

\section{Related Work}

\subsection{Geometry and Locality in ViTs}

A central difficulty in modern vision representation learning is balancing {local geometric fidelity} and {global semantic abstraction}.
Chen \etal\cite{chen2023rethinking} directly investigates the representation trade-off between local and global features and shows that many models need to balance both for dense tasks. 
Empirically, though semantically rich, deeper layers tend to lose local structure and positional distinctiveness~\cite{park2022vision,yuan2021hrformer}.
Conversely, architectures emphasizing locality~\cite{li2021localvit,chen2021regionvit,tu2022maxvit} maintain sharper geometric features but lack semantic robustness under large view or appearance changes.
Self-supervised foundation models illustrate this trade-off in distinct forms, such as DINOv2~\cite{oquab2023dinov2} exhibit weakened geometric correspondence~\cite{el2024probing}, while predictive models like I-JEPA~\cite{assran2023self} further emphasize semantic abstraction at the expense of spatial precision.
At the time of this study, even state-of-the-art foundation models such as DINOv3~\cite{simeoni2025dinov3} explicitly state the persistent local–global dilemma, where improved global consistency comes at the expense of local spatial fidelity.
In this work, we revisit the local–global feature balancing from the perspective of positional encoding, revealing how PEs shape the local–global feature trade-off.

\subsection{The Role of Positional Embedding} % Written by Jian

Transformers are permutation-invariant and thus require explicit spatial priors.
PEs provide such cues by encoding the 2D layout of image patches. Mainstream positional encoding methods include absolute~\cite{carion2020end}, relative~\cite{wu2021rethinking}, and rotary~\cite{su2024roformer}. 
Notably, a series of studies reported that the choice of positional encoding methods can heavily affect the performance on tasks that demand spatial reasoning.
DETR~\cite{carion2020end} first demonstrated that omitting PEs causes a sharp drop in detection accuracy, as the model loses absolute localization cues.
Similarly, CPVT~\cite{chu2021conditional} and DeiT~\cite{touvron21deit} reported that dropping or mismatching PEs reduces classification accuracy and harms robustness to input resolution changes, indicating that absolute spatial priors improve model generalization.
For correspondence-driven tasks, explicit positional cues are even more critical. LoFTR~\cite{sun2021loftr} uses 2D positional encodings in its Local Feature Transformer and notes that positional information, together with global receptive fields, is crucial for obtaining high-quality matches.
STTR~\cite{li2021revisiting} further shows that relative positional information is necessary for resolving ambiguity in textureless regions and for producing stable disparity patterns.
Recent 3D transformer models~\cite{liu2022petr,li2021revisiting} introduce disparity- or depth-aware PEs to capture spatial structure across multiple views.
Overall, PEs remain indispensable for preserving geometric coherence, even though their geometric effects remain underexplored. This work provides a token-level analysis of the impact of PEs.

% \subsection{Probing and Interpreting Visual Representations}

% Token-level probing and attention analysis:
% Press et al. (2021) “Train Short, Test Long,” Li et al. (2021) “Revisiting Attention.”

% Representation probing:
% Probe3D~\cite{el2024probing}, Feature2GS~\cite{chen2025feat2gs},
% Hila Chefer et al. “Transformer Interpretability Beyond Attention.”

% Geometric and structure probing:
% works exploring patch topology or distance embedding (e.g., Sinha \& Zisserman 2022 “Depth Probes for ViTs”).

% Possibly mention DINO and I-JEPA analytical papers (Assran et al. 2023; Oquab et al. 2023) to tie in self-supervised perspectives.

% Establish that most interpretability work focuses on attention maps or semantics — whereas your approach probes the spatial geometry encoded in the features themselves.

\section{PEs as Spatial Kernels}

Transformers process a sequence of tokens without any built-in notion of spatial order.
For image inputs, this means that tokens corresponding to local patches are treated as unordered elements unless additional positional cues are provided.
To encode spatial structure, ViTs introduce PEs that supply information about the location of each token.
Let $x_i \in \mathbb{R}^d$ denote the feature of the $i$-th token.
The query and key projections are given by:
\[
q_i = W_Q x_i, \qquad k_j = W_K x_j,
\]
and the raw attention logits (before softmax) are:
\begin{equation}
\label{eq:att_logit}
\alpha_{ij} = \frac{1}{\sqrt{d}} q_i^\top k_j.
\end{equation}
Note that this representation has not added any positional signal yet.
Depending on how the positional signal is formulated, three main variants are commonly used: \emph{absolute}, \emph{relative}, and \emph{rotary} encodings.

\subsection{Background: Absolute, Relative, and Rotary PEs}

\paragraph{Absolute positional encodings.}
Each token receives a unique embedding that encodes its absolute position $p_i=f(i)$ on the image grid, such that:
\begin{equation}
q_i = W_Q (x_i+p_i), \qquad k_j = W_K (x_j+p_j),
\end{equation}
Common formulations include sinusoidal encodings~\cite{vaswani2017attention,dosovitskiy2020image} and learned embeddings~\cite{touvron21deit}.
By encoding absolute coordinates, absolute PEs are not translation-invariant, leading to position-dependent attention patterns.

\paragraph{Relative positional encodings.}
Relative encodings~\cite{shaw2018self,he2020deberta,press2021train} do not modify tokens. Position is directly injected in the logits with a function $b(\Delta_{ij})$ of the displacement $\Delta_{ij}=j-i$:
\begin{equation}
\alpha_{ij} = \tfrac{1}{\sqrt{d}} q_i^\top k_j + b(\Delta_{ij}),
\label{eq:rela_encoding}
\end{equation}
that depends only on relative offset rather than absolute location.
This enhances translation invariance and stability across resolutions.

\paragraph{Rotary positional encodings (RoPE).}
Rotary PEs~\cite{su2024roformer} inject position by rotating queries and keys in a complex-valued space. Given the projected queries and keys $q_i$ and $k_j$, RoPE assigns each position $i$ an angle
$\theta_i$ derived from fixed sinusoidal functions, and applies
\[
q_i \leftarrow \mathcal{R}_{\theta_i}({q}_i), \qquad k_j \leftarrow \mathcal{R}_{\theta_j}({k}_j),
\]
Since rotations compose additively, the attention logit becomes:
This results in dot-products of the form:
\begin{equation}
q_i^\top k_j = {q}_i^\top \mathcal{R}_{\theta_j - \theta_i}({k}_j),
\label{eq:rope}
\end{equation}
which depends only on the relative phase $\theta_j - \theta_i$, effectively encoding angular or offset-based relationships in a multiplicative manner.

\subsection{A Kernel Interpretation of PEs}
\label{supp:pe_kernel_derivation}

We provide an interpretive decomposition of the attention mechanism that clarifies how PEs influence spatial interactions between tokens.
We emphasize that this analysis does not introduce a new attention formulation, nor does it claim a formal equivalence between attention and kernel methods.
Instead, it offers a conceptual lens for understanding how different positional encoding schemes impose structured spatial priors on token interactions.

\paragraph{Absolute \& Relative Positional Encoding (RPE).}
We begin from the standard attention logit formulation introduced in Eq.~\eqref{eq:att_logit}.
% Referring to~\cite{raffel2020exploring}, in relative encodings, the attention bias depends on the displacement $\Delta_{ij}=i-j$. Combining with~\cref{eq:rela_encoding}, under the zero-mean assumption
% we obtain:
% \begin{equation}
% \mathbb{E}[\alpha_{ij}] \;\propto\; b(\Delta_{ij}),
% \label{eq:rpe_kernel}
% \end{equation}
% given the common parameterization of:
% \begin{equation}
% q_i = W_Q x_i,\quad k_j = W_K x_j.
% \end{equation}
% \Cref{eq:rpe_kernel} reveals that RPE induces a {translation-stationary kernel} that depends solely on token displacement. The kernel amplitude typically decays with increasing $|\Delta_{ij}|$, thereby biasing attention toward spatially neighboring tokens.
\label{sec:pe_kernel}
Following~\cite{press2021train,li2021revisiting}, we expand the dot-product attention with additive positional embeddings into four components:
\begin{align}
\alpha_{ij}
&= \underbrace{\tfrac{1}{\sqrt{d}} x_i^\top W_Q^\top W_K x_j}_{\text{content--content}}
+ \underbrace{\tfrac{1}{\sqrt{d}} x_i^\top W_Q^\top W_K p_j}_{\text{content $\to$ position}} \nonumber\\
&\quad + \underbrace{\tfrac{1}{\sqrt{d}} p_i^\top W_Q^\top W_K x_j}_{\text{position $\to$ content}}
+ \underbrace{\tfrac{1}{\sqrt{d}} p_i^\top W_Q^\top W_K p_j}_{\text{position--position}}.
\label{eq:att_decomp}
\end{align}
This decomposition separates content-driven similarity from positional contributions.
The first term encodes content-based similarity (semantic affinity between tokens), while the remaining three terms inject spatial bias through positional embeddings.
The final term depends solely on the positional embeddings and the learned projection matrices, and therefore captures a purely positional interaction.

To make this effect explicit, we consider the expected contribution of the attention logits under mild and commonly used assumptions.
Specifically, in pretrained ViTs, features are normalized by pre-LayerNorm and exhibit approximately zero mean ($\mathbb{E}[x]=0$) across tokens and images.
Under this setting, cross terms involving both content and position are typically smaller in expectation than the purely positional interaction.
While these assumptions are not exact and need not hold for every token or layer, they provide a useful approximation for analyzing the dominant spatial bias induced by positional encodings.
Under this approximation, the expected attention logit can be written as
\begin{equation}
\label{eq:spatial_kernel}
\mathbb{E}[\alpha_{ij}] \;\propto\; p_i^\top M p_j + g(\Delta_{ij}),
\end{equation}
where $M=W_Q^\top W_K$ represents a learned bilinear form and $g$ is any explicit relative bias.
Equation~\eqref{eq:spatial_kernel} reveals that positional encodings induce an implicit spatial interaction pattern between tokens.

\paragraph{Rotary positional encodings (RoPE).}
Unlike additive positional embeddings, RoPE does not introduce explicit positional vectors. Instead, it modulates the interaction between queries and keys through multiplicative phase shifts.
For each each query or key vector $q_i,k_i\in \mathbb{R}^d$ (\eg $d=768$), RoPE applies independent 2D rotations to each pair of adjacent feature dimensions:
\begin{equation}
    (q_{2b-1},q_{2b})\quad \text{for}\;b=1,\dots,\frac{d}{2}.
\end{equation}
Referring to~\cref{eq:rope}, for one 2D feature pair $(q_{2b-1},q_{2b})$ and $(k_{2b-1},k_{2b})$, we have:
\begin{equation}
    \alpha_{ij}^{b}=[q_{2b-1}\;\;q_{2b}]\mathcal{R}_{(i-j)\theta_b} \begin{bmatrix}
    k_{2b-1} \\
    k_{2b}
    \end{bmatrix}.
\end{equation}
Expanding to:
\begin{align}
    \label{eq:att_rot_decomp}
    \alpha_{ij}^{b}=&(q_{2b-1}k_{2b-1}+q_{2b}k_{2b})\text{cos}(\Delta \phi_{ij}^b) \nonumber \\
    & + (q_{2b}k_{2b-1}-q_{2b-1}k_{2b})\text{sin}(\Delta \phi_{ij}^b),
\end{align}
where we use $\Delta \phi_{ij}^b = (j-i)\theta_b$ to denote the difference in rotation angles.
Under the zero-mean expectation, we assume both the $2b-1$-th and $2b$-th components of $q$ and $k$ share the same expected correlation value $\lambda_b$.
Therefore, by assuming the expected covariance between corresponding query-key components is the same $\mathbb{E}[q_{2b-1}k_{2b-1}]=\mathbb{E}[q_{2b}k_{2b}]=\lambda_b$ and others$=0$, we get:
\begin{equation}
    \label{eq:pe_kernel_rot}
    \mathbb{E}[\alpha_{ij}^b]=\lambda_b \cos(\Delta \phi_{ij}^b).
\end{equation}
Hence, the expected dot product between a rotated query-key pair 
depends solely on the relative phase $(\theta_i - \theta_j)$, 
demonstrating that RoPE encodes a cosine-shaped positional kernel.
This expression indicates that, in expectation, the interaction strength between tokens depends on a cosine function of their relative positional offset.

\paragraph{The PE-induced Kernel.}
We refer to the mentioned interactions as \emph{implicit spatial kernels}, not in the sense of fixed or explicitly defined kernel functions, but as the learned positional prior that governs how strongly tokens at different spatial locations attend to one another.
This kernel interpretation is not meant as an exact equivalence, but as a diagnostic abstraction.
In absolute encodings, this kernel takes the form of a non-stationary bilinear Gram matrix $p_i^\top M p_j$, while in relative and rotary encodings, it becomes a stationary or phase-stationary kernel depending on relative displacement.
It raises the possibility that ViT representations can be factorized into content- and position-dependent components.

Under this interpretation, PEs do not merely annotate token locations. Instead, they induce a learned spatial interaction prior that structures the representation space itself.
The resulting representation is then defined with respect to this PE kernel. Kernel visualizations are provided in supplementary.

\begin{table}[b]
    \centering
    \scriptsize
    \rowcolors{1}{}{verylightgray}
    \setlength{\abovecaptionskip}{0em}
    \setlength{\belowcaptionskip}{-1em}
    \caption{\textbf{Spatial anchoring fails despite identical content under overlapping crops.} Cross-view token similarity is evaluated within the overlapping region of two crops that share identical pixel content. Increasing crop offset induces positional misalignment and leads to a pronounced drop in correspondence.}
    \begin{tabular}{l|c|ccc  HHH}
        \toprule
         % & & \multicolumn{3}{c}{Vanilla PE}  \\
         % \cmidrule(lr){3-5}
         & PE & $\Delta x=1$ & $\Delta x=2$ & $\Delta x=3$ \\
        \midrule
        \textbf{DINO}     & Abs. &   0.9782 & 0.9765 & 0.9627  & 0.9980 & 0.9809 & 0.9766   \\
        \textbf{DINOv2}   & Abs. &   0.6775 & 0.5760 & 0.5272  & 0.7446 & 0.6318 & 0.5782  \\
        \textbf{DINOv3}   & Rot. &   0.9261 & 0.7377 & 0.6564  & 0.9744 & 0.9657 & 0.9616  \\
        \textbf{MLCD}     & Rot. &   0.8209 & 0.7345 & 0.6786  & 0.8412 & 0.7660 & 0.7175 \\
        \textbf{I-JEPA}    & Abs. &   0.7318 & 0.7283 & 0.7280  & 0.8646 & 0.8576 & 0.8568  \\
        % \textbf{MetaCLIP} & Abs. &   0.7836 & 0.7589 & 0.7496  & 0.9799 & 0.9281 & 0.9109   \\
        \textbf{DeiT}     & Abs. &   0.8184 & 0.8003 & 0.7947  & 0.9400 & 0.9107 & 0.9031  \\
        \textbf{SigLIP2}  & Abs. &   0.6468 & 0.6920 & 0.6872  & 0.7143 & 0.6312 & 0.6199   \\ 
        \textbf{BEiT}     & Rel. &   0.9998 & 0.9997 & 0.9997  & 0.9999 & 0.9999 & 0.9999  \\
        \textbf{Data2Vec} & Rel. &   0.9850 & 0.9678 & 0.9469  & 0.9978 & 0.9948 & 0.9997 \\
        \textbf{ViT}      & Abs. &   0.7913 & 0.7737 & 0.7416  & 0.9498 & 0.9354 & 0.9338   \\
        \textbf{CLIP}     & Abs. &   0.6758 & 0.6665 & 0.6711  & 0.9234 & 0.8844 & 0.8652   \\
        \textbf{SAM}      & Abs. &   0.9969 & 0.9945 & 0.9921  & 0.9977 & 0.9971 & 0.9971    \\
        % SegFormer& N/A. &   0.9723 & 0.9529 & 0.9315  & -      & -      & - \\
        \textbf{Swin}     & Rel. &   0.6928 & 0.6928 & 0.5899  & 0.8448 & 0.8293 & 0.8189 \\
        \bottomrule
    \end{tabular}
    \label{tab:positional_probe_overlap}
\end{table}

\begin{table*}
\centering
\scriptsize
\setlength{\tabcolsep}{5pt}
\rowcolors{5}{}{verylightgray}
\setlength{\abovecaptionskip}{0em}
\setlength{\belowcaptionskip}{-1em}
\caption{
\textbf{Stereo probing results at 448× resolution.}
Removing and shuffling positional embeddings (\textbf{Zeroed Out PE} and \textbf{Shuffled PE}) disrupts spatial correspondence, leading to larger disparity errors (EPE↑) and weaker spatial anchoring.
Though visually uninterpretable (see supplementary), \textbf{pairwise shuffled PE} generally preserve correspondence for absolute PEs.
Conversely, models with intact positional encodings (\textbf{Vanilla PE}) retain sharper geometric priors, achieving lower EPE and higher recall (R@k).
}
\begin{tabular}{l|HcHcHcH|cHcHcH|ccc|ccc}
\toprule
\multirow{2}{*}{Model} & LI &
\multicolumn{6}{c|}{\textbf{Vanilla PE}} &
\multicolumn{6}{c|}{\textbf{Zeroed Out PE}} &
\multicolumn{3}{c|}{\textbf{Pairwise Shuffled PE}} &
\multicolumn{3}{c}{\textbf{Shuffled PE}} \\
\cmidrule(lr){3-8} \cmidrule(lr){9-14}
\cmidrule(lr){15-17} \cmidrule(lr){18-20} &
& EPE↓ & D1↓ & R@1↑ & R@2↑ & R@5↑ & EC-SIM↑
& EPE↓ & D1↓ & R@1↑ & R@2↑ & R@5↑ & EC-SIM↑
& EPE↓ & R@1↑ & R@5↑
& EPE↓ & R@1↑ & R@5↑\\
\midrule
DINO                    & 0.731 & 11.40 & 0.843 & 0.146 & 0.320 & 0.560 & 0.989 & 24.21 & 0.861 & 0.126 & 0.265 & 0.477 & 0.846
& 8.35  & 0.146 & 0.564 & 63.86  & 0.032 & 0.129 \\
DINOv2                  & 0.541 & 12.64 & 0.810 & 0.131 & 0.296 & 0.515 & 0.890 & 38.08 & 0.908 & 0.087 & 0.179 & 0.315 & 0.69  
& 9.93  & 0.143 & 0.551 & 65.29  & 0.028 & 0.114\\
DINOv3                  & 0.635 & 11.66 & 0.828 & 0.129 & 0.267 & 0.474 & 0.803 & 51.89 & 0.934 & 0.080 & 0.166 & 0.293 & 0.605 
& 60.36  & 0.034 & 0.133 & 60.51  & 0.034 & 0.133 \\
I-JEPA                  & 0.844 &  9.90 & 0.787 & 0.151 & 0.322 & 0.561 & 0.940 & 44.97 & 0.927 & 0.088 & 0.183 & 0.321 & 0.782
& 8.69  & 0.148 & 0.562 & 67.37  & 0.028 & 0.112\\
% MetaCLIP                & 0.306 & 31.61 & 0.923 & 0.140 & 0.311 & 0.534 & 0.910 & 23.74 & 0.824 & 0.131 & 0.270 & 0.470 & 0.665 \\
DEiT                    & 0.304 & 14.69 & 0.809 & 0.152 & 0.320 & 0.553 & 0.908 & 26.22 & 0.861 & 0.122 & 0.254 & 0.436 & 0.640
& 10.42  & 0.152 & 0.562 & 66.87  & 0.027 & 0.111\\
SigLIP2                 & 0.653 & 11.04 & 0.656 & 0.141 & 0.299 & 0.519 & 0.760 & 38.99 & 0.909 & 0.088 & 0.183 & 0.312 & 0.678
& 8.35  & 0.152 & 0.562 & 65.71  & 0.029 & 0.116\\
BeiT                    & 0.118 & 28.98 & 0.901 & 0.143 & 0.298 & 0.531 & 0.939 & 46.31 & 0.929 & 0.115 & 0.243 & 0.441 & 0.872
& 42.72  & 0.136 & 0.509 & 60.57  & 0.054 & 0.210\\
Data2Vec                & 0.263 & 17.55 & 0.919 & 0.153 & 0.324 & 0.564 & 0.986 & 39.64 & 0.907 & 0.112 & 0.239 & 0.424 & 0.786
& 29.02  & 0.143 & 0.525 & 62.87  & 0.052 & 0.194\\
MLCD                    & 0.564 & 23.25 & 0.811 & 0.112 & 0.237 & 0.416 & 0.745 & 38.36 & 0.859 & 0.086 & 0.182 & 0.314 & 0.566
& 59.36  & 0.033 & 0.133 & 59.39  & 0.033 & 0.134 \\
SAM                     & 0.589 & 19.67 & 0.877 & 0.153 & 0.330 & 0.569 & 0.961 & 47.07 & 0.894 & 0.156 & 0.332 & 0.574 & 0.964
& 9.56 & 0.152 & 0.534 & 58.10  & 0.026 & 0.107\\
CLIP                    & 0.445 & 21.39 & 0.793 & 0.148 & 0.310 & 0.531 & 0.847 & 53.80 & 0.941 & 0.081 & 0.166 & 0.290 & 0.541
& 19.24  & 0.145 & 0.535 & 66.75  & 0.026 & 0.106\\
Swin                    & 0.488 & 30.87 & 0.900 & 0.087 & 0.182 & 0.331 & 0.734 & 34.89 & 0.906 & 0.088 & 0.182 & 0.323 & 0.661
& 40.18  & 0.077 & 0.279 & 52.01  & 0.056 & 0.213\\
SwinV2                  & & 27.53 & 0.903 & 0.095 & 0.198 & 0.360 & 0.811 & 33.99 & 0.917 & 0.088 & 0.186 & 0.329 & 0.792 
& 11.74  & 0.145 & 0.554 & 44.72  & 0.065 & 0.240 \\
\bottomrule
\end{tabular}
\label{tab:stereo_upsample_448}
\end{table*}

\section{Content vs. Position: When Geometry Breaks in ViTs}

Spatially consistent feature representations are often assumed to arise from visual content similarities. However, as formalized in~\cref{eq:att_decomp,eq:att_rot_decomp}, token representations inherit a PE-induced kernel, rather than being purely content-driven. This coupling is important for multi-view geometry. Prior work shows that PEs can degrade cross-view alignment~\cite{sun2021loftr,el2024probing}, but the underlying reason is still under-explored.
Thus, we design multi-view correspondence experiments to isolate PE's effects. Notably, multi-view correspondence serves as a stress test for spatial consistency, rather than as the primary task of interest.

\subsection{ViT Representations Fail Under Identical Content}
\label{sec:overlap}

We consider overlapping image crops as a minimal correspondence setting, which preserves image content while selectively perturbs PEs.

\paragraph{Setup.} Given an image $\mathbf{I}$, we generate two overlapping crops, $\mathbf{I}_1$ and $\mathbf{I}_2$, such that the overlapping region contains identical pixels but with misaligned PE kernels.
Since the overlapping regions contain identical content, any similarity degradation cannot be attributed to semantic variation.

We use the test split from Imagenette~\cite{Howard_Imagenette_2019} and measure the cosine similarity of cross-view tokens on the overlapping regions.
We repeat this analysis for increasing horizontal displacements $\Delta x \in \{1,2,3\}$, which induce progressively larger positional reference mismatch.

\paragraph{Results.} As shown in~\Cref{tab:positional_probe_overlap}, we observe a pronounced degradation in correspondence as the horizontal offset between crops increases, despite identical visual input. This occurs consistently across models with different PEs, suggesting that it is not an artifact of a particular architecture or positional encoding scheme. Given identical visual content in the overlapping regions, any degradation in cross-view spatial anchoring must arise from mismatched PEs rather than insufficient visual representation.

\subsection{Positional Mismatch in Stereo Correspondence}
\label{subsec:remove_pe_locality}

Overlapping crops provide a token-level probe of PE mismatch.
In addition, previous studies~\cite{oquab2023dinov2,amir2021deep} show that ViT tokens preserve spatial locality and positional structure also at patch-level granularity. Yet, sub-patch correspondence behavior across views has not been systematically examined.
We therefore design geometric correspondence experiments using a stereo dataset whose disparities predominantly lie within the size of a single token. This setup provides a direct and interpretable measure of whether fine positional anchoring at the sub-patch scale is preserved across views.
When anchoring is preserved, correspondence remains localized, resulting in sub-patch end-point errors (EPE). Conversely, when anchoring is weakened or mismatched, correspondence peaks drift away, leading to larger EPE.

\paragraph{Setup}
Given left and right tokens $\mathbf{F}_L, \mathbf{F}_R$ extracted from a rectified stereo pair, we compute a 4D correlation volume to measure pairwise feature similarity across views.
The construction of the 4D correlation volume is conceptually related to the correlation layers employed in optical flow estimation~\cite{dosovitskiy2015flownet,sun2018pwc,teed2020raft}.
For each token location $(i,j)$ in $\mathbf{F}_L$, we evaluate its correspondence along the epipolar line in the right view with a \textit{soft-argmax} operation.
We evaluate on four PE settings, including 1) vanilla PE, 2) zero out PE, 3) shuffled PE, and 4) pairwise shuffled PE.
Zeroed-out PE removes positional information entirely. Shuffled variants preserve positional information but assign it inconsistently, whereas pairwise shuffled maintains the same PE permutations across views.
Technical details for 4D correlation volume and PE probing variants are provided in the supplementary.

\begin{figure*}[t]
    \centering
    \scriptsize
    \setlength{\tabcolsep}{0pt}
    
    \renewcommand{\arraystretch}{0.25}
    \begin{tabular}{@{}cc cc cc cc@{}}

        \multicolumn{3}{c}{\hspace{1em} \textbf{ Input Left Image \hspace{3em} Input Right Image}}
        & \multicolumn{3}{l}{\hspace{1em} \textbf{SAM Feature Visualization (Left only)}}
        & \multicolumn{2}{c}{\hspace{1em} \textbf{Epipolar Peak Response}} \\
        \cmidrule(lr){1-3} \cmidrule(lr){4-6} \cmidrule(lr){7-8}
        \multicolumn{3}{c}{
            \setlength{\tabcolsep}{1pt} 
            \begin{tabular}{@{}cc@{}}
                 \includegraphics[width=0.166\linewidth]{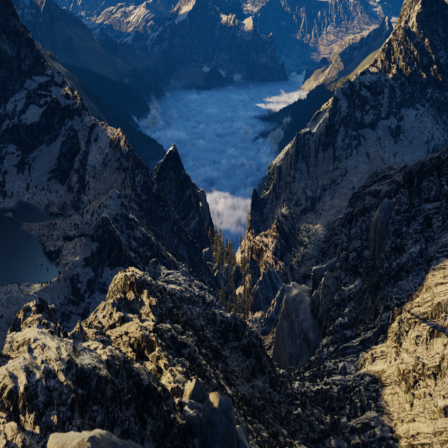}
                 &
                 \includegraphics[width=0.166\linewidth]{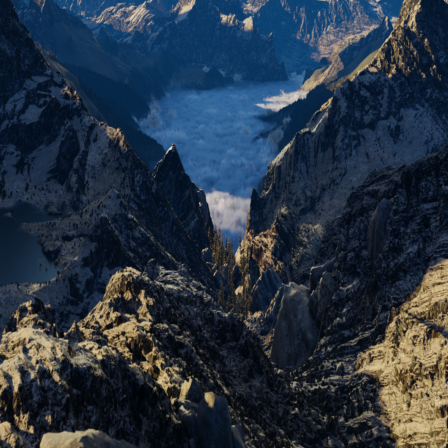} \\
            \end{tabular}
        }
        &
        \multicolumn{3}{c!{\vrule width 1pt}}{
            \setlength{\tabcolsep}{0pt} 
            \begin{tabular}{@{}cc@{}}
                {w/ PE} & {w/o PE} \\
                 \includegraphics[width=0.167\linewidth,trim={12.6cm 0 12.6cm 1.2cm},clip]{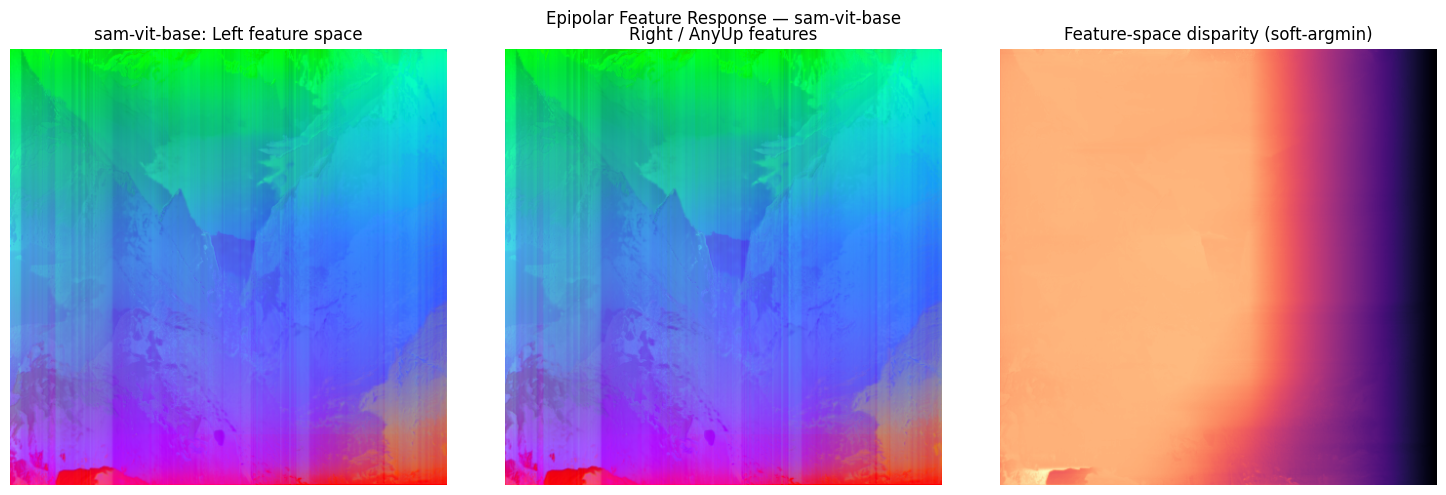}
                 &
                 \includegraphics[width=0.167\linewidth,trim={12.6cm 0 12.6cm 1.2cm},clip]{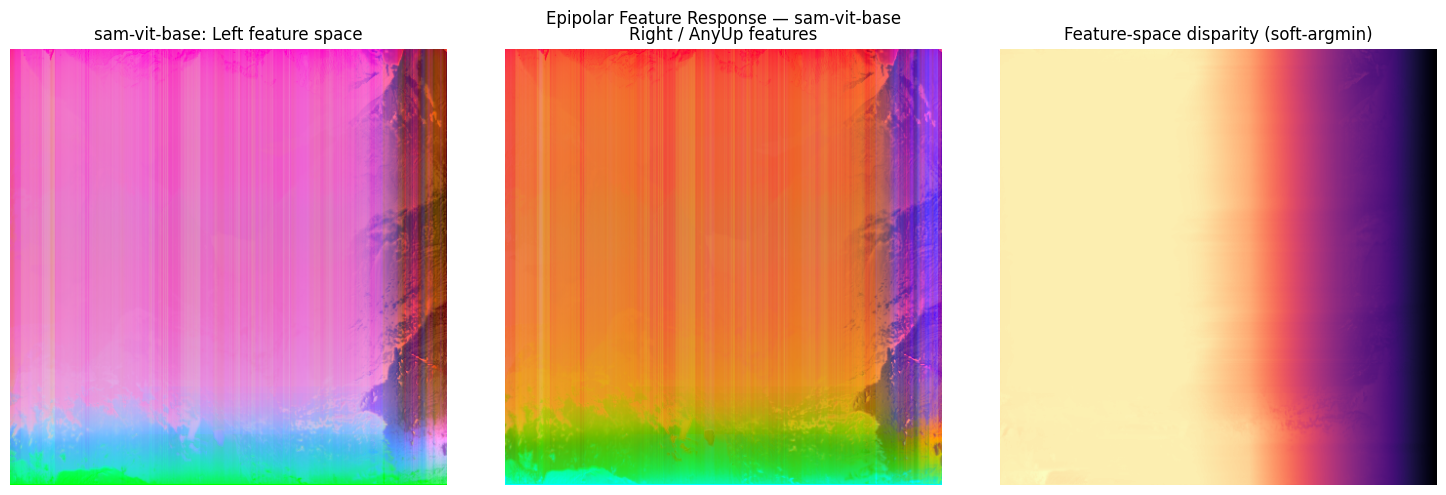} \\
            \end{tabular}
        }
        &
        \multicolumn{2}{c}{
            \setlength{\tabcolsep}{0pt} 
            \begin{tabular}{cc}
                \multicolumn{2}{l}{\hspace{1em} \textbf{SAM}}  \\
                \cmidrule(lr){1-1}
                {w/ PE} & {w/o PE} \\
                 \includegraphics[width=0.12\linewidth,trim={25.2cm 0 0 1.2cm},clip]{misc/disp/epipolar_response_pos_anyup_pca_sam-vit-base.png}
                 &
                 \includegraphics[width=0.12\linewidth,trim={25.2cm 0 0 1.2cm},clip]{misc/disp/epipolar_response_no_pos_anyup_pca_sam-vit-base.png} \\
            \end{tabular}
        }

         % \multicolumn{2}{c}{\textbf{SAM} w/ PE} & \multicolumn{2}{c}{\textbf{SAM} w/o PE}\\
         % \includegraphics[width=0.12\linewidth,trim={12.6cm 0 12.6cm 1.2cm},clip]{misc/disp/epipolar_response_pos_anyup_pca_sam-vit-base.png} &
         % \includegraphics[width=0.12\linewidth,trim={25.2cm 0 0 1.2cm},clip]{misc/disp/epipolar_response_pos_anyup_pca_sam-vit-base.png} &
         % \includegraphics[width=0.12\linewidth,trim={12.6cm 0 12.6cm 1.2cm},clip]{misc/disp/epipolar_response_no_pos_anyup_pca_sam-vit-base.png} &
         % \includegraphics[width=0.12\linewidth,trim={25.2cm 0 0 1.2cm},clip]{misc/disp/epipolar_response_no_pos_anyup_pca_sam-vit-base.png}
         \\
         \cmidrule[1pt](lr){1-6}

        \multicolumn{2}{l}{\hspace{1em} \textbf{DINO}}
        & \multicolumn{2}{l}{\hspace{1em} \textbf{DINOv2}} 
        & \multicolumn{2}{l}{\hspace{1em} \textbf{DINOv3}} 
        & \multicolumn{2}{l}{\hspace{1em} \textbf{I-JEPA}}  \\
        \cmidrule(lr){1-1} \cmidrule(lr){3-3} \cmidrule(lr){5-5} \cmidrule(lr){7-7}

         {\scriptsize w/ PE} & {\scriptsize w/o PE}
         & {\scriptsize w/ PE} & {\scriptsize w/o PE}
         & {\scriptsize w/ PE} & {\scriptsize w/o PE} 
         & {\scriptsize w/ PE} & {\scriptsize w/o PE} \\

         % {\textbf{DINO} w/ PE} & {\textbf{DINO} w/o PE}
         % & {\textbf{DINOv2} w/ PE} & {\textbf{DINOv2} w/o PE}
         % & {\textbf{DINOv3} w/ PE} & {\textbf{DINOv3} w/o PE} 
         % & {\textbf{I-JEPA} w/o PE} & {\textbf{I-JEPA} w/o PE} \\
         
         \includegraphics[width=0.12\linewidth,trim={25.2cm 0 0 1.2cm},clip]{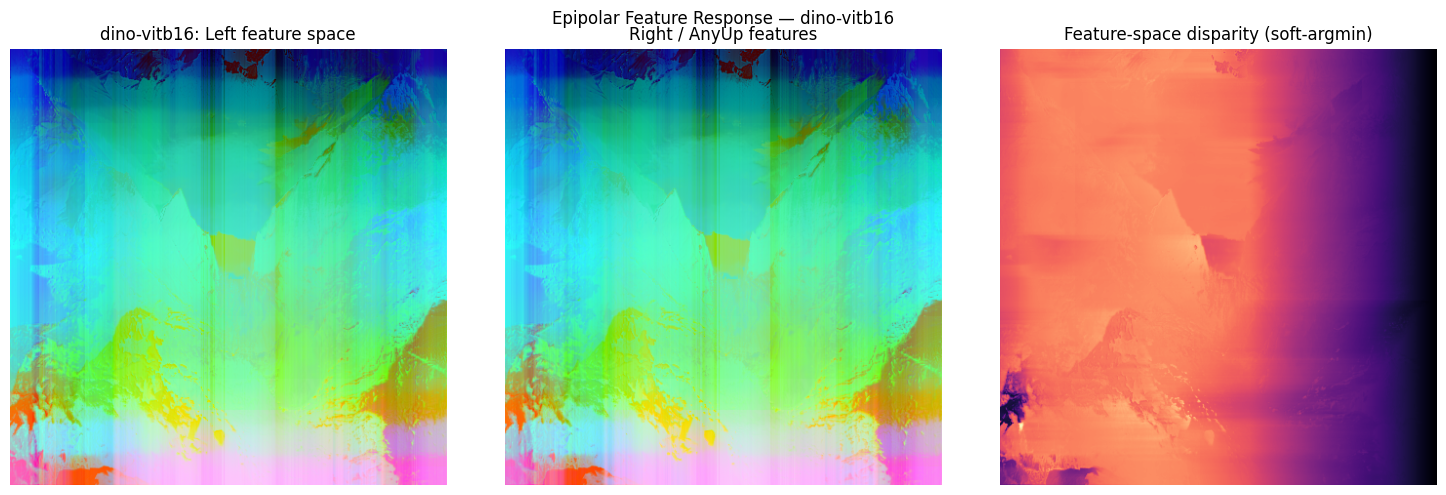} &
         \includegraphics[width=0.12\linewidth,trim={25.2cm 0 0 1.2cm},clip]{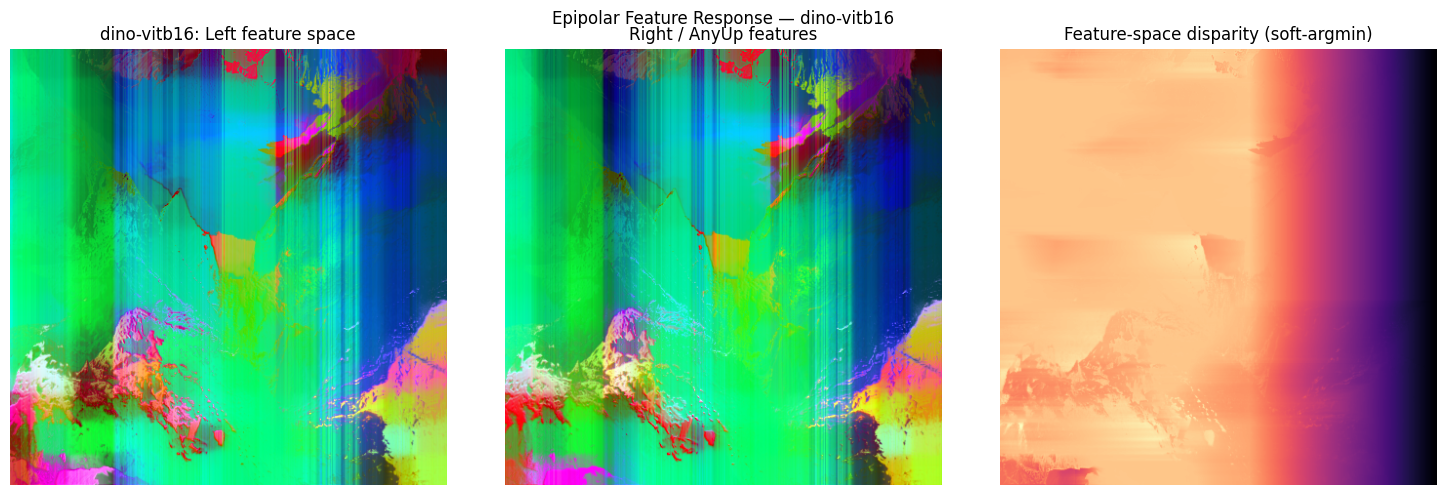}
         &
         \includegraphics[width=0.12\linewidth,trim={25.2cm 0 0 1.2cm},clip]{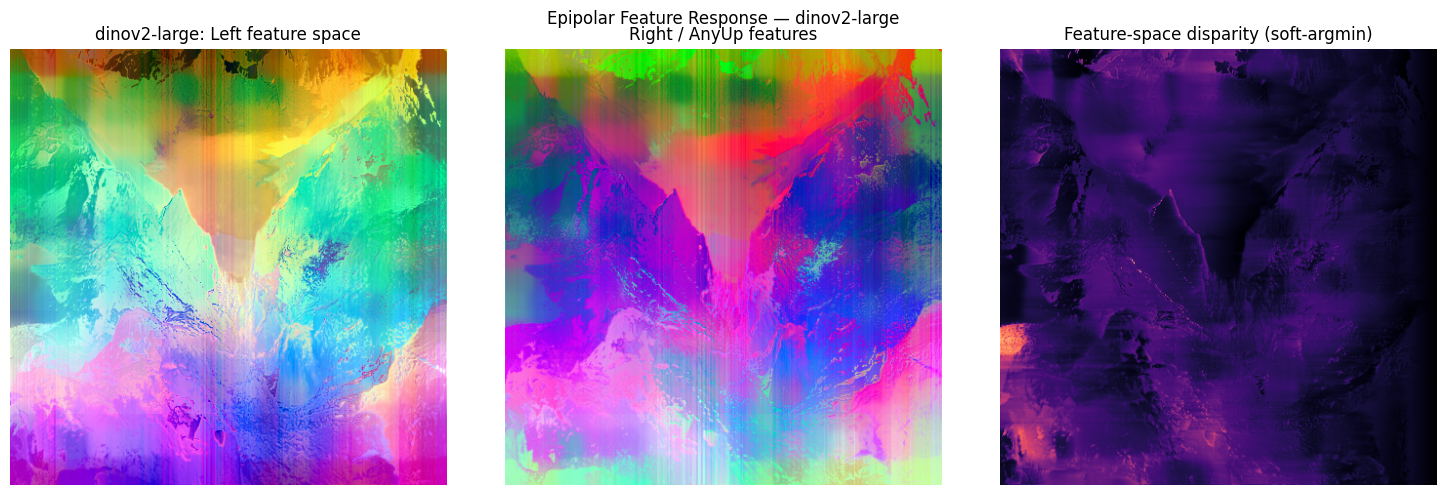} &
         \includegraphics[width=0.12\linewidth,trim={25.2cm 0 0 1.2cm},clip]{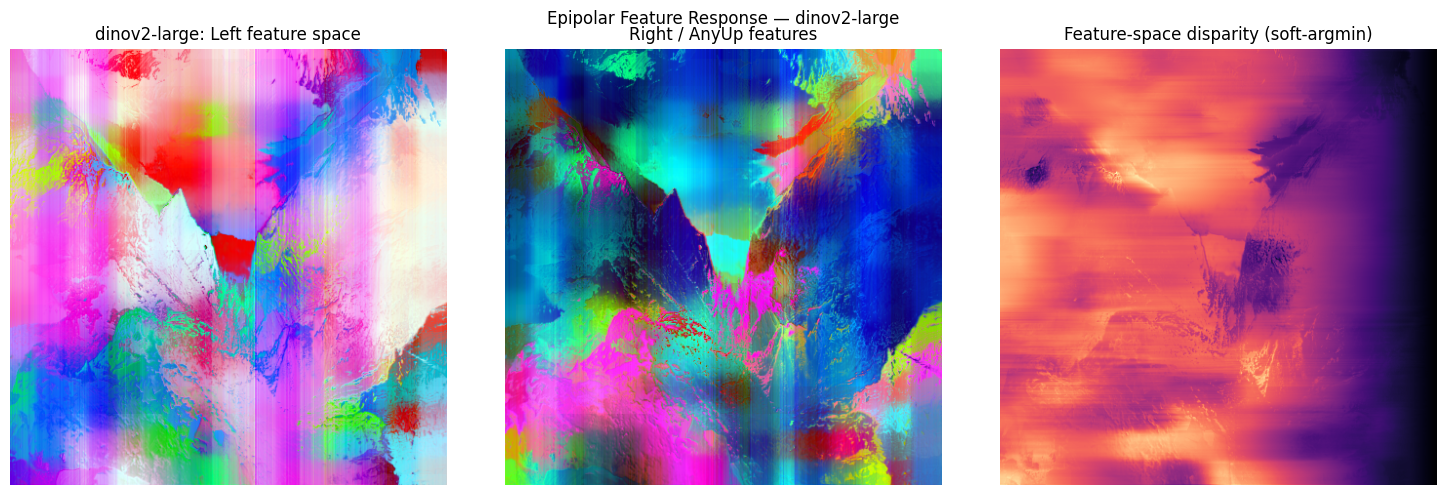} 
         &
         \includegraphics[width=0.12\linewidth,trim={25.2cm 0 0 1.2cm},clip]{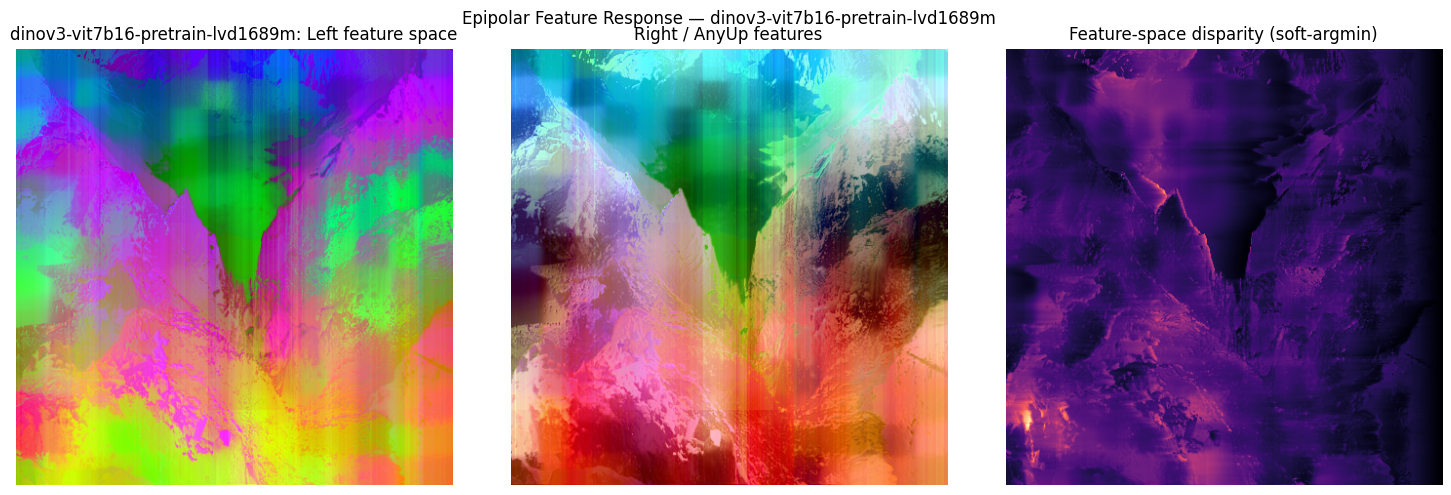} &
         \includegraphics[width=0.12\linewidth,trim={25.2cm 0 0 1.2cm},clip]{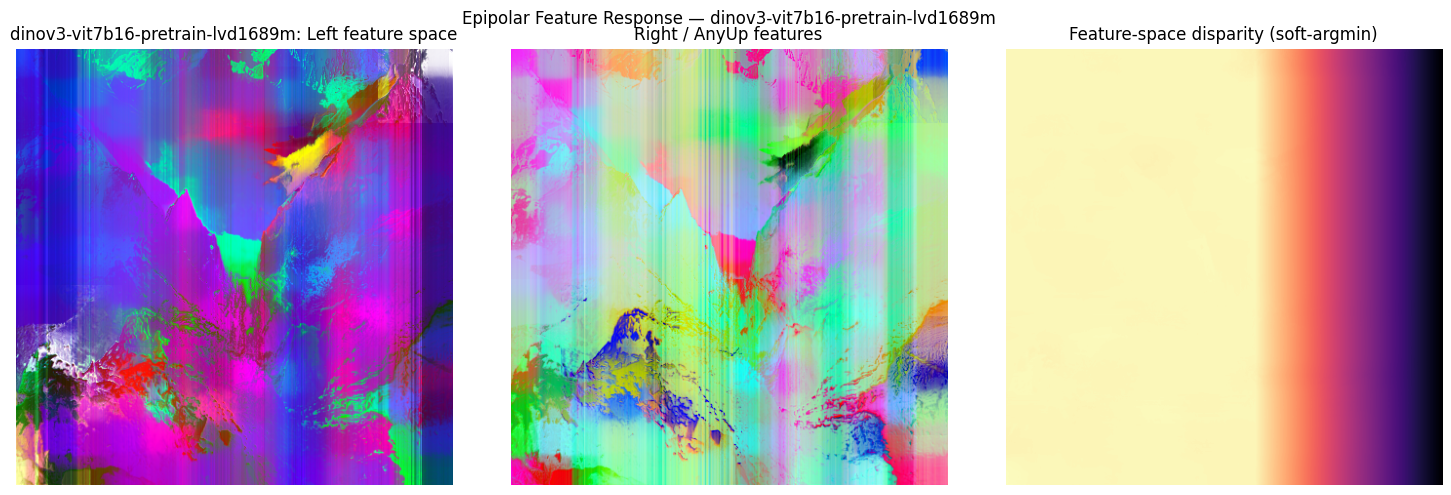}
         &
         \includegraphics[width=0.12\linewidth,trim={25.2cm 0 0 1.2cm},clip]{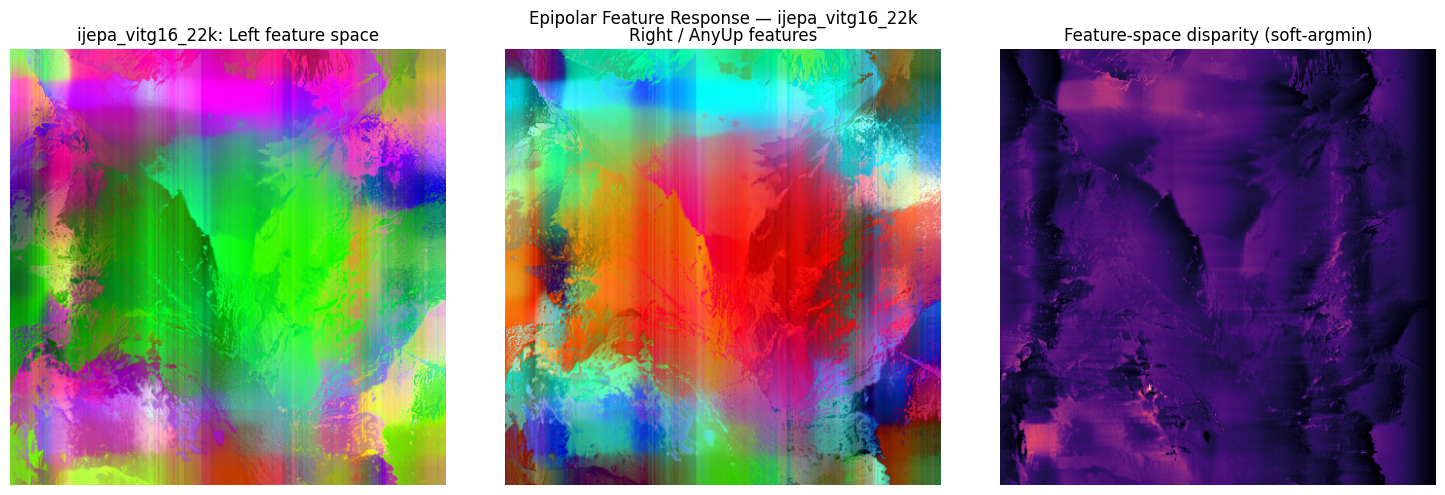} &
         \includegraphics[width=0.12\linewidth,trim={25.2cm 0 0 1.2cm},clip]{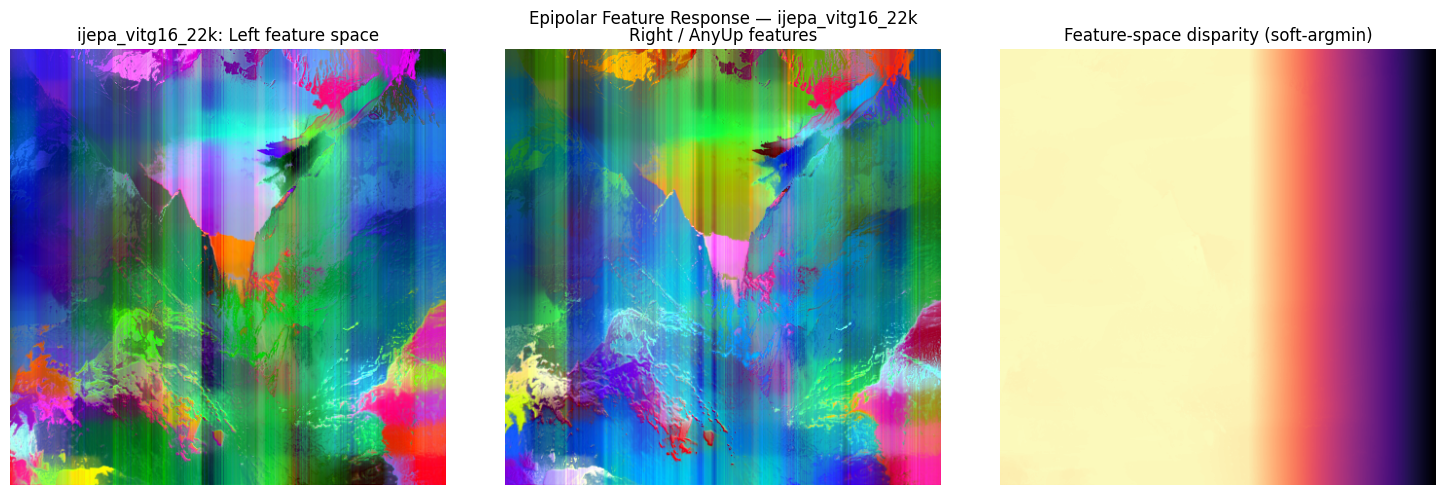} \\

         % {\textbf{MetaClip}} & {}
         % & {\textbf{DEiT}} & {}
         % & {\textbf{SigLip2}} & {} 
         % & {\textbf{BeiT}} & {} \\
        \multicolumn{2}{l}{\hspace{1em} \textbf{CLIP}}
        & \multicolumn{2}{l}{\hspace{1em} \textbf{DEiT}} 
        & \multicolumn{2}{l}{\hspace{1em} \textbf{SigLip2}} 
        & \multicolumn{2}{l}{\hspace{1em} \textbf{BeiT}}  \\
        \cmidrule(lr){1-1} \cmidrule(lr){3-3} \cmidrule(lr){5-5} \cmidrule(lr){7-7}

         {w/ PE} & {w/o PE}
         & {w/ PE} & {w/o PE}
         & {w/ PE} & {w/o PE} 
         & {w/ PE} & {w/o PE} \\
         \includegraphics[width=0.12\linewidth,trim={25.2cm 0 0 1.2cm},clip]{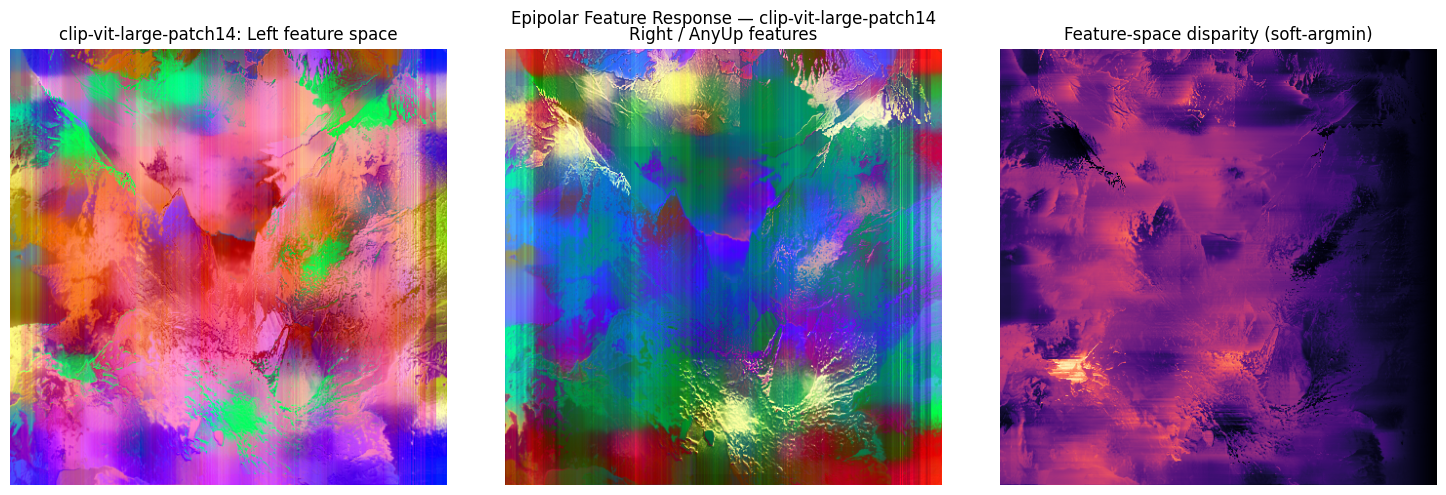} &
         \includegraphics[width=0.12\linewidth,trim={25.2cm 0 0 1.2cm},clip]{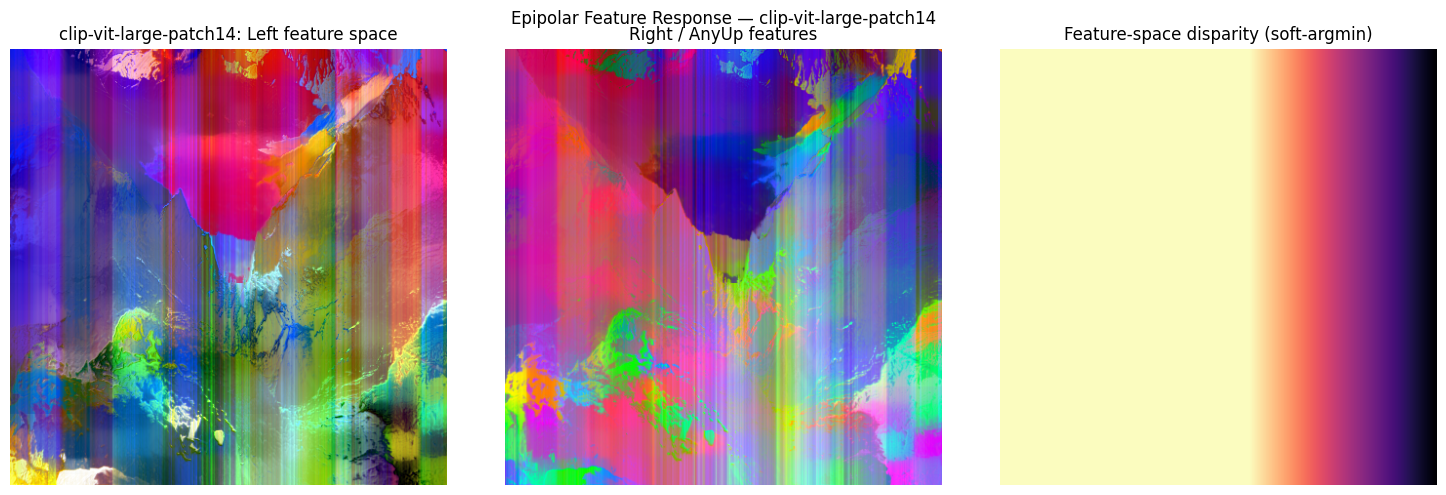}
         &
         \includegraphics[width=0.12\linewidth,trim={25.2cm 0 0 1.2cm},clip]{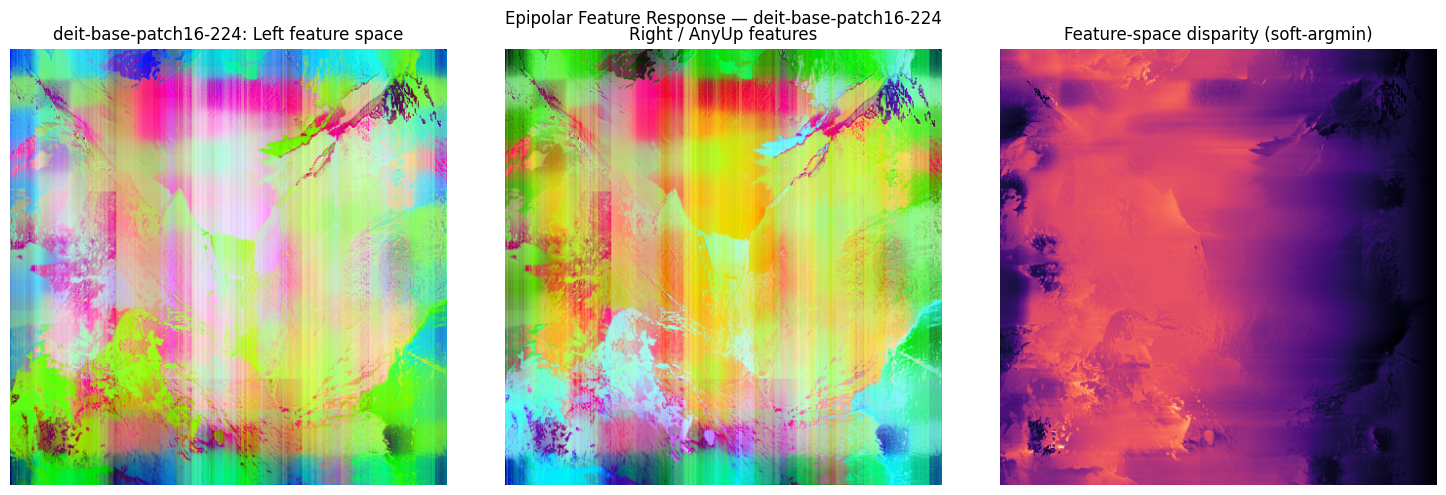} &
         \includegraphics[width=0.12\linewidth,trim={25.2cm 0 0 1.2cm},clip]{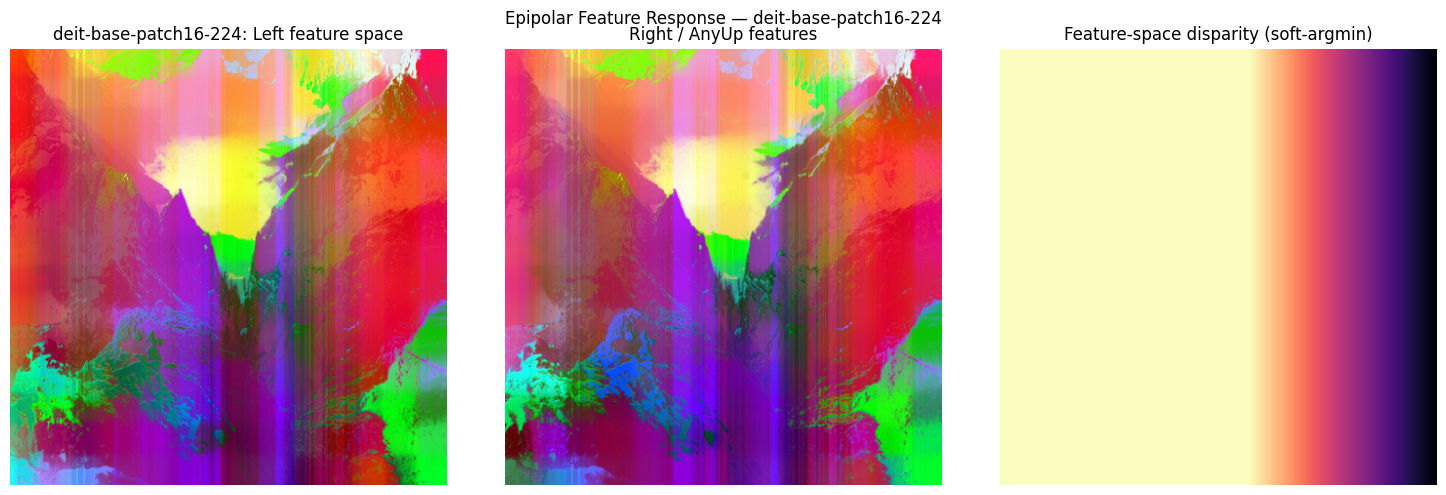} 
         &
         \includegraphics[width=0.12\linewidth,trim={25.2cm 0 0 1.2cm},clip]{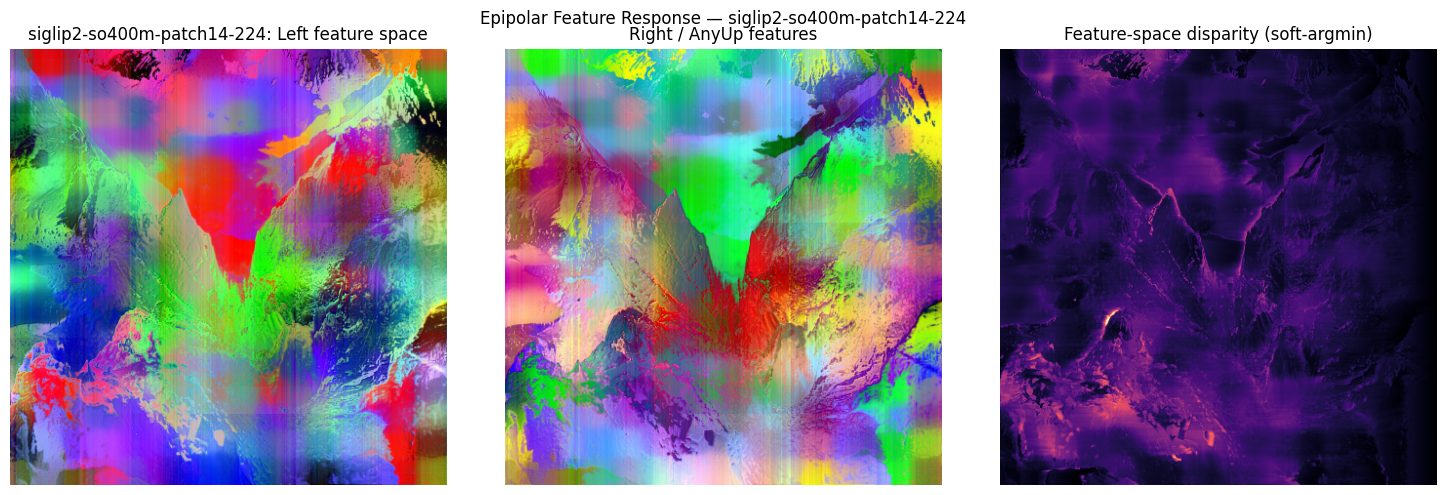} &
         \includegraphics[width=0.12\linewidth,trim={25.2cm 0 0 1.2cm},clip]{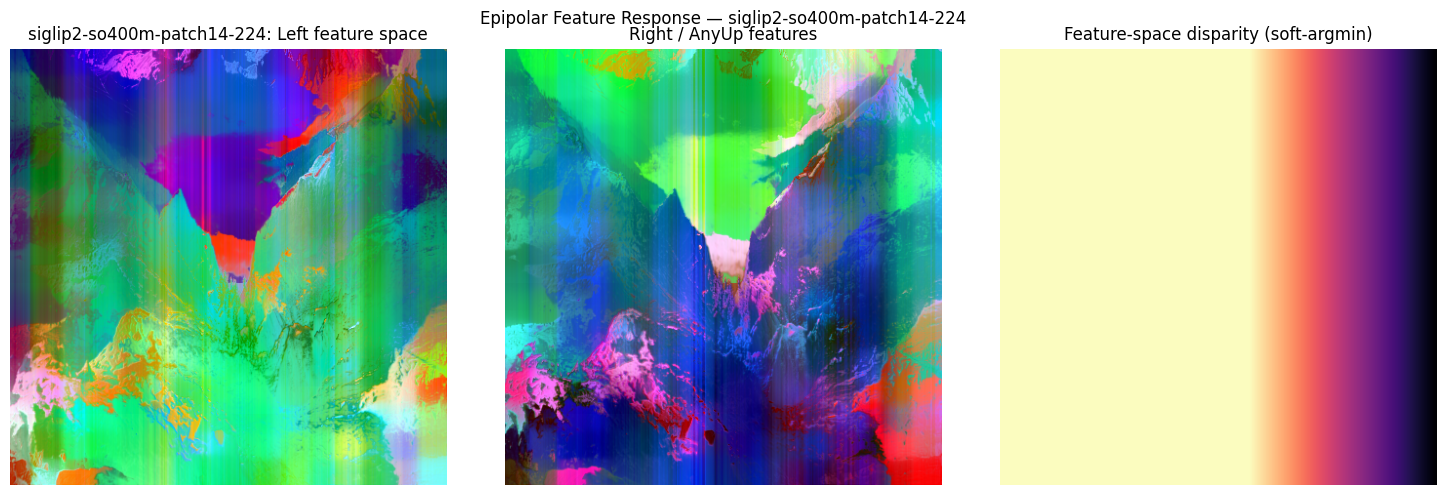}
         &
         \includegraphics[width=0.12\linewidth,trim={25.2cm 0 0 1.2cm},clip]{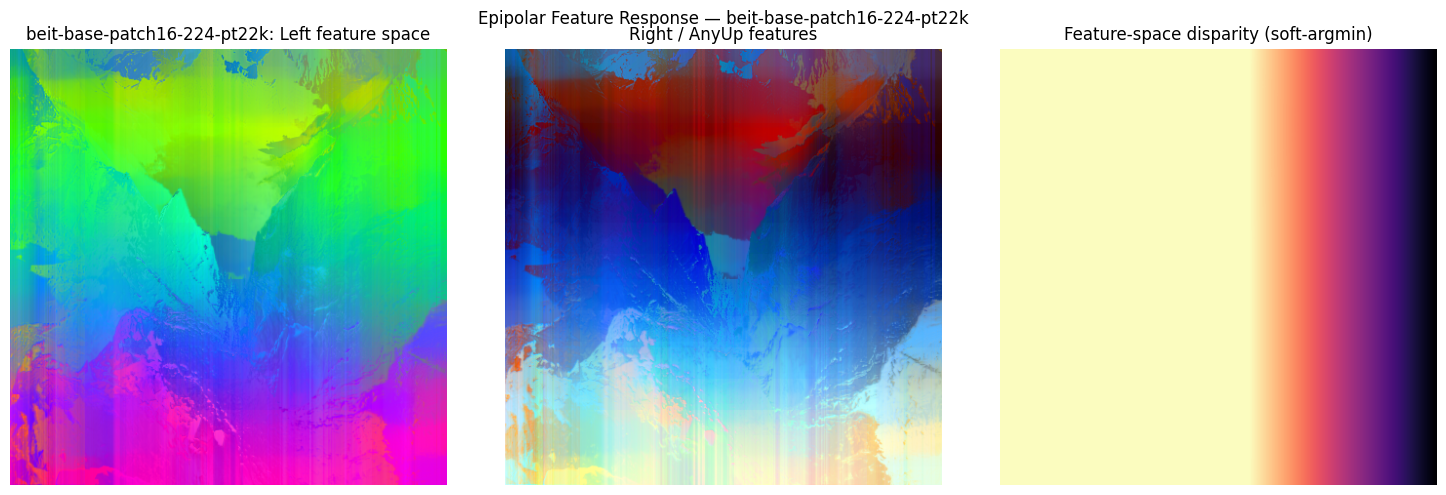} &
         \includegraphics[width=0.12\linewidth,trim={25.2cm 0 0 1.2cm},clip]{misc/disp/epipolar_response_pos_anyup_pca_beit-base-patch16-224-pt22k.png} \\
         
        \multicolumn{2}{l}{\hspace{1em} \textbf{Swin}}
        & \multicolumn{2}{l}{\hspace{1em} \textbf{SwinV2}} 
        & \multicolumn{2}{l}{\hspace{1em} \textbf{MLCD}} 
        & \multicolumn{2}{l}{\hspace{1em} \textbf{Data2Vec}}  \\
        \cmidrule(lr){1-1} \cmidrule(lr){3-3} \cmidrule(lr){5-5} \cmidrule(lr){7-7}

         {w/ PE} & {w/o PE}
         & {w/ PE} & {w/o PE}
         & {w/ PE} & {w/o PE} 
         & {w/ PE} & {w/o PE} \\
         \includegraphics[width=0.12\linewidth,trim={25.2cm 0 0 1.2cm},clip]{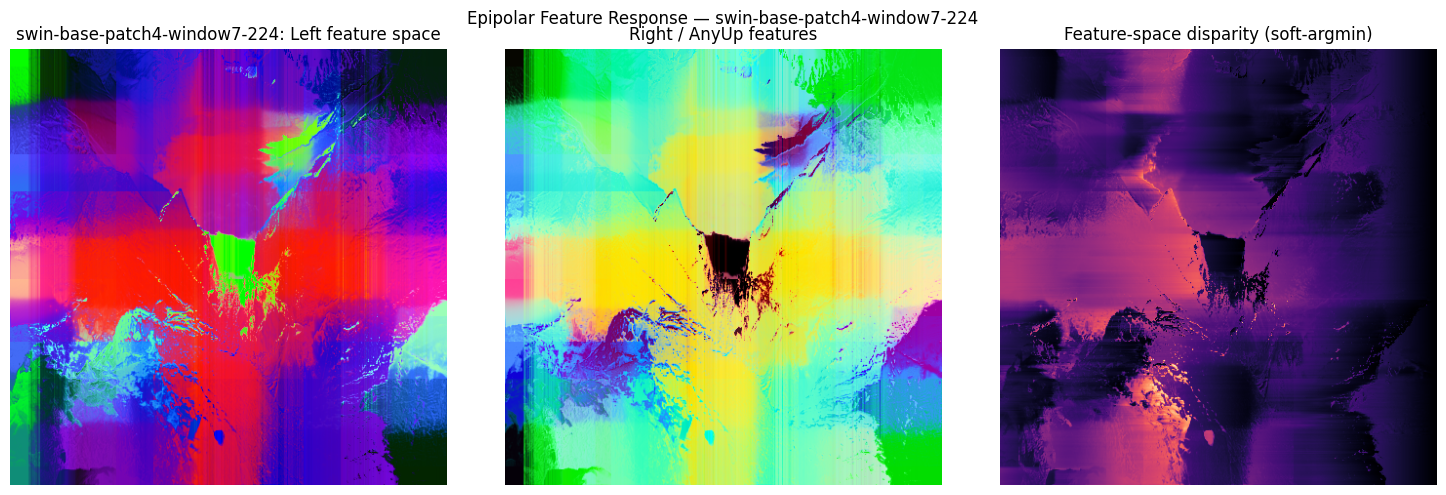} &
         \includegraphics[width=0.12\linewidth,trim={25.2cm 0 0 1.2cm},clip]{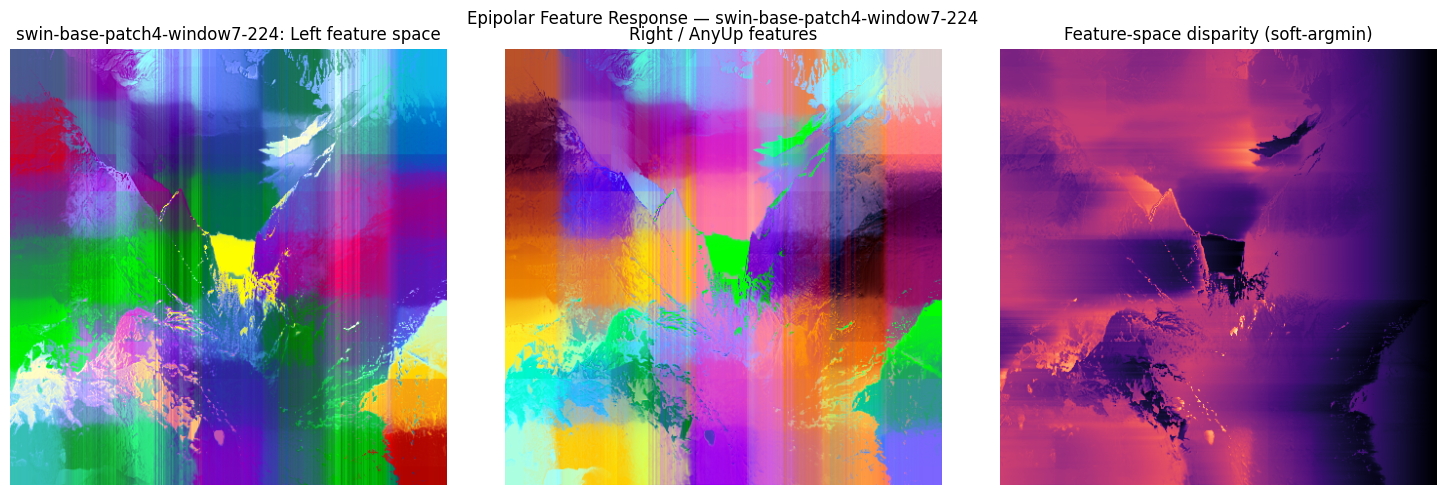}
         &
         \includegraphics[width=0.12\linewidth,trim={25.2cm 0 0 1.2cm},clip]{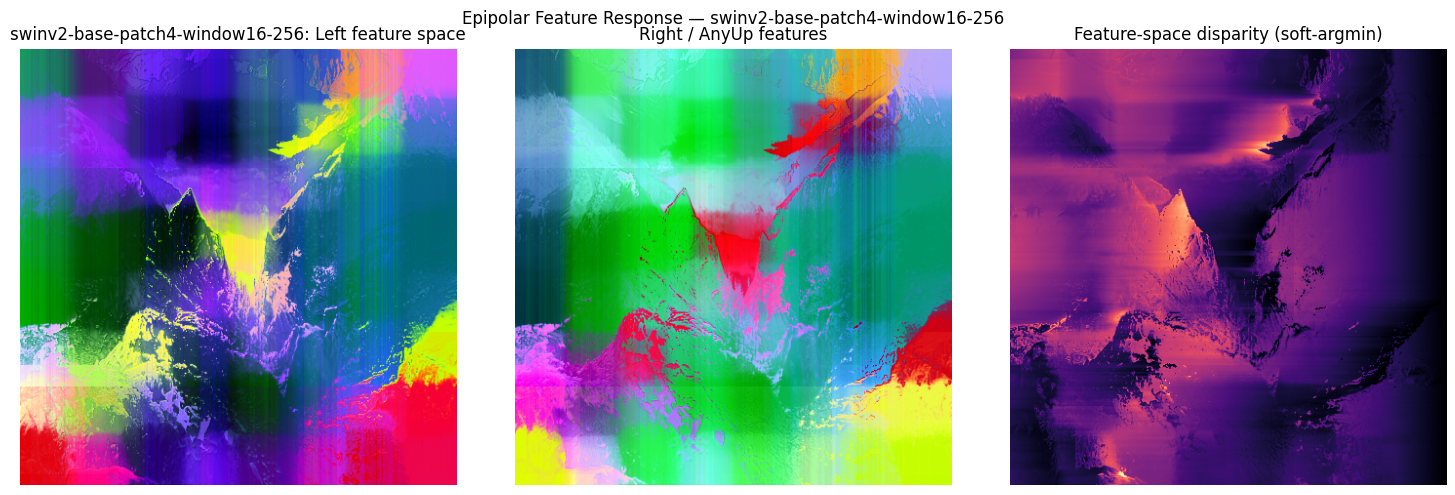} &
         \includegraphics[width=0.12\linewidth,trim={25.2cm 0 0 1.2cm},clip]{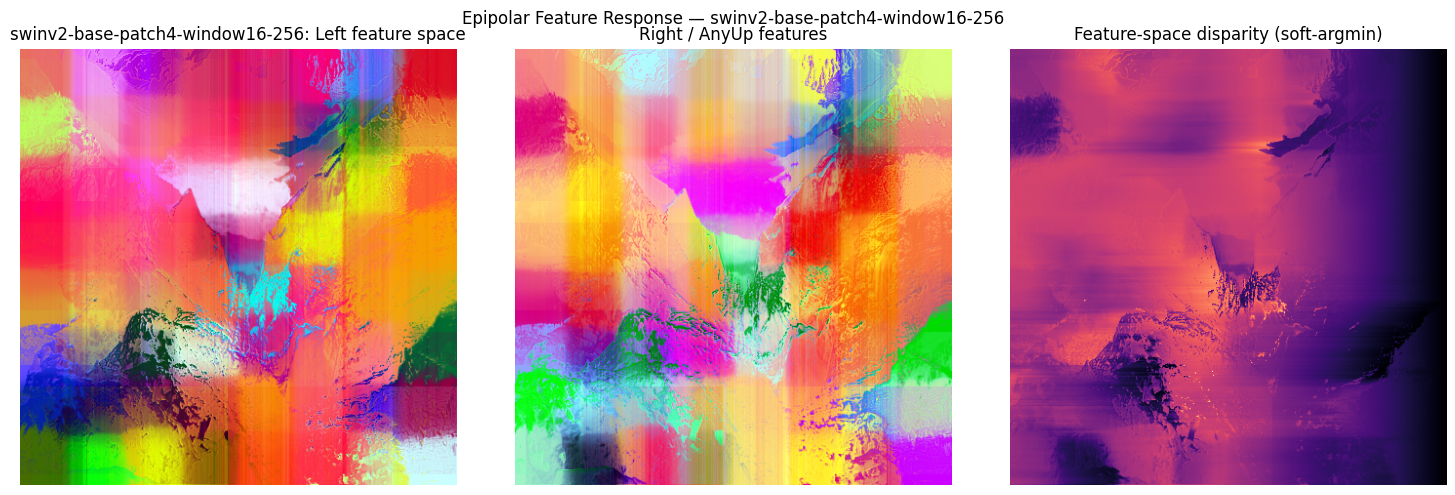} 
         &
         \includegraphics[width=0.12\linewidth,trim={25.2cm 0 0 1.2cm},clip]{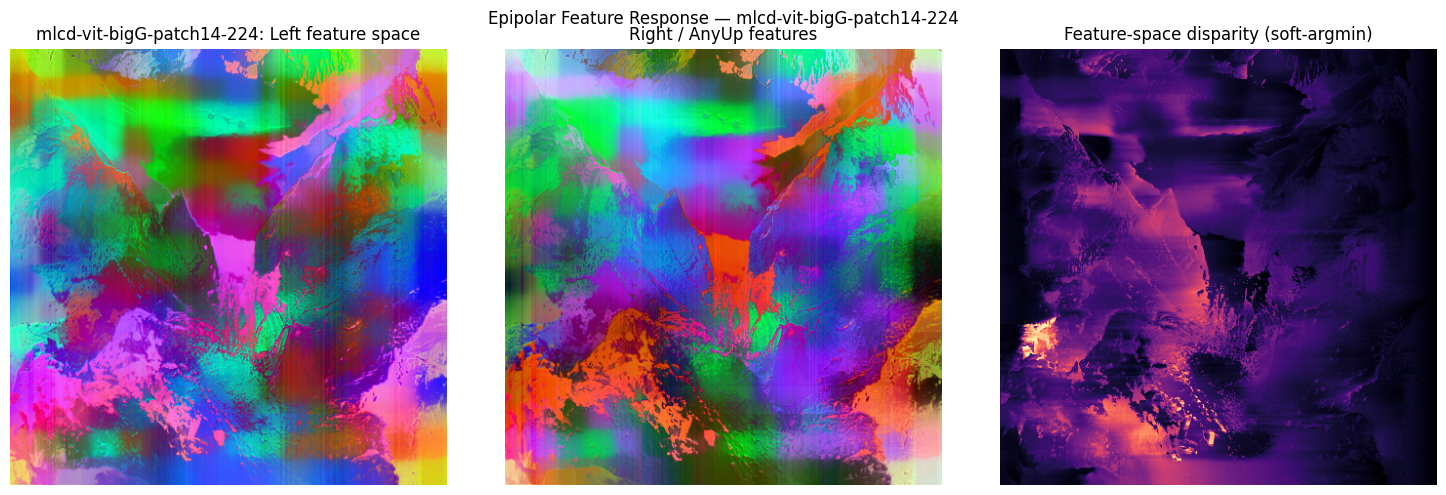} &
         \includegraphics[width=0.12\linewidth,trim={25.2cm 0 0 1.2cm},clip]{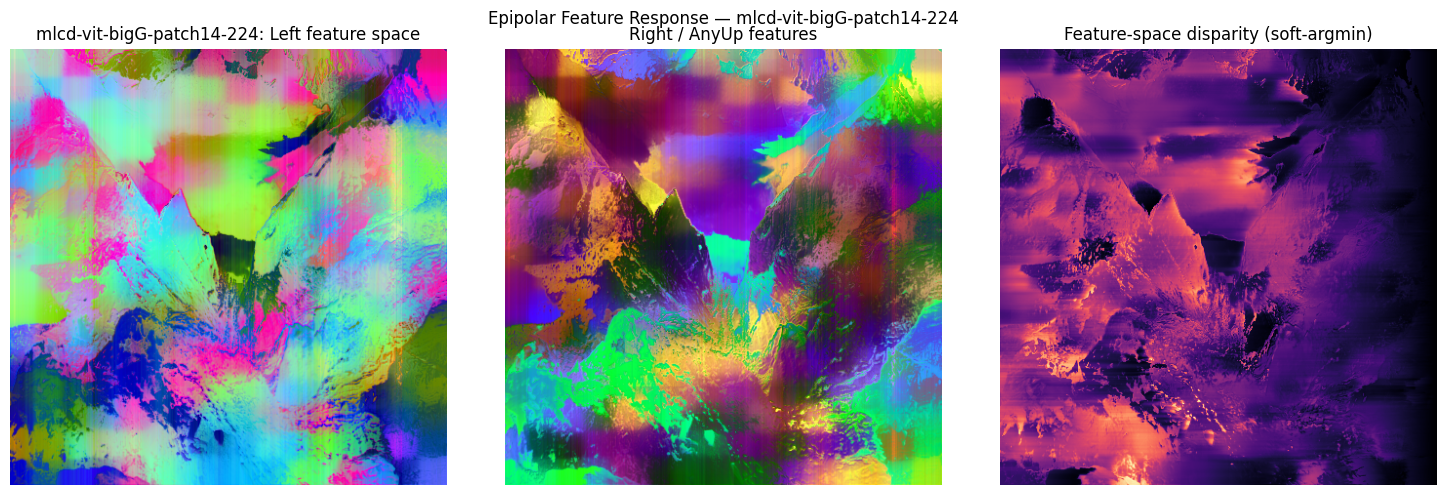}
         &
         \includegraphics[width=0.12\linewidth,trim={25.2cm 0 0 1.2cm},clip]{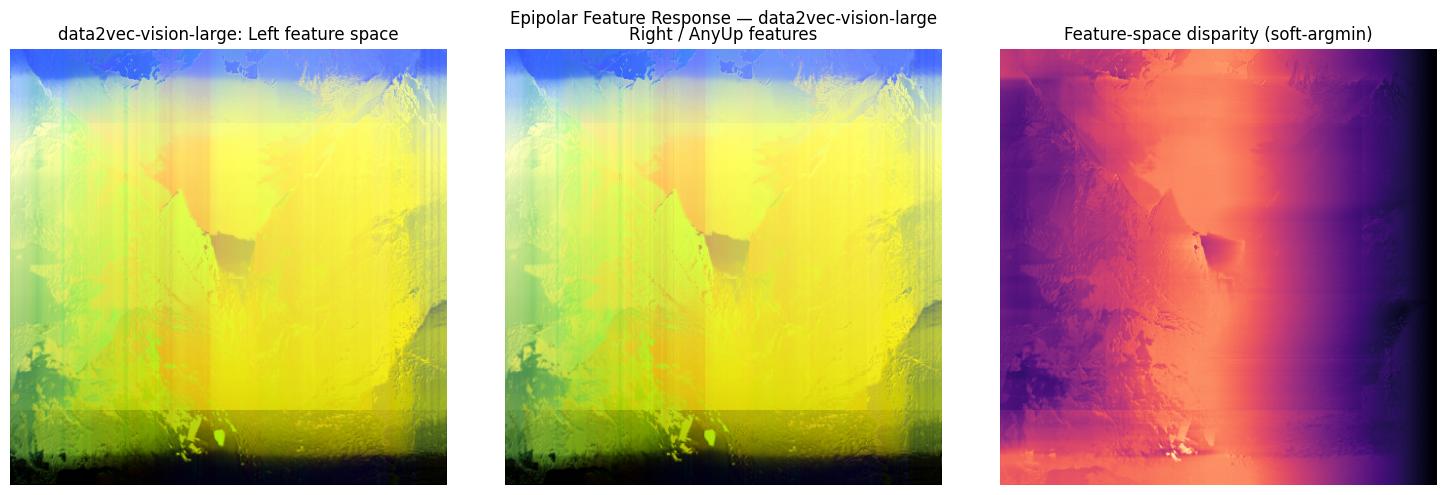} &
         \includegraphics[width=0.12\linewidth,trim={25.2cm 0 0 1.2cm},clip]{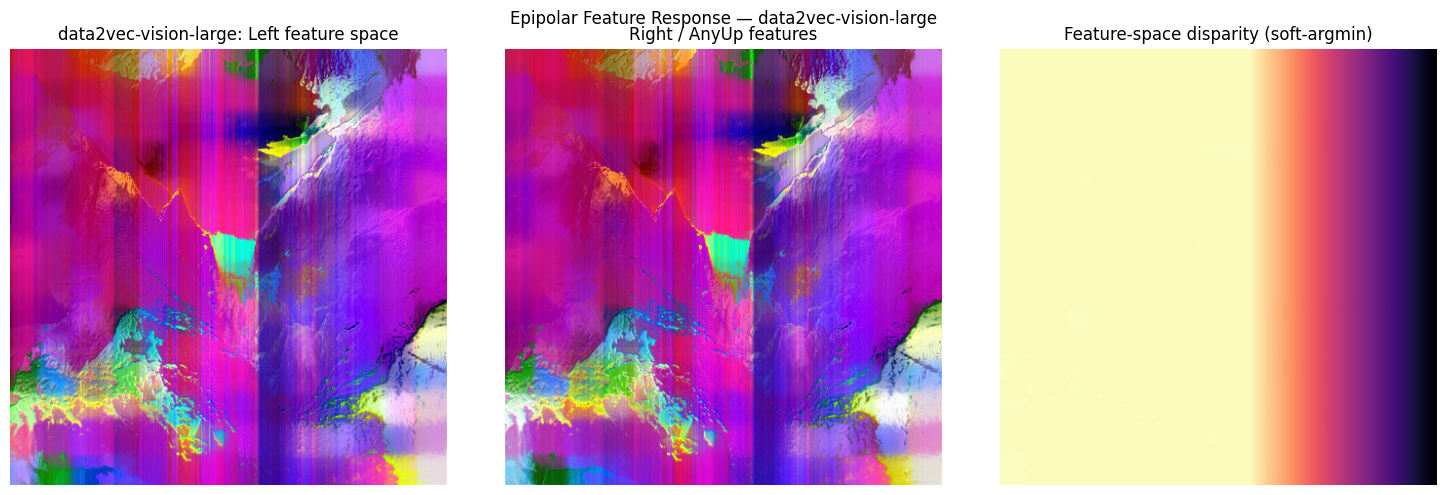}
         \\
    \end{tabular}
    \setlength{\belowcaptionskip}{-1.5em}
    \setlength{\abovecaptionskip}{0em}
    \caption{\textbf{Epipolar peak response visualization.} 
    \textbf{Top-left}: an example stereo pair and the corresponding SAM feature maps with and without PEs. Removing PE substantially weakens the spatial anchoring in the token representations.
    \textbf{Remaining panels} show the epipolar peak responses for each evaluated model. With PEs enabled, nearly all models (except BEiT) exhibit well-localized epipolar peaks, indicating preserved spatial anchoring. After removing PEs, epipolar responses vanish for most architectures, revealing that spatial anchoring degrades without positional signals.
    % Models such as DINO, DINOv2, Swin, SwinV2, and MLCD retain partial geometric cues, but their responses become noisy or unstable once PEs are removed.
    We use AnyUp~\cite{wimmer2025anyup} to obtain semantically meaningful feature visualizations. Please refer to the supplementary for (pairwise) shuffled PE visualizations.
    }
    \label{fig:epipolar_plot}
\end{figure*}

We use stereo pairs from the Spring~\cite{mehl2023spring} dataset at $448\times448$ resolution.
We sample the first ten pairs from each scene, resulting in a total of 370 pairs.
Left and right tokens are resized to $448\times 448$ for the 4D correlation volume construction.
We compute EPE and Recall@$n\; \text{and}\; n\in\{1,5\}$ to quantify geometric correspondence accuracy.
Lower EPE indicates higher correspondence precision, while higher Recall@$n$ reflects stronger epipolar consistency.
% Metric definitions are provided in the supplementary material.

\paragraph{Results.} We present quantitative and qualitative results in~\cref{tab:stereo_upsample_448} and~\cref{fig:epipolar_plot}, respectively. Quantitative metrics are derived from bilinearly upsampled tokens, whereas AnyUp~\cite{wimmer2025anyup} is used for producing semantically meaningful visualizations. Additional visualization for shuffled PEs are provided in the supplementary.

With vanilla PEs, almost all models (except BeiT, discussion provided in supplementary) attain sub-patch correspondence, with mean EPE below the effective patch size.
Epipolar peak responses exhibit sharp peaks (except BeiT and SAM), indicating that the PE-induced kernel provides spatial anchoring.
However, zeroing out or shuffling PEs can significantly collapse the correspondence. EPE typically increases beyond one patch size, Recall@$n$ drops substantially, and epipolar responses become diffuse or unstable. 
Interestingly, though visually uninterpretable (see supplementary), pairwise shuffling largely remains correspondence for the models with absolute PEs. This behavior indicates that, for absolute PEs, geometric correspondence depends primarily on the consistency of PEs across views rather than on their precise spatial correctness.

\subsection{On The Role of PE Consistency}

We observe that pairwise shuffling, where both views share the same permutation of PEs, does not always degrade correspondence as severely as independent random shuffling. Although the resulting spatial organization can be distorted or visually corrupted (see supplementary), a permuted but shared PE frame allows tokens to remain partially anchored across views from token representations, especially with absolute PEs.
In contrast, independently shuffled PEs across views collapse the entire spatial anchoring.

A natural interpretation is that, geometric alignment in ViT representations does not solely depend on the visual content, but also on the consistency of PEs across views.

% \paragraph{Residual geometric structure without PEs.}
% Despite the overall degradation, several architectures, including DINO, DINOv2, Swin, SwinV2, and MLCD, retain weak but non-negligible epipolar peak responses after PE kernels altered.
% While correspondence accuracy is substantially reduced, response profiles remain partially localized.
% This indicates that positional embeddings are not the sole source of spatial structure. Architectural inductive biases such as local attention windows, hierarchical feature aggregation, or multi-scale processing may implicitly favor spatial coherence.

\begin{takeaway}
$\S$~ \textbf{Positional embeddings go beyond providing coordinates.}
They contribute substantially to the spatial organization of ViT representations.

\vspace{.5em}
$\S$~ \textbf{Spatial scaffolding requires positional consistency.}
With consistent PEs, cross-view tokens are conditioned on the same PE kernel, allowing spatial relationships to be preserved.
\end{takeaway}

\section{Consistent PEs for Multi-View Geometry}

Now, we identify inconsistent PEs as a primary cause of geometric degradation in ViTs. We next study approaches for reconciling the inconsistency for multi-view geometry.

\subsection{Restoring PE Consistency via Token Re-Indexing}
\label{sec:reindexing}
In representation learning, geometric correspondence is typically attributed to shared visual content across views.
However, the representations in ViTs are anchored to an internal positional reference frame. A natural question follows:
\begin{quoting}
\textit{How does internal positional reference frame across views affect spatial anchoring in ViTs?}
\end{quoting}
Our hypothesis is simple: ViT representations live in a model's internal positional reference frame.
Therefore, for multi-view tasks, scene-level geometric structure may naturally emerge when this internal frame is consistently aligned across views.
We propose a simple, training-free token re-indexing operation. 

\begin{figure}[h]
\centering
\scalebox{.8}{%
\input{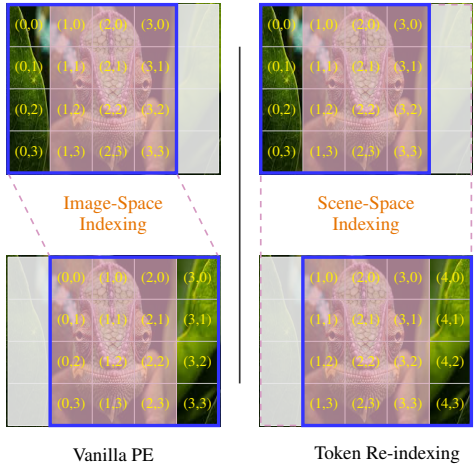}
}
\setlength{\belowcaptionskip}{-1em}
\setlength{\abovecaptionskip}{0em}
\caption{\textbf{Illustration of token re-indexing.} The upper and bottom represent two feature maps that contain overlapping regions (in pixel space, annotated in red color) along the horizontal axis.}
\label{fig:overlap}
\end{figure}

\begin{table}[b]
    \centering
    \scriptsize
    \setlength{\tabcolsep}{1.5pt}
    \rowcolors{5}{}{verylightgray}
    \setlength{\abovecaptionskip}{0em}
    \setlength{\belowcaptionskip}{-1em}
    \caption{\textbf{Token re-indexing performance.} Aligned PE kernels increase similarity, indicating improved spatial consistency.}
    \begin{tabular}{l|c|ccc | ccc}
        \toprule
         & & \multicolumn{3}{c|}{Vanilla PE} & \multicolumn{3}{c}{Aligned PE} \\
         \cmidrule(lr){3-5}\cmidrule(lr){6-8}
         & PE & $\Delta x=1$ & $\Delta x=2$ & $\Delta x=3$ & $\Delta x=1$ & $\Delta x=2$ & $\Delta x=3$ \\
        \midrule
        \textbf{DINO}     & Abs. &   0.9782 & 0.9765 & 0.9627  & 0.9980 & 0.9809 & 0.9766   \\
        \textbf{DINOv2}   & Abs. &   0.6775 & 0.5760 & 0.5272  & 0.7446 & 0.6318 & 0.5782  \\
        \textbf{DINOv3}   & Rot. &   0.9261 & 0.7377 & 0.6564  & 0.9744 & 0.9657 & 0.9616  \\
        \textbf{MLCD}     & Rot. &   0.8209 & 0.7345 & 0.6786  & 0.8412 & 0.7660 & 0.7175 \\
        \textbf{I-JEPA}    & Abs. &   0.7318 & 0.7283 & 0.7280  & 0.8646 & 0.8576 & 0.8568  \\
        % \textbf{MetaCLIP} & Abs. &   0.7836 & 0.7589 & 0.7496  & 0.9799 & 0.9281 & 0.9109   \\
        \textbf{DeiT}     & Abs. &   0.8184 & 0.8003 & 0.7947  & 0.9400 & 0.9107 & 0.9031  \\
        \textbf{SigLIP2}  & Abs. &   0.6468 & 0.6920 & 0.6872  & 0.7143 & 0.6312 & 0.6199   \\ 
        \textbf{BEiT}     & Rel. &   0.9998 & 0.9997 & 0.9997  & 0.9999 & 0.9999 & 0.9999  \\
        \textbf{Data2Vec} & Rel. &   0.9850 & 0.9678 & 0.9469  & 0.9978 & 0.9948 & 0.9997 \\
        \textbf{ViT}      & Abs. &   0.7913 & 0.7737 & 0.7416  & 0.9498 & 0.9354 & 0.9338   \\
        \textbf{CLIP}     & Abs. &   0.6758 & 0.6665 & 0.6711  & 0.9234 & 0.8844 & 0.8652   \\
        \textbf{SAM}      & Abs. &   0.9969 & 0.9945 & 0.9921  & 0.9977 & 0.9971 & 0.9971    \\
        % SegFormer& N/A. &   0.9723 & 0.9529 & 0.9315  & -      & -      & - \\
        \textbf{Swin}     & Rel. &   0.6928 & 0.6928 & 0.5899  & 0.8448 & 0.8293 & 0.8189 \\
        \bottomrule
    \end{tabular}
    \label{tab:positional_probe}
\end{table}

\paragraph{Setup.}
We adopt the same content-position isolation settings as per~\Cref{sec:overlap}. With two overlapping crops $\mathbf{I}_1$ and $\mathbf{I}_2$, we introduce a {training-free token re-indexing} procedure that enforces PE kernel alignment across views, as illustrated in \cref{fig:overlap}.
Simply, we ensure that tokens that correspond to the same scene-space location within the overlaps are assigned with identical positional indices.
The details of the token re-indexing for different positional encoding methods are presented in the supplementary.

\paragraph{Results.}
Table~\ref{tab:positional_probe} reports token-wise cosine similarities before and after re-indexing across all positional encoding types.
With aligned PEs, token similarities are consistently restored across all models, without any training.
This ``free-lunch" improvement provides direct evidence that the observed degradation arises from model's view-dependent internal positional reference frames, rather than intrinsic differences in architectural modeling.

Token re-indexing is not proposed as a practical algorithm, but as a controlled intervention to test whether positional alignment alone is sufficient to restore spatial anchoring.
The success of this training-free alignment indicates that geometric information is preserved in the representation but expressed relative to mismatched PEs in multi-view tasks.

\subsection{Implicit Positional Alignment in VGGT}

With the role of PE kernel misalignment clarified, it remains an open question that:
\begin{quoting}
\textit{Why do some multi-view systems succeed despite being built on ViT backbones with standard positional embeddings?}
\end{quoting}
VGGT provides a particularly instructive case.
It employs a ViT backbone to extract per-view token features, followed by an aggregator that jointly processes tokens across views.
Importantly, the positional encoding mechanism of its ViT backbone (DINOv2) is left unchanged.
In this section, we analyze the behavior of VGGT for multi-view geometry under the lens of PE kernels.

\paragraph{Setup.}
To understand how VGGT achieves robust multi-view representation, we analyze the layer-wise evolution of cross-view token similarity within its aggregator.
We use the same overlapping-crop dataset construction as in \cref{sec:overlap}, which induces positional reference mismatch while preserving identical visual content.

First, for corresponding tokens across views, we measure cosine similarity as a function of aggregator depth (24-layer in total). We repeat this analysis for increasing horizontal displacements $\Delta x \in \{1,2,3\}$.
Second, for each aggregation layer, we train lightweight MLP to predict the row and column indices from individual token features under $\Delta x=3$, for probing how much positional information are preserved for each layer.
These experiments provide a controlled way to probe the evolution of positional information and spatial consistency across aggregation layers.

\begin{figure}
    \centering
    \begin{subfigure}{0.85\linewidth}
        \resizebox{\linewidth}{!}{\scriptsize% Requires: \usepackage{pgfplots}
%           \pgfplotsset{compat=1.18}
\begin{tikzpicture}
\begin{axis}[
    width=\linewidth,
    height=4.5cm,
    ylabel={Token similarity},
    xmin=-1, xmax=24,
    ymin=0.55, ymax=1.00,
    grid=both,
    major grid style={line width=.2pt,draw=gray!40},
    minor grid style={line width=.1pt,draw=gray!25},
    legend pos=south east,
    legend cell align=left,
    legend style={font=\scriptsize, nodes={scale=0.7, transform shape}},
    tick align=outside,
    tick pos=left,
    axis lines=left,
]

% delta_x = 1
\addplot+[mark=*, mark size=1.2pt] coordinates {
(0,0.8770315914192899)
(1,0.8198645116112625)
(2,0.7988911916913423)
(3,0.7905532698281671)
(4,0.8467383713683383)
(5,0.7690014697140929)
(6,0.758098654494509)
(7,0.7477587999248699)
(8,0.7389269279365385)
(9,0.7318569522525531)
(10,0.7298484816085054)
(11,0.745523635821527)
(12,0.7422187063698856)
(13,0.7897737522717648)
(14,0.8269222048538037)
(15,0.807065166669562)
(16,0.8406469205246437)
(17,0.9290337499435953)
(18,0.9295404455083937)
(19,0.9197582346358756)
(20,0.9090439864185823)
(21,0.925110226007683)
(22,0.9537618825489051)
(23,0.9902856227336736)
};
\addlegendentry{\scriptsize $\Delta x=1$}

% delta_x = 2
\addplot+[mark=square*, mark size=1.2pt] coordinates {
(0,0.8300649391415162)
(1,0.7538915185481614)
(2,0.7274246902912551)
(3,0.7181287628812848)
(4,0.7951239453071004)
(5,0.6924735180229373)
(6,0.6790306654335768)
(7,0.6665620659373686)
(8,0.6565068127424314)
(9,0.6498441732585308)
(10,0.652317312487274)
(11,0.6785101028663806)
(12,0.6800347592097435)
(13,0.7437689559280023)
(14,0.7920820486035707)
(15,0.7685950684935886)
(16,0.8070020704793833)
(17,0.9131539421508851)
(18,0.9117292512944177)
(19,0.8905046937907544)
(20,0.8860658623293315)
(21,0.9048269200470686)
(22,0.9407650167976037)
(23,0.9872641329124358)
};
\addlegendentry{\scriptsize $\Delta x=2$}

% delta_x = 3
\addplot+[mark=triangle*, mark size=1.4pt] coordinates {
(0,0.801331842382668)
(1,0.7142230573351175)
(2,0.6850645398899396)
(3,0.675663906178018)
(4,0.7651889369104157)
(5,0.6481655562967973)
(6,0.6334859422899313)
(7,0.6200467522906674)
(8,0.6096443527827681)
(9,0.6037628989355627)
(10,0.6096429282196185)
(11,0.6421333319543578)
(12,0.6466311612100076)
(13,0.7197873960201706)
(14,0.7727665739729546)
(15,0.7459645868561661)
(16,0.7869508721677931)
(17,0.9026820879361295)
(18,0.8990268144005428)
(19,0.8686172123353496)
(20,0.8701257668056216)
(21,0.8904030423785907)
(22,0.9311924125172213)
(23,0.984803654875629)
};
\addlegendentry{\scriptsize $\Delta x=3$}

\end{axis}
\end{tikzpicture}}
    \end{subfigure}
    \begin{subfigure}{0.85\linewidth}
        \resizebox{\linewidth}{!}{\scriptsize% Requires: \usepackage{pgfplots}
% Optional: \pgfplotsset{compat=1.18}
\begin{tikzpicture}
    \begin{axis}[
        width=\linewidth,
        height=4cm,
        xlabel={Aggregator layer index},
        ylabel={Pos. Pred. accuracy},
        xmin=-1, xmax=24,
        ymin=0.70, ymax=1.05,
        xtick={0,5,10,15,20},
        ytick={0.7,0.8,0.9,1.0},
        grid=both,
        major grid style={line width=.2pt,draw=gray!40},
        minor grid style={line width=.1pt,draw=gray!25},
    legend style={font=\scriptsize, nodes={scale=0.7, transform shape}},
        legend pos=south east,
        legend cell align=left,
        tick align=outside,
        tick pos=left,
        axis lines=left,
    ]

% \addplot+[mark=*, mark size=1.2pt] coordinates {
    % Row
    \addplot+[
        thick,
        mark=*,
        mark size=1.4pt,
    ] coordinates {
        (0,0.864) (1,0.933) (2,0.928) (3,0.911) (4,0.908) (5,0.900)
        (6,0.893) (7,0.891) (8,0.899) (9,0.908) (10,0.912) (11,0.944)
        (12,0.945) (13,0.976) (14,0.992) (15,0.992) (16,0.996) (17,0.995)
        (18,0.998) (19,0.998) (20,0.999) (21,0.999) (22,0.999) (23,0.999)
    };
    \addlegendentry{\scriptsize Row}

% Epoch 1 Index 0 test Row 0.864 Col 0.900 Both 0.780
% Epoch 1 Index 1 test Row 0.933 Col 0.947 Both 0.885
% Epoch 1 Index 2 test Row 0.928 Col 0.914 Both 0.849
% Epoch 1 Index 3 test Row 0.911 Col 0.905 Both 0.826
% Epoch 1 Index 4 test Row 0.908 Col 0.897 Both 0.816
% Epoch 1 Index 5 test Row 0.900 Col 0.888 Both 0.801

% Epoch 1 Index 6 test Row 0.893 Col 0.876 Both 0.784
% Epoch 1 Index 7 test Row 0.891 Col 0.875 Both 0.782
% Epoch 1 Index 8 test Row 0.899 Col 0.879 Both 0.792
% Epoch 1 Index 9 test Row 0.908 Col 0.868 Both 0.790
% Epoch 1 Index 10 test Row 0.912 Col 0.881 Both 0.805
% Epoch 1 Index 11 test Row 0.944 Col 0.882 Both 0.833

% Epoch 1 Index 12 test Row 0.945 Col 0.880 Both 0.833
% Epoch 1 Index 13 test Row 0.976 Col 0.972 Both 0.949
% Epoch 1 Index 14 test Row 0.992 Col 0.980 Both 0.972
% Epoch 1 Index 15 test Row 0.992 Col 0.986 Both 0.978
% Epoch 1 Index 16 test Row 0.996 Col 0.993 Both 0.989
% Epoch 1 Index 17 test Row 0.995 Col 0.994 Both 0.989

% Epoch 1 Index 18 test Row 0.998 Col 0.998 Both 0.996
% Epoch 1 Index 19 test Row 0.998 Col 0.998 Both 0.997
% Epoch 1 Index 20 test Row 0.999 Col 0.998 Both 0.996
% Epoch 1 Index 21 test Row 0.999 Col 0.997 Both 0.996
% Epoch 1 Index 22 test Row 0.999 Col 0.997 Both 0.996
% Epoch 1 Index 23 test Row 0.999 Col 0.994 Both 0.994

    % Col
    \addplot+[
        thick,
        mark=square*,
        mark size=1.4pt,
    ] coordinates {
        (0,0.900) (1,0.947) (2,0.914) (3,0.905) (4,0.897) (5,0.888)
        (6,0.876) (7,0.875) (8,0.879) (9,0.868) (10,0.881) (11,0.882)
        (12,0.880) (13,0.972) (14,0.980) (15,0.986) (16,0.993) (17,0.994)
        (18,0.998) (19,0.998) (20,0.998) (21,0.997) (22,0.997) (23,0.994)
    };
    \addlegendentry{\scriptsize Col}

    % Both
    \addplot+[
        thick,
        mark=triangle*,
        mark size=1.6pt,
    ] coordinates {
        (0,0.780) (1,0.885) (2,0.849) (3,0.826) (4,0.816) (5,0.801)
        (6,0.784) (7,0.782) (8,0.792) (9,0.790) (10,0.805) (11,0.833)
        (12,0.833) (13,0.949) (14,0.972) (15,0.978) (16,0.989) (17,0.989)
        (18,0.996) (19,0.997) (20,0.996) (21,0.996) (22,0.996) (23,0.994)
    };
    \addlegendentry{\scriptsize Both}

    \end{axis}
\end{tikzpicture}}
    \end{subfigure}
\setlength{\belowcaptionskip}{-2em}
\setlength{\abovecaptionskip}{0em}
    \caption{\textbf{Layer-wise evaluation for the aggregator of VGGT}.
    \textbf{Upper:} \textit{cross-view token similarity}. We report cosine similarity between corresponding tokens across views as a function of the VGGT aggregator depth (24 layers), for increasing horizontal displacements $\Delta x=\{1,2,3\}$. As depth increases, similarity consistently recovers and converges toward near-perfect ($\text{Sim}\approx1$) alignment for all displacements. \textbf{Lower:} \textit{position decodability}. we report probing accuracy for predicting the positional index (row, column, both) for each token under $\Delta x=3$.}
    \label{fig:vggt_delta}
\end{figure}
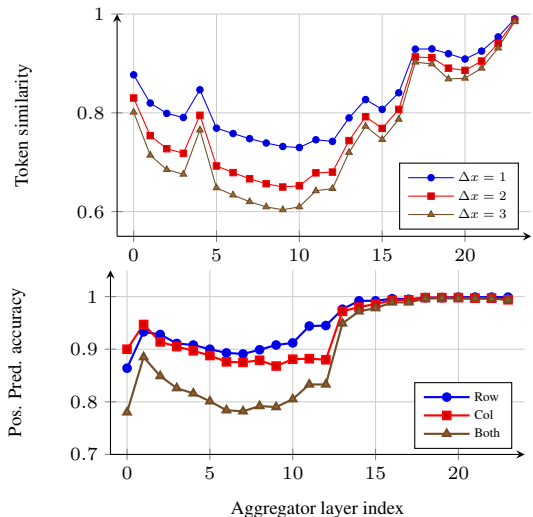

\paragraph{Results.}
As shown in~\cref{fig:vggt_delta}, across all displacement magnitudes, similarity initially decreases in the early layers, reaching a minimum around layers 6--10.
This effect is more pronounced for larger $\Delta x$, where positional reference mismatch is stronger.
Beyond the mid-depth layers, similarity begins to recover and increases rapidly in the later stages of the aggregator.
By the final layers, similarity converges toward near-perfect alignment for all $\Delta x$ values.

Similarly, the positional decoding probes present a complementary behavior. The position indices remain highly decodable throughout the aggregator, while the positional signals surge significantly at the $13^{\text{th}}$ layer, reaching almost 100\% decoding accuracy afterwards.

VGGT does not remove positional information from tokens, nor modify the backbone PE scheme.
These results show that VGGT may learn a robust multi-view representation by implicitly inferring a canonical latent coordinate system that reconciles PEs across views.

% These findings suggest that VGGT succeeds not by discarding positional information, but by learning to down-weight or cancel positional biases that conflict with scene-level geometry.

% \subsection{Flow Probe for Motion Consistency Analysis}

% To further evaluate whether foundation vision encoders capture \textbf{geometric and temporal consistency}, we introduce a lightweight \emph{Flow Probe} that estimates dense optical flow directly from frozen backbone features.  
% Unlike task-specific fine-tuning, this diagnostic probe is designed to isolate the representational capability of the pretrained features with minimal additional learning.

% \begin{figure*}
%     \centering
%     \begin{tabular}{cccccc}
%         \includegraphics[width=0.15\linewidth,trim={15.8cm 0 0cm 0cm},clip]{misc/optical_flow/flow_val_dino-vitb16_002_000.png}  & 
%         \includegraphics[width=0.15\linewidth,trim={15.8cm 0 0cm 0cm},clip]{misc/optical_flow/flow_val_dinov2-large_002_000.png}  & 
%         \includegraphics[width=0.15\linewidth,trim={15.8cm 0 0cm 0cm},clip]{misc/optical_flow/flow_val_deit-base-patch16-224_002_000.png}  & 
%         \includegraphics[width=0.15\linewidth,trim={15.8cm 0 0cm 0cm},clip]{misc/optical_flow/flow_val_I-JEPA_vitg16_22k_002_000.png}  & 
%         \includegraphics[width=0.15\linewidth,trim={15.8cm 0 0cm 0cm},clip]{misc/optical_flow/flow_val_metaclip-b16-fullcc2.5b_002_000.png}  & 
        
%         \\
%         & 
%     \end{tabular}
    
%     \caption{Caption}
%     \label{fig:placeholder}
% \end{figure*}

\section{Discussion}

This work identifies a previously unarticulated failure mode in ViTs, in which spatial anchoring degrades under PE mismatch.
Unlike prior analyses~\cite{el2024probing,chen2025feat2gs} that examine ViT representations as inseparable tokens, our study isolates how much of a model’s spatial behavior originates from visual content versus the positional mechanism itself.

% \paragraph{PEs as geometric regulators.}
% Across a diverse set of foundation ViTs, we observe that PEs are not only coordinate tags, but also act as an implicit spatial kernel that anchors local geometry. By modulating attention as a function of spatial offset~\Cref{eq:att_decomp,eq:att_rot_decomp}, PEs act as the primary control mechanism for geometric organization in ViTs by determining token interactions.
% This perspective unifies prior observations on locality, correspondence, and robustness, and clarifies why changes in positional design can have disproportionate effects on geometric behavior.

\paragraph{Under Unified Lens of Positional Embeddings.}
Several prior works~\cite{sun2021loftr, li2021revisiting} have reported that PEs can degrade geometric alignment, correspondence accuracy, or cross-view consistency in ViTs.
By holding visual content fixed and selectively intervening on PEs, our experiments isolate the causal role of PEs in shaping geometric representations, resulting in a more nuanced and unified interpretation.
% While PEs stabilize spatial structure within a single view, they also introduce a critical limitation for multi-view geometry, as also confirmed from previous works~\cite{sun2021loftr, li2021revisiting}.

Under which these previously reported failure cases can be understood as consequences of misaligned PE-induced spatial kernels, rather than as evidence that PEs themselves are harmful.
When PEs are defined with respect to image-space coordinate systems, tokens extracted from different views or crops may be embedded in incompatible positional reference frames, even when they correspond to identical scene content.
As stated in~\cref{sec:reindexing}, a simple training-free re-indexing can significantly recover a large fraction of the lost geometric correspondence.
The success of VGGT provides further confirms this insight.

Overall, our analysis provides a unified view of prior PE failures, that multi-view geometric degradation arises not from the presence of PEs or a lack of geometric information, but from the inconsistent PEs across views.
% This perspective reconciles previous observations in the literature and highlights the importance of PEs as a key factor for reliable multi-view geometry in ViTs.

\textbf{Design Implications.} Our analysis does not propose a new positional encoding scheme.
Instead, it suggests that architectural gains may arise from how positional information is represented, aligned, or transformed, rather than always focusing on strengthening visual feature extraction.
In particular, our experimental results demonstrate that a significant portion of geometric inconsistency originates from inconsistent PEs. This suggests that architectural improvements should prioritize PE consistency for multi-view tasks.

\paragraph{Limitations and Future Work.}
Our analysis isolates the representational role of PEs at inference time.
This probing setup reveals how PEs structure spatial relations in pretrained ViTs, but it does not capture how a model would behave if positional mechanisms were changed during training. 
Thus, our conclusions speak to the inherent dependencies learned under standard PE schemes rather than reflecting how a model would adapt if retrained under altered positional encoding schemes entirely.
Inspired by recent works such as~\cite{li2025cameras,gopalakrishnan2025decoupling}, an interesting next step is to move beyond purely image-space positional frames.
Our token re-indexing and VGGT probing experiment suggest that part of the failures of multi-view representation learning arises from PE misalignment rather than representation collapse.
This points toward future positional designs that incorporate geometric constraints, such as multi-view alignment or scene-aware PE systems, to better bridge image-space indexing with scene-space structure.

\section{Conclusion}
We presented a set of controlled experiments examining how positional embeddings (PEs) influence several spatial properties of vision transformer (ViT) representations. 
Across 14 foundation ViTs, we showed that PEs critically regulate spatial interactions and geometric consistency for multi-view representations.
We hope these findings provide a clearer empirical basis for future work on understanding and designing positional mechanisms in ViTs and multi-view geometry.

\clearpage
\newpage

\section*{Impact Statement}

This work provides a fundamental analysis of how positional embeddings shape spatial reasoning in vision transformers, particularly in multi-view settings. By isolating positional mechanisms from visual content, our findings clarify why existing models may succeed or fail under viewpoint changes, and provide diagnostic tools for analyzing spatial consistency at the representation level.

Overall, we view this work as contributing foundational insights into representation learning, with the primary impact that provides a basis for more principled analysis and design of positional mechanisms for future vision models.

\bibliography{main}
\bibliographystyle{icml2026}

%%%%%%%%%%%%%%%%%%%%%%%%%%%%%%%%%%%%%%%%%%%%%%%%%%%%%%%%%%%%%%%%%%%%%%%%%%%%%%%
%%%%%%%%%%%%%%%%%%%%%%%%%%%%%%%%%%%%%%%%%%%%%%%%%%%%%%%%%%%%%%%%%%%%%%%%%%%%%%%
% APPENDIX
%%%%%%%%%%%%%%%%%%%%%%%%%%%%%%%%%%%%%%%%%%%%%%%%%%%%%%%%%%%%%%%%%%%%%%%%%%%%%%%
%%%%%%%%%%%%%%%%%%%%%%%%%%%%%%%%%%%%%%%%%%%%%%%%%%%%%%%%%%%%%%%%%%%%%%%%%%%%%%%
\newpage
\appendix
% \onecolumn

\clearpage
\setcounter{page}{1}
\maketitlesupplementary
\appendix

\section{Technical Details}

This section presents technical details regarding how we perform positional weighting, the derivation of relative and rotary positional kernels, and how do we construct the 4D correlation volume and relevant metrics to capture epipolar peak response.

\subsection{Positional Weighting}
\label{sec:pos_weighting}

We define a scalar coefficient $\omega \in [0,1]$, referred to as the \emph{positional weight}, that modulates the strength of the positional signal. We adapt the formulation to each type of positional encoding used in modern ViTs.

\subsubsection{Absolute and Relative PEs}
\paragraph{Absolute PEs (ViT, DeiT, CLIP).}
Absolute PEs are directly scaled by the weighting factor:
\begin{equation}
    \mathbf{P}' = \omega \cdot \mathbf{P},
\end{equation}
where $\mathbf{P}$ denotes the pretrained absolute PE matrix.

\paragraph{Relative PEs (Swin, SwinV2, BEiT, Data2Vec).}
Similar to absolute PEs, for models employing relative position bias or continuous positional MLPs, we apply a multiplicative scaling:
\begin{equation}
    \mathbf{B}' = \omega \cdot \mathbf{B},
\end{equation}
where $\mathbf{B}$ represents either the discrete bias table or the learned continuous bias embedding.

\subsubsection{Rotary PEs}
RoPE define a rotation operator
$\mathbf{R}_\theta = 
\begin{pmatrix}
\cos\theta & -\sin\theta\\[2pt]
\sin\theta & \cos\theta
\end{pmatrix}$.
Crucially,  RoPE is {multiplicative} in the sense of a rotation; there is no additive positional vector to “scale” as in absolute and relative PEs. Therefore, weighting the positional effect must operate on the {rotation} itself, not by multiplying with a scalar.

Differently, RoPE encode positions by \emph{rotating} each 2D subspace of the query/key vectors by a phase that depends on the token index~$p$ and the subspace frequency~$\omega_i$.
For a 2D slice $(x_{2i},x_{2i+1})$ of a vector $x\in\mathbb{R}^d$ (with $d$ even), the standard RoPE transform is $R(\theta_i)\!\begin{bmatrix}x_{2i}\\ x_{2i+1}\end{bmatrix}$, where $\theta_i \;=\; p\,\omega_i$ and $\omega_i \;=\; 10000^{-2i/d}.$
The conceptually clean approach is to scale the {phase}:
\begin{equation}
\label{eq:phase_scaling}
R_w(\theta_i) \;:=\; R(w\,\theta_i)
\;=\;
\begin{bmatrix}
\cos(w\theta_i) & -\sin(w\theta_i)\\
\sin(w\theta_i) & \cos(w\theta_i)
\end{bmatrix}.
\end{equation}
Such that $R_0(\theta_i)=I$ and $R_1(\theta_i)=R(\theta_i)$. For fixed $p$, decreasing $w$ reduces the {angular rate} across positions, thereby softening position sensitivity without altering vector norms. Therefore, the attention kernel induced by RoPE (through $q^\top k$ after rotation) varies smoothly with $w$, preserving relative geometry while controlling positional contrast.

However, direct updating $\theta$ is not supported in many libraries. Thus, to ease the implementation, we interpolate between the identity (no rotation) and full rotation as:
\begin{equation}
    \cos' = 1 - \omega + \omega \cos\theta, 
    \qquad
    \sin' = \omega \sin\theta.
\end{equation}
This formulation preserves orthogonality while continuously modulating
the angular contribution of positional rotation.

However, as many libraries cache only $(\cos\theta,\sin\theta)$, not $\theta$ itself, we use an effective surrogate is to interpolate the rotation toward the identity:
\begin{equation}
\label{eq:rope_interpolation}
\tilde R_w \;\approx\; (1-w)\,I + w\,R(\theta),
\end{equation}
such that:
\begin{equation}
\begin{cases}
\cos'(p,i) \;=\; (1-w)\cdot 1 + w\cdot \cos\theta_i \;=\; 1-w + w\cos\theta_i,\\[2pt]
\sin'(p,i) \;=\; (1-w)\cdot 0 + w\cdot \sin\theta_i \;=\; w\sin\theta_i.
\end{cases}
\end{equation}
While \eqref{eq:rope_interpolation} is not equal to $R(w\theta)$ for intermediate $w$, it is a stable and faithful {blend} between identity and the full rotation.

\begin{figure}[t]
\centering
\begin{tikzpicture}
\begin{axis}[
    width=0.88\linewidth,      % larger figure
    height=0.48\linewidth,
    xlabel={$\theta$},
    ylabel={value},
    xmin=0, xmax=3.2,
    ymin=-1.1, ymax=1.1,
    xtick={0,1.57,3.14},
    xticklabels={$0$, $\pi/2$, $\pi$},
    ytick={-1, -0.5, 0, 0.5, 1},
    grid=both,
    legend style={
        font=\scriptsize,          % smaller legend font
        at={(0.5,-0.85)},
        anchor=south,
        legend columns=2,          % <-- 1 line legend
        column sep=8pt,
    },
]

% true phase-scaled rotation
\addplot[blue, thick] 
    {cos(deg(0.4*x))};
\addlegendentry{$\cos(w\theta)$ (true)}

% reference cos(theta)
\addplot[gray, thin, dotted] 
    {cos(deg(x))};
\addlegendentry{$\cos(\theta)$ (reference)}

% interpolated surrogate
\addplot[orange, thick, dashed] 
    {1-0.4 + 0.4*cos(deg(x))};
\addlegendentry{$1-w + w\cos\theta$ (interp)}

% ---- MANUAL LEGEND (NO & , NO MATRIX, NO ERROR) ----
\node[anchor=north west, font=\scriptsize, text depth=0pt]
    at (rel axis cs:0.12,-0.18)
    {\tikz{\draw[blue,thick] (0,0) -- (6mm,0);} \(\cos(w\theta)\) (true)};

\node[anchor=north west, font=\scriptsize, text depth=0pt]
    at (rel axis cs:0.55,-0.18)
    {\tikz{\draw[gray,thin,dotted] (0,0) -- (6mm,0);} \(\cos(\theta)\) (reference)};

\node[anchor=north west, font=\scriptsize, text depth=0pt]
    at (rel axis cs:0.25,-0.32)
    {\tikz{\draw[orange,thick,dashed] (0,0) -- (6mm,0);} \(1-w + w\cos\theta\) (interpolated)};

\end{axis}
\end{tikzpicture}

\caption{\textbf{Comparison of phase-scaling vs interpolation for RoPE.}
The interpolated rotation $(1-w)I + wR(\theta)$ closely follows the true phase-scaled rotation $R(w\theta)$ in magnitude, validating it as an effective implementation surrogate.}
\label{fig:rope_interpolation}
\end{figure}
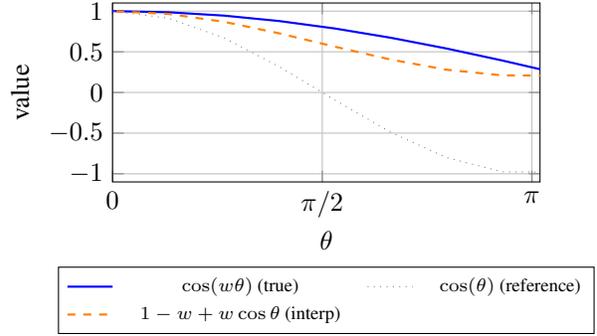

% \begin{figure}[t]
% \centering
% \begin{tikzpicture}

% \begin{axis}[
%     width=0.48\linewidth,
%     height=0.48\linewidth,
%     xlabel={$\theta$},
%     ylabel={value},
%     xmin=0, xmax=3.2,
%     ymin=-1.1, ymax=1.1,
%     xtick={0,1.57,3.14},
%     xticklabels={$0$, $\pi/2$, $\pi$},
%     ytick={-1, -0.5, 0, 0.5, 1},
%     grid=both,
%     clip=false,
%     legend style={draw=none},
% ]

% % ---- PLOTS ----
% \addplot[blue, thick] {cos(deg(0.4*x))};
% \addlegendentry{true}

% \addplot[gray, thin, dotted] {cos(deg(x))};
% \addlegendentry{ref}

% \addplot[orange, thick, dashed] {1-0.4 + 0.4*cos(deg(x))};
% \addlegendentry{interp}

% % ---- MANUAL LEGEND (NO & , NO MATRIX, NO ERROR) ----
% \node[anchor=north west, font=\scriptsize, text depth=0pt]
%     at (rel axis cs:0.12,-0.18)
%     {\tikz{\draw[blue,thick] (0,0) -- (6mm,0);} \(\cos(w\theta)\) (true)};

% \node[anchor=north west, font=\scriptsize, text depth=0pt]
%     at (rel axis cs:0.55,-0.18)
%     {\tikz{\draw[gray,thin,dotted] (0,0) -- (6mm,0);} \(\cos(\theta)\) (reference)};

% \node[anchor=north west, font=\scriptsize, text depth=0pt]
%     at (rel axis cs:0.25,-0.32)
%     {\tikz{\draw[orange,thick,dashed] (0,0) -- (6mm,0);} \(1-w + w\cos\theta\) (interpolated)};

% \end{axis}
% \end{tikzpicture}

% \caption{\textbf{Comparison of phase-scaling vs interpolation for RoPE.}
% The interpolated rotation $(1-w)I + wR(\theta)$ closely follows the true rotation $R(w\theta)$.}
% \label{fig:rope_interpolation}
% \end{figure}

\subsection{Epipolar Peak Response Capturing}

\subsubsection{4D Correlation Volume}

To measure how well a model preserves epipolar geometry, we construct a full \emph{4D correlation volume} between the left and right feature maps. This formulation is conceptually related to correlation layers in optical flow~\cite{dosovitskiy2015flownet,sun2018pwc,teed2020raft} that retains the entire 2D search space.

Given normalized feature embeddings 
$\hat{\mathbf{F}}_L, \hat{\mathbf{F}}_R \in \mathbb{R}^{B \times C \times H \times W}$,
the 4D correlation is defined as:
\begin{equation}
    \mathbf{C}(x,y,x',y')
    = \hat{\mathbf{F}}_L(x,y)^\top \hat{\mathbf{F}}_R(x',y'),
    \label{eq:4d_corr}
\end{equation}
yielding a dense affinity tensor
$\mathbf{C} \in \mathbb{R}^{B \times H \times W \times H \times W}$.
This volume encodes all potential cross-view correspondences without imposing rectification or epipolar-line constraints.

To obtain a differentiable point estimate of correspondence, we apply a spatial softmax over the search domain:
\begin{equation}
p(x',y'\mid x,y)
    = \frac{\exp(\tau\,\mathbf{C}(x,y,x',y'))}
           {\sum_{x'',y''} \exp(\tau\,\mathbf{C}(x,y,x'',y''))},
\end{equation}
where $\tau$ controls the sharpness.
The expected displacement is then:
\begin{equation}
    \hat{\mathbf{u}}(x,y) 
    = \mathbb{E}_{p(x',y'\mid x,y)}[(x'-x,~y'-y)].
\end{equation}
This “soft argmax’’ provides a differentiable approximation to hard matching and is widely used in deep stereo and multi-view reconstruction~\cite{im2019dpsnet,li2021revisiting,zhe2023geomvsnet,liu2023epipolar}.

\paragraph{Rectified Stereo as a Special Case.}
Similar to the traditional stereo cost volumes~\cite{kendall2017end,chang2018pyramid}, rectified stereo geometry restricts the \emph{true} matches to lie on the horizontal epipolar line $y' = y$. Under this geometric constraint, the relevant slice of the 4D volume reduces to a standard 1D cost volume:
\begin{equation}
    C(x,y,d)
    = \mathbf{C}(x,y,\,x+d,\,y),
    \label{eq:1d_corr}
\end{equation}
where $d$ is the horizontal displacement.
The disparity estimate is then obtained through the familiar soft-argmin operation:
\begin{equation}
    \hat{d}(x,y)
    = \sum_{d} d \cdot \text{softmax}\!\left(\tau\, C(x,y,d)\right).
\end{equation}
Thus, the traditional 1D stereo formulation emerges naturally as the \emph{rectified geometric slice} of the full 4D correspondence volume, rather than a separate construction.

\subsubsection{Metrics}

We evaluate how faithfully the feature-induced cost volume reflects the true epipolar geometry.

\paragraph{End-Point Error (EPE).}
EPE measures the average absolute deviation between the predicted and ground-truth disparities:
\begin{equation}
    \text{EPE}
    = \frac{1}{|\Omega|}
      \sum_{(x,y)\in\Omega} 
      \big|\hat{d}(x,y) - d^{gt}(x,y)\big|.
\end{equation}
Lower values indicate that the peak of the similarity distribution aligns closely with the correct epipolar correspondence.

\paragraph{Epipolar Recall@n.}
To quantify how sharply the correlation volume peaks around the correct match, we compute Recall@n. The percentage of pixels for which ground truth disparity lies among the top-$n$ highest responses in $C(x,y,d)$.  
A higher Recall@n indicates a more confident and better localized epipolar peak.

\section{More Experiments}

\begin{figure}[b]
\scalebox{.7}{%
\begin{tikzpicture}[
    font=\sffamily,
    gridstyle/.style={draw=gray!50, very thin},
    feature_node/.style={fill=blue!20, text=blue!80!black, font=\bfseries},
    patch_node/.style={fill=green!20, text=green!60!black, font=\bfseries},
    offset_patch_node/.style={fill=red!20, text=red!80!black, font=\bfseries},
    arrowstyle/.style={-latex, thick, shorten >=1pt},
    labelstyle/.style={font=\small\itshape}
]

% Define grid dimensions and spacing
\def\gridsize{5}
\def\patchsize{1}
\def\mapsize{0.6}
\def\spacing{6}

% Draw Feature Map on the left
\node[label={[font=\bfseries,xshift=2.5cm, yshift=-.8cm]above:Feature Map ($F$)}] (fm_anchor) at (0,0) {};
\draw[gridstyle] (fm_anchor) grid ++(\gridsize*\patchsize, \gridsize*\patchsize);

% Draw Image Patches on the right
\node[label={[font=\bfseries,xshift=2.5cm,yshift=-.8cm]above:Reconstructed Feature Map ($F'$)}] (img_anchor) at (\spacing, 0) {};
\draw[gridstyle] (img_anchor) grid ++(\gridsize*\patchsize, \gridsize*\patchsize);

% --- Define coordinates for the feature and patches ---
% Using (i,j) notation where (0,0) is the bottom-left corner
\def\i{2} % x-coordinate
\def\j{2} % y-coordinate
\def\k{1} % offset value

% Feature cell f_{i,j}
\coordinate (f_ij) at ({\i*\mapsize + \mapsize/2}, {\j*\mapsize + \mapsize/2});
\fill[blue!20] (\i*\patchsize - \patchsize, \j*\patchsize - \patchsize) rectangle ++(\patchsize, \patchsize);
\node[feature_node, minimum size=\mapsize*10pt, inner sep=0pt] at (f_ij) {$f_{i,j}$};

% Local patch p_{i,j}
% \coordinate (p_ij) at (\spacing + \i*\patchsize + \patchsize/2, {\j*\patchsize + \patchsize/2});
% \node[patch_node, minimum size=\patchsize*10pt, inner sep=0pt] at (p_ij) {$p_{i,j}$};

% Gray out the (i,j) block in the right grid
\fill[gray!30] (\spacing + \i*\patchsize - \patchsize, \j*\patchsize - \patchsize) rectangle ++(\patchsize, \patchsize);
\node at (\spacing + \i*\mapsize + \mapsize/2, \j*\mapsize + \mapsize/2) {};

% Offset patch p_{i+k, j+k}
\pgfmathsetmacro{\ik}{\i+\k}
\pgfmathsetmacro{\jk}{\j+\k}

\fill[red!20] (\spacing + \ik*\patchsize, \jk*\patchsize) rectangle ++(\patchsize, \patchsize);
\coordinate (p_ik_jk) at (\spacing + \ik*\patchsize + \patchsize/2, {\jk*\patchsize + \patchsize/2});
\node[offset_patch_node, minimum size=\patchsize*10pt, inner sep=0pt] at (p_ik_jk) {$\hat{p}_{i+k, j+l}$};

% --- Draw arrows and labels ---
% Arrow for local reconstruction (k=0)
% \draw[arrowstyle, blue!80] (f_ij) to[bend left=10]
%     node[midway, above, sloped, labelstyle, text=blue!80!black] {Local Reconstruction ($k=0$)} (p_ij);

% Arrow for offset reconstruction (k>0)
\draw[arrowstyle, red!80] (f_ij) to[bend right=15]
    node[midway, below, sloped, labelstyle, text=red!80!black] {\small Offset Reconstruction ($k\neq0$)} (p_ik_jk);

\end{tikzpicture}
}
\caption{
% Visual demonstration of token-offset probing. We restore the model features to 2D feature map representation. The feature vector $f_{i,j}$ from the source map (left) is used to reconstruct a feature vector $\hat{p}$ at a spatially offset location $(i+k, j+l)$ in the target grid (right). The original location $(i, j)$ in the target grid is grayed out to illustrate it is not the reconstruction target.
\textbf{Visualization of token-offset probing}.
A source token $f_{i,j}$ from $F$ (left) is used to reconstruct $\hat{p}_{i+k,\,j+l}$ at an offset position in $F'$ (right), with the original $(i,j)$ location grayed out.
Note that flattened transformer tokens lack 2D spatial relations. We reshape them into a feature map layout for interpretability.
}
\label{fig:feature_offset}
\end{figure}
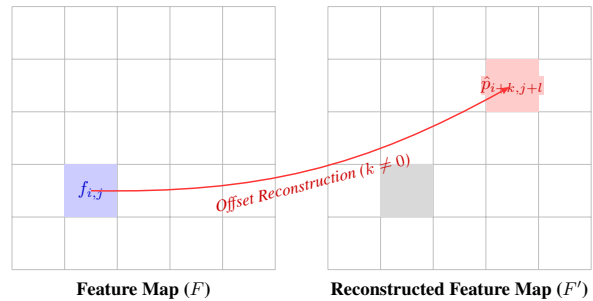

\input{figures/offset_res}

\subsection{PE Strength Sweep}
\label{sec:pe_token_correlation}

Unlike convolutional networks, which enforce spatial continuity through local receptive fields, ViTs operate on sets of tokens with no inherent notion of spatial adjacency.
In principle, attention is permutation-invariant, meaning that any token could interact with any other, independent of their image positions.
Yet, in practice, ViT features exhibit surprisingly smooth spatial organization where neighboring patches often encode correlated information and preserve local geometry.
This raises a fundamental, unanswered question:
\begin{quote}
\textit{Where does this spatial continuity come from in a model with no built-in notion of space?}
\end{quote}

\subsubsection{Method}
As discussed in~\cref{sec:pe_kernel}, we hypothesize that PEs are responsible for inducing an \textit{implicit spatial kernel} that governs how attention strength decays with distance.
To test this, we design a token-offset probing experiment that directly measures how token similarity varies with spatial displacement and the weight of PEs.
If PEs truly modulate inter-token interactions, reducing their weight should weaken similarity, while a coherent spatial representation should exhibit decreasing similarity with increasing distance.
By observing the shape and steepness of this decay, we can then understand how positional signals translate into spatial inductive bias.

\subsubsection{Setup}
Given a reference token $\mathcal{T}_{i,j}$ at spatial position $(i,j)$, we examine its relationship with a displaced token $\mathcal{T}_{i+\Delta x,,j+\Delta y}$ located at an offset $(\Delta x, \Delta y)$ within the same feature map, as illustrated in~\cref{fig:feature_offset}.
We evaluate the \emph{cosine similarity} between tokens $\mathcal{T}_{i,j}$ and $\mathcal{T}_{i+\Delta x,j+\Delta y}$ to understand the local geometric continuity.
We report the average similarity over all valid displacement pairs.
For simplicity, we consider symmetric integer offsets $k \in \{1, 2, 3\}$ with $\Delta x = \Delta y = k$, and apply a decaying PE strength $\omega \in \{0.0, 0.2, 0.4, 0.6, 0.8, 1.0\}$.

We carry out our experiments based on \textit{Imagenette}\cite{Howard_Imagenette_2019} with a size of $224\times224$.
Most transformer-based foundation vision encoders use a patch size of $16$ or $14$, resulting in a fixed grid of $14 \times 14$ or $16 \times 16$. The feature grids are downsampled to $16 \times 16$ for the models that require a larger input size. 

% We design three families of probes to answer the question:
% \emph{To what degree do positional embeddings govern the representation of local geometry within tokens?}

\subsubsection{Analyses}

\Cref{fig:pos_weight_multi_offset} illustrates the token-offset experiment results across 14 visual foundation models.

% \paragraph{Local continuity manifests as token similarity decay.}
\paragraph{Positional embeddings go beyond providing coordinates.}
Apart from BEiT, larger offsets consistently weaken the similarity, reflecting a notion of local continuity. This behavior indicates that positional components contribute explicitly to spatial discrimination, which aligns with~\cref{sec:pe_kernel}.
With decayed PE strength, the model shifts from a locally structured representation to a globally homogeneous one.
Interestingly, we observe no significant differences among various PE strength and token offset settings for BEiT, indicating the lack of locality. This observation also aligns with our later experiment in~\cref{subsec:remove_pe_locality}.
Additionally, averaged over all encoders, the area under the curve (AUC) drops by ${\sim}8\%$ from lowest to highest PE strength, while the end-to-end decay (PE strength from 0 to 1) rises by ${\sim}58\%$. In this protocol, PE directly controls how concentrated (local) the token interactions are.

% \paragraph{Training objective-dependent spatial sensitivity.}
% The rate of this decay varies.
% {I-JEPA} and {DINOv2/3} show large locality gains with PE, while DeiT/SAM remain globally correlated even with strong PE strength.
% This difference aligns with each model’s training objective. 
% Self-distillation and contrastive models (\eg, DINO, DINOv2) show steep, monotonic similarity decay, reflecting a tight local structure.
% Vision–language and segmentation models (\eg, CLIP, SAM) display shallower decay, emphasizing semantic rather than spatial grouping.

% \begin{takeaway}
% \textbf{Positional embeddings go beyond providing coordinates.} They contribute substantially to the spatial organization of ViT representations.
% \end{takeaway}

\subsection{Do tokens know where they are?}

To understand if tokens are aware of its position as expected, we design a simple probing experiment to evaluate the positional awareness of various ViT-based foundation models.
Given an encoder’s feature map $\mathbf{F} \in \mathbb{R}^{C\times H\times W}$ extracted from an input image, we train a lightweight \emph{position predictor} that takes either a single feature token or a local $N\times N$ neighborhood and predicts its absolute row and column indices $(r, c)$ on the feature grid.
Both row and column positions are modeled as categorical classification tasks with cross-entropy losses.
The encoder is frozen during training, and only the probe is optimized.
The evaluation metrics include separate accuracies for row and column predictions, as well as the joint ``both'' accuracy for exact 2D localization.

\Cref{tab:positional_encoding} demonstrates that nearly all tested ViTs exhibit strong positional awareness, even though their encoding mechanisms differ widely.
Models trained with explicit absolute or relative embeddings (e.g., DINO, DEiT, SAM) achieve almost perfect localization accuracy. In general, all position encoding methods allow positional information to be linearly recovered, suggesting that positional cues are deeply embedded in the learned representation regardless of the encoding formulation.

Differently, Swin/SwinV2 rely on local windows and shifted windows. Token locations in Swin models are explicitly encoded directly, we might expect it to rely less on positional embeddings.
By design, Swin series models emphasize relative positions within windows and hierarchical structure, so that absolute 2D positions are much less linearly decodable in those models.
In addition, rotary embedding methods such as DINOv3 and MLCD exhibit also a slight weaker performance than those absolute and relative embedding method.
We believe it is because rotary encoding preserves direction (angular displacement) without encoding scales, thus it does not give the model a unique positional signature tied to a specific grid location. As a result, recovering exact row/column indices becomes harder.

\begin{table}
    \centering
    \scriptsize
    \begin{tabular}{l|ccc}
        \toprule
         & row & col & both \\
        \midrule
        DINO       & 0.999 & 0.999 & 0.999 \\
        DINOv2     & 0.926 & 0.961 & 0.893 \\
        DINOv3     & 0.796 & 0.846 & 0.687 \\
        I-JEPA     & 0.995 & 0.998 & 0.994 \\
        DEit       & 0.996 & 0.996 & 0.993 \\
        Siglip2    & 0.995 & 0.993 & 0.989 \\
        BeiT       & 0.822 & 0.813 & 0.658 \\
        dat2vec    & 0.994 & 0.988 & 0.982 \\
        MLCD       & 0.848 & 0.801 & 0.697\\
        ViT        & 0.997 & 0.997 & 0.994 \\
        SAM        & 0.979 & 0.973 & 0.956 \\
        Clip       & 0.888 & 0.874 & 0.810 \\
        Swin       & 0.803 & 0.682 & 0.561 \\
        SwinV2     & 0.602 & 0.418 & 0.270 \\
        \bottomrule
    \end{tabular}
    \caption{\textbf{Token position prediction accuracy.} 
    Values denote the probability of correctly predicting the token’s absolute row, column, and joint 2D index.}
    \label{tab:positional_encoding}
\end{table}

% \subsection{Token Similarity Decay Visualization}

% \begin{figure*}
%     \centering
%     \setlength{\tabcolsep}{0pt}
%     \input{figures/kernel_single}
%     \caption{.}
%     \label{fig:epipolar_vis_all}
% \end{figure*}

% \begin{figure*}
%     \centering
%     \includegraphics[width=1.\linewidth]{misc/world_grid.drawio.pdf}
%     \caption{Caption}
%     \label{fig:re_index_monkaa}
% \end{figure*}

\subsection{Do Tokens Preserve Spatial Identity?}

Local continuity alone does not reveal whether the model preserves spatial identity.
Thus, we further investigate:
\begin{quote}
\textit{Without PEs, are tokens at different positions distinguishable in feature space?}
\end{quote}
Following the setup in~\cref{sec:pe_token_correlation}, we further measure whether a displaced token $\mathcal{T}{i+\Delta x, j+\Delta y}$ can be linearly reconstructed from its reference $\mathcal{T}{i,j}$ using a single linear layer.

\paragraph{Results.}
As shown in~\cref{tab:tokenoffset}, when PEs are removed, reconstruction similarity remains uniformly high across all offsets, indicating that features at different spatial locations become nearly interchangeable.
This suggests that without positional cues, ViTs drift toward a globally invariant representation space, effectively removing positional identity.
In contrast, with PEs enabled, reconstruction similarity decreases with increasing offset for almost all models.
This indicates that tokens encode spatially localized information that cannot be linearly inferred from neighboring positions.
PEs therefore preserve spatial distinctiveness, not just local continuity.
% These results confirm that positional embeddings act as a structural regulator of locality: they preserve position-dependent variation while preventing the network from collapsing into purely global representations.

Notably, models such as DINO1/2, MLCD, and Swin1/2 still retain modest positional separation even without PEs.
This is likely due to architectural inductive biases, such as local windows in Swin, provide weak additional spatial anchoring.
This aligns with the epipolar peak responses observed in the same models in~\Cref{fig:epipolar_plot}, reinforcing that certain inductive biases can sustain limited locality when positional information is removed.

\subsection{PE-induced Kernel Visualization}

We present visualization of PE-induced kernels here, as per the decomposition in~\cref{eq:att_decomp}.

\begin{figure}[h]
     \centering
     \begin{subfigure}[b]{.48\linewidth}
         \centering
         \includegraphics[width=.48\linewidth,trim={6.3cm 5.4cm 2.6cm 1cm},clip]{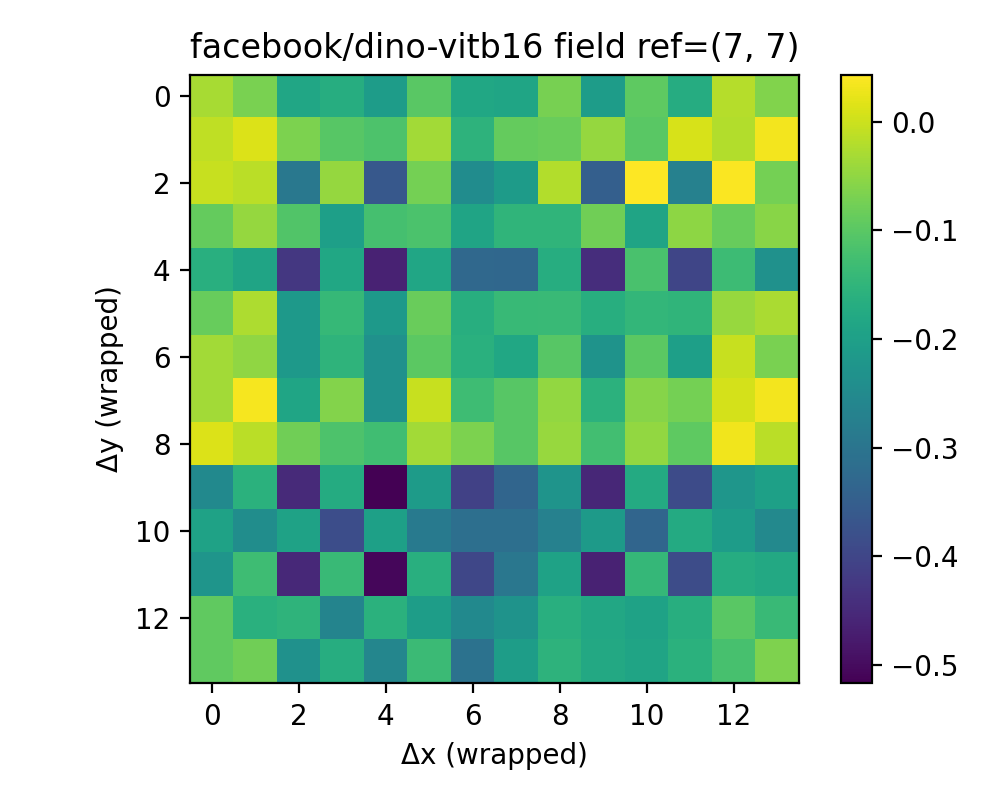}
         \includegraphics[width=.48\linewidth,trim={5.95cm 5.4cm 2.95cm 1cm},clip]{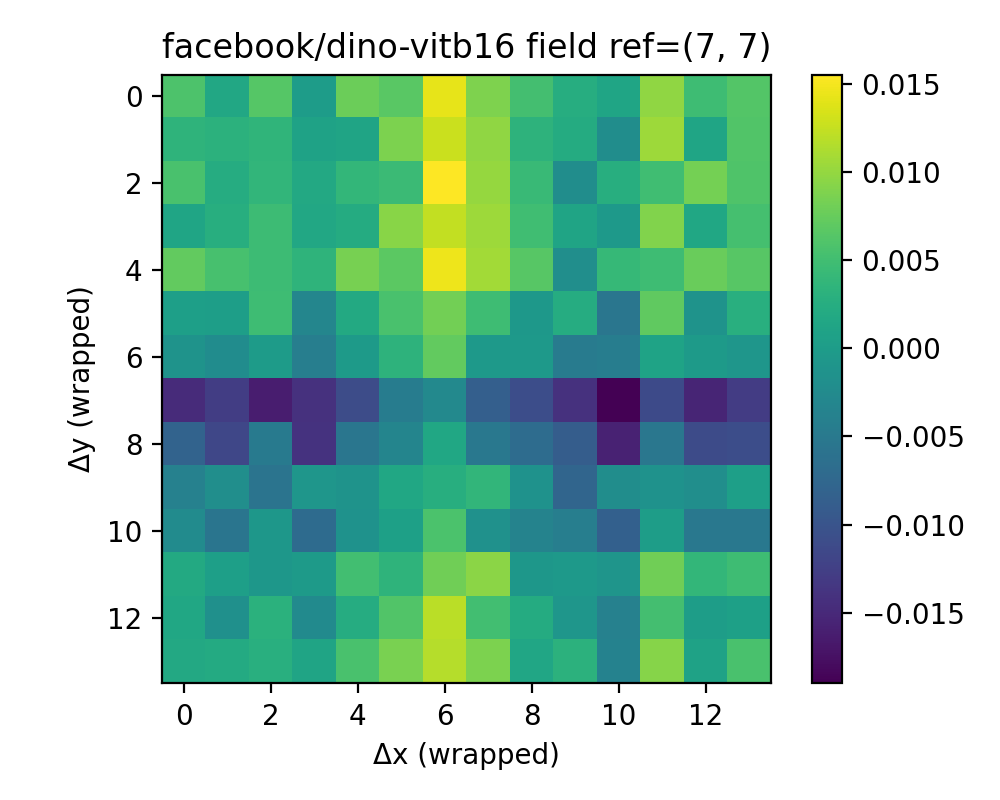}
         \caption{DINO}
     \end{subfigure}
     \hfill
     \begin{subfigure}[b]{.48\linewidth}
         \centering
         \includegraphics[width=.48\linewidth,trim={6.25cm 5.4cm 2.65cm 1cm},clip]{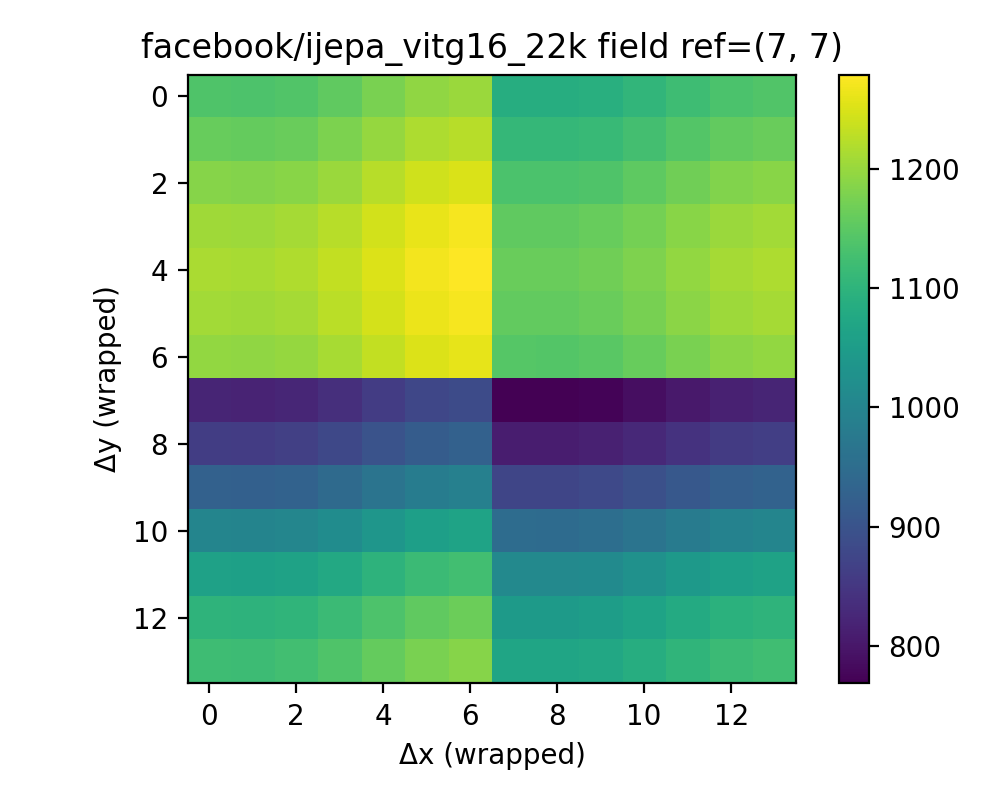}
         \includegraphics[width=.48\linewidth,trim={6.3cm 5.4cm 2.6cm 1cm},clip]{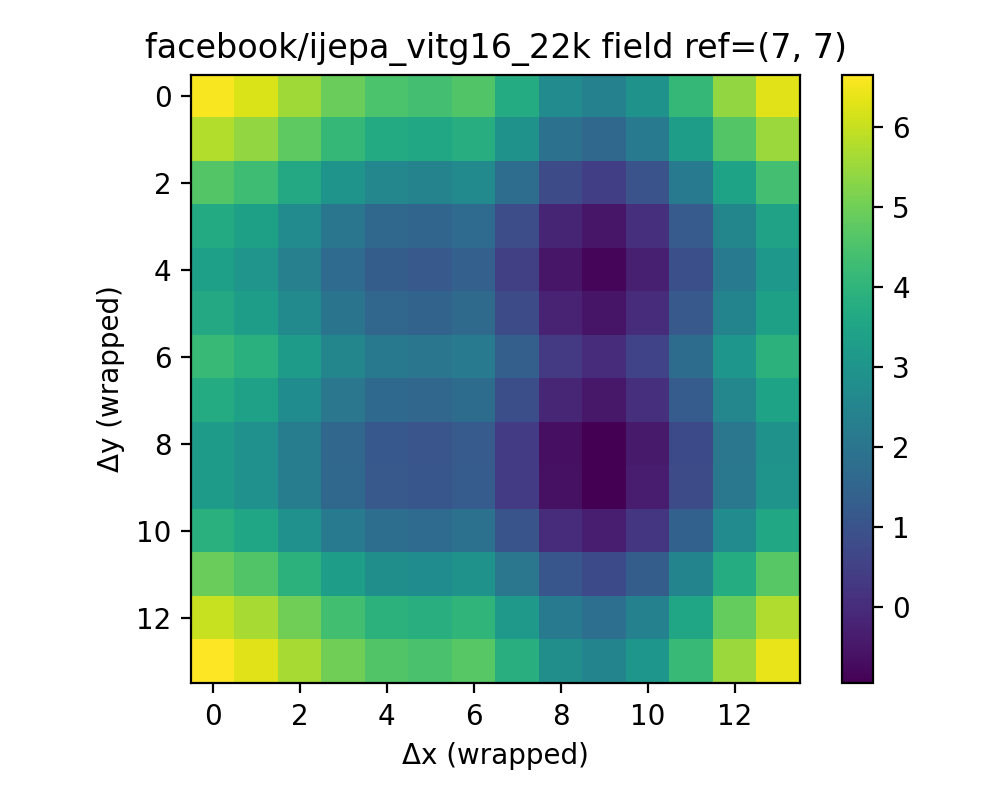}
         \caption{I-JEPA}
     \end{subfigure}
     \hfill
     \begin{subfigure}[b]{.48\linewidth}
         \centering
         \includegraphics[width=.48\linewidth,trim={6.23cm 5.46cm 2.95cm .97cm},clip]{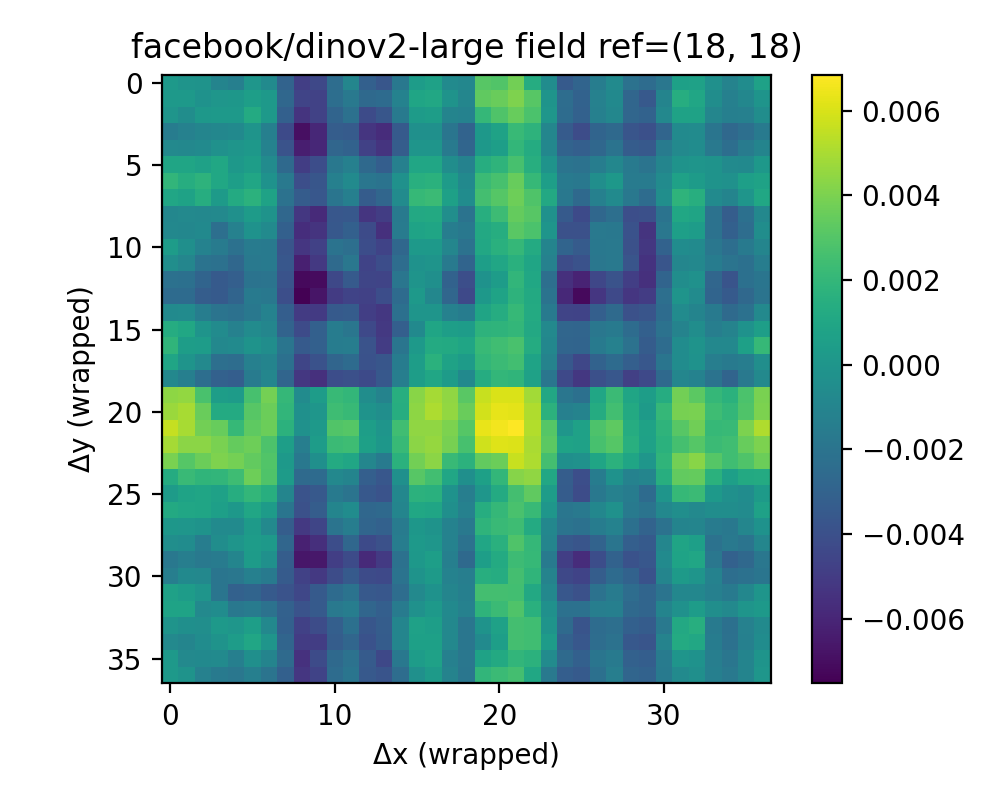}
         \includegraphics[width=.48\linewidth,trim={6.23cm 5.46cm 2.95cm .97cm},clip]{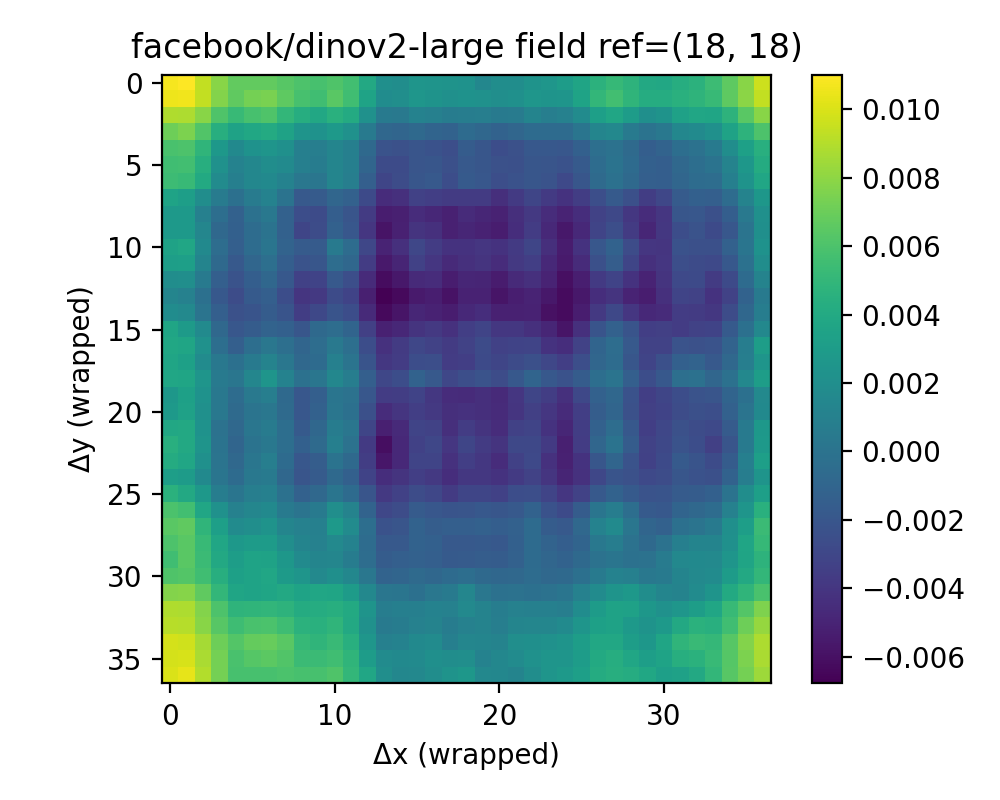}
         \caption{DINOv2}
     \end{subfigure}
     \hfill
     \begin{subfigure}[b]{.48\linewidth}
         \centering
         \includegraphics[width=.48\linewidth,trim={6.78cm 5.85cm 2.58cm .99cm},clip]{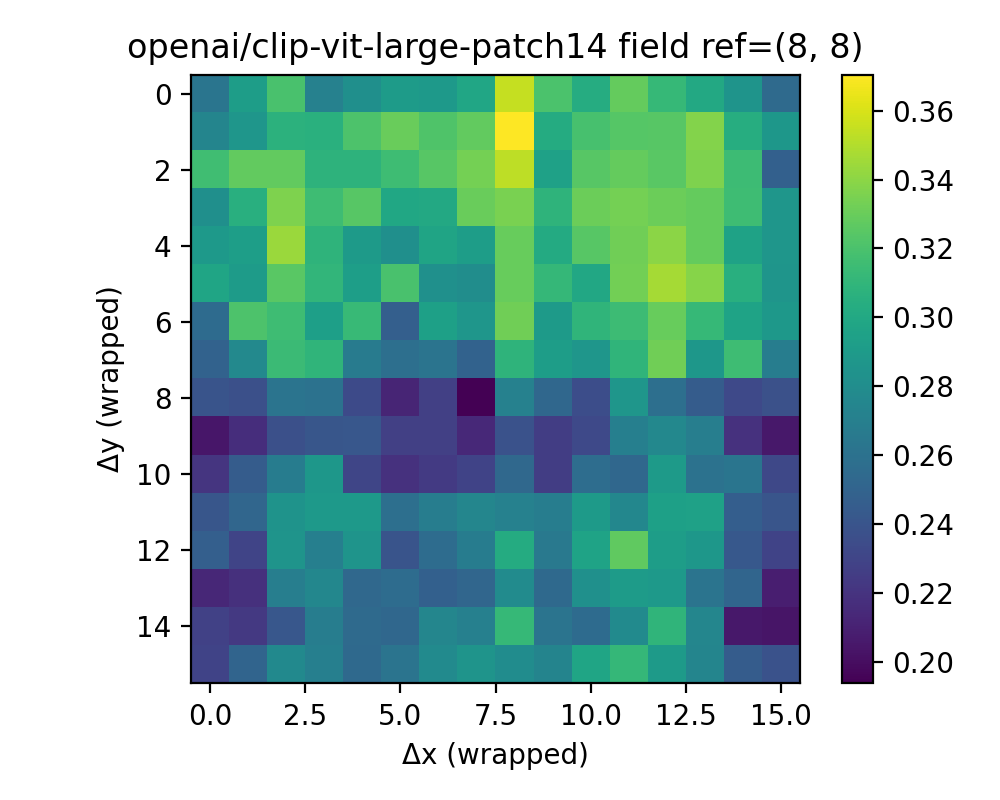}
         \includegraphics[width=.48\linewidth,trim={6.67cm 5.85cm 2.7cm .99cm},clip]{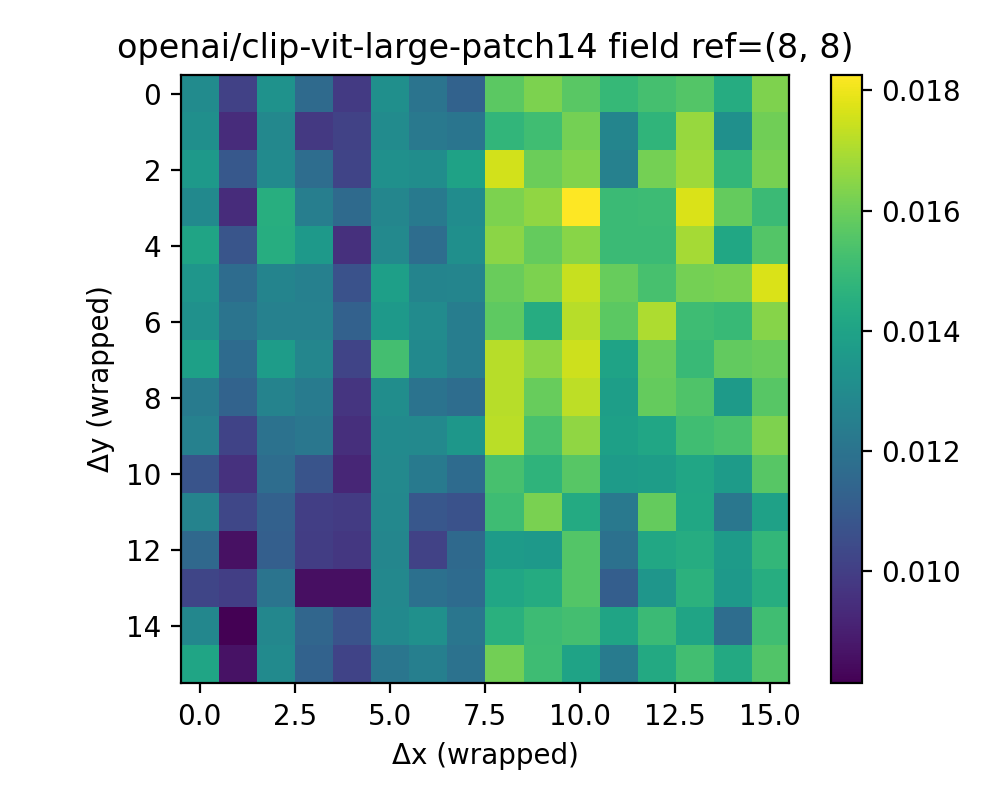}
         \caption{CLIP}
     \end{subfigure}
     \hfill
     \begin{subfigure}[b]{.48\linewidth}
         \centering
         \includegraphics[width=.48\linewidth,trim={2.45cm 1.6cm 2.8cm .99cm},clip]{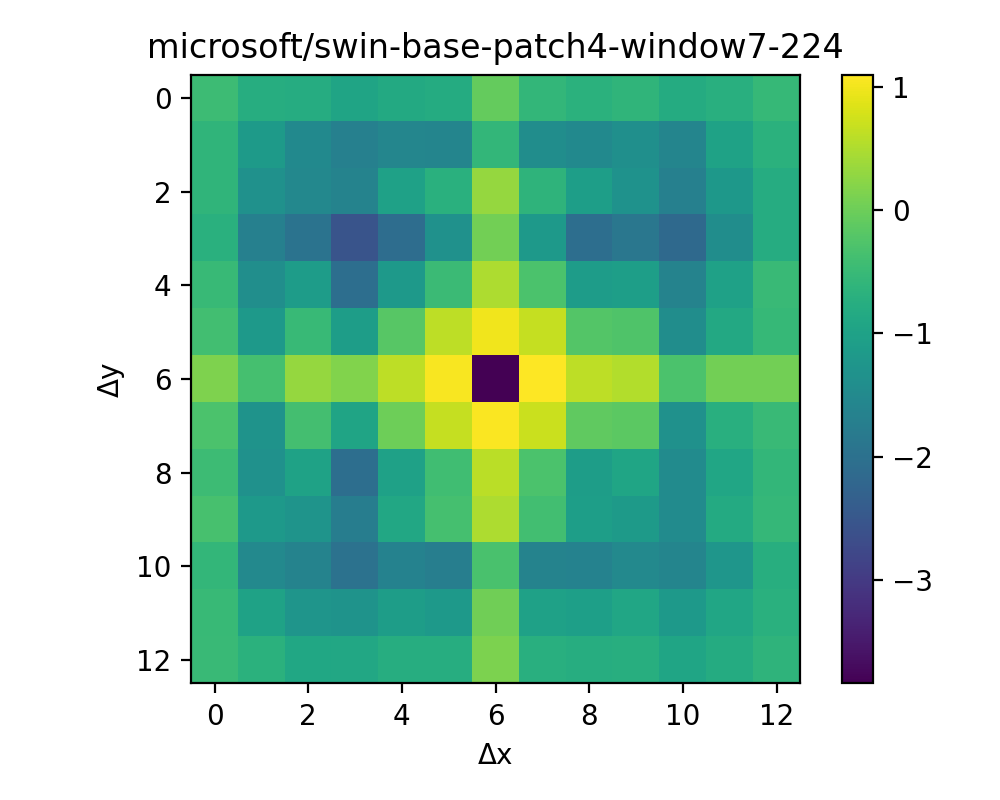}
         \includegraphics[width=.48\linewidth,trim={2.45cm 1.6cm 2.8cm .99cm},clip]{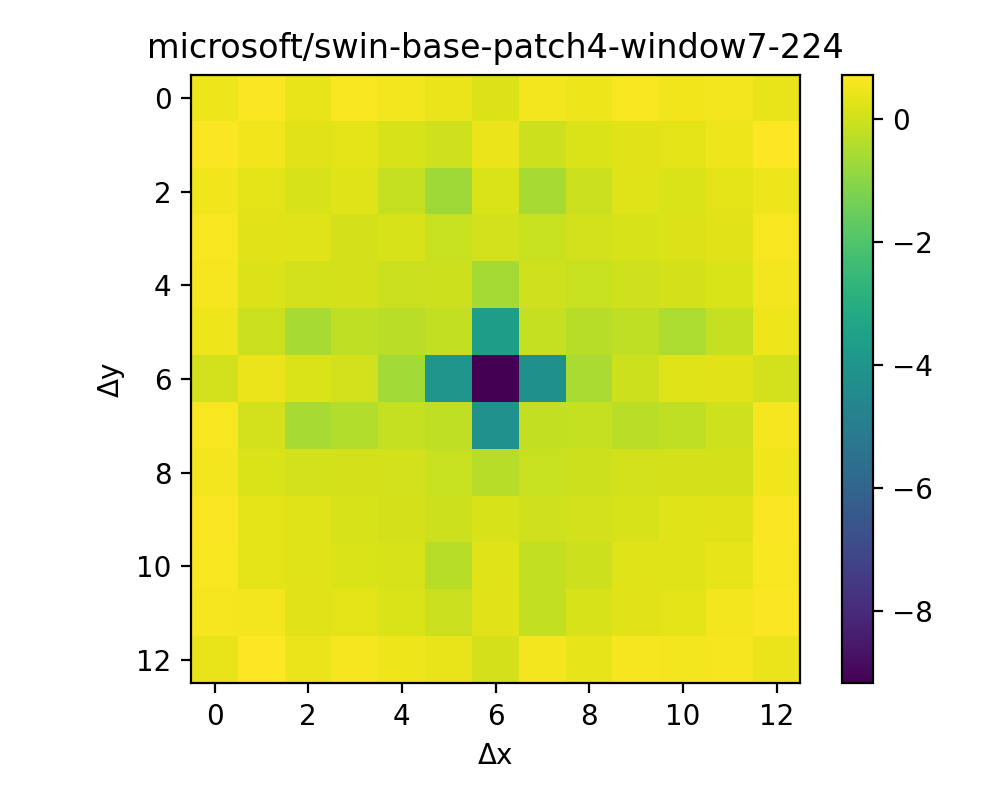}
         \caption{Swin}
     \end{subfigure}
     \hfill
     \begin{subfigure}[b]{.48\linewidth}
         \centering
         \includegraphics[width=.48\linewidth,trim={2.45cm 1.6cm 2.8cm .99cm},clip]{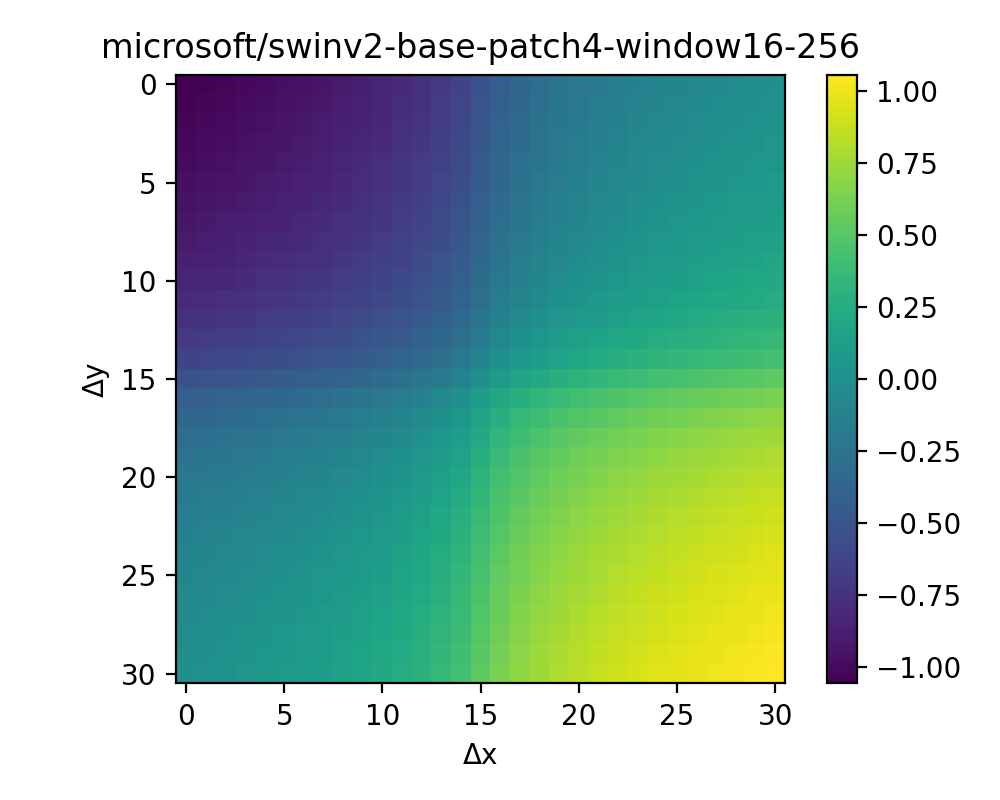}
         \includegraphics[width=.48\linewidth,trim={2.45cm 1.6cm 2.8cm .99cm},clip]{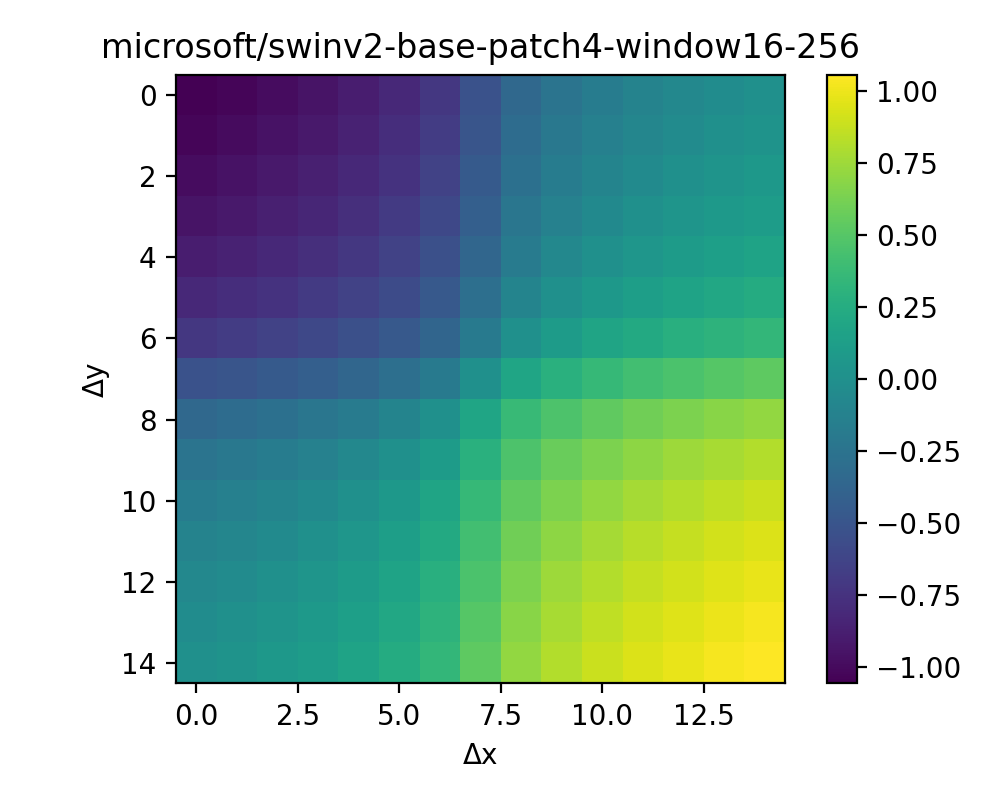}
         \caption{Swin v2}
     \end{subfigure}
     \caption{Visualization of PE-induced kernels. The left and right images are the the kernels from the first and last attention layers.}
    \label{fig:kernel_vis}
\end{figure}

\begin{table}[h]
    \centering
    \scriptsize
    \setlength{\tabcolsep}{1pt}
    \rowcolors{2}{verylightgray}{}
    \setlength{\abovecaptionskip}{0em}
    \setlength{\belowcaptionskip}{0em}
    \caption{
    \textbf{Token-offset reconstruction.}.
    Each entry reports the cosine similarity between a reference token and its displaced neighbor token at offset $k$, in a \textit{reconstruction/{\color{gray}\scriptsize baseline}} format. Without PEs, similarity becomes uniformly high across offsets, indicating that positional distinctiveness collapses.}
    \begin{tabular}{l|cc cH| cccH}
        \toprule
        & \multicolumn{4}{c|}{w/ PE} & \multicolumn{4}{c}{w/o PE} \\
        \cmidrule(lr){2-5} \cmidrule(lr){6-9}
          & $k=1$ & $k=2$ & $k=3$ & $k=5$  & $k=1$ & $k=2$ & $k=3$ & $k=5$ \\
        \midrule
        DINO      & 0.983\,/\,{\color{gray}\tiny 0.956} & 0.981\,/\,{\color{gray}\tiny 0.9334} & 0.980\,/\,{\color{gray}\tiny 0.917} & 0.981\,/\,{\color{gray}\tiny 0.904}
        & 0.981\,/\,{\color{gray}\tiny 0.970} & 0.979\,/\,{\color{gray}\tiny 0.965} & 0.979\,/\,{\color{gray}\tiny 0.962} & 0.979\,/\,{\color{gray}\tiny 0.960}
        \\
        {DINOv2}  & 0.722\,/\,{\color{gray}\tiny 0.606} & 0.692\,/\,{\color{gray}\tiny 0.573} & 0.670\,/\,{\color{gray}\tiny 0.472} & 0.660\,/\,{\color{gray}\tiny 0.402}
        & 0.909\,/\,{\color{gray}\tiny 0.879} & 0.902\,/\,{\color{gray}\tiny 0.858} & 0.900\,/\,{\color{gray}\tiny 0.618} & 0.900\,/\,{\color{gray}\tiny 0.514} 
        \\
        {DINOv3}   & 0.762\,/\,{\color{gray}\tiny 0.688} & 0.713\,/\,{\color{gray}\tiny 0.549} & 0.689\,/\,{\color{gray}\tiny 0.468} & 0.673\,/\,{\color{gray}\tiny 0.389}
        & 0.998\,/\,{\color{gray}\tiny 0.996} & 0.998\,/\,{\color{gray}\tiny 0.996} & 0.998\,/\,{\color{gray}\tiny 0.996} & 0.998\,/\,{\color{gray}\tiny 0.996} 
        \\
        {IJEPA}    & 0.912\,/\,{\color{gray}\tiny 0.664} & 0.915\,/\,{\color{gray}\tiny 0.537} & 0.919\,/\,{\color{gray}\tiny 0.427} & 0.924\,/\,{\color{gray}\tiny 0.271}
        & 0.998\,/\,{\color{gray}\tiny 0.998} & 0.998\,/\,{\color{gray}\tiny 0.997} & 0.998\,/\,{\color{gray}\tiny 0.998} & 0.998\,/\,{\color{gray}\tiny 0.998}
        \\
        % {MetaCLIP} & 0.870\,/\,{\color{gray}\tiny 0.730} & 0.861\,/\,{\color{gray}\tiny 0.774} & 0.849\,/\,{\color{gray}\tiny 0.709} & 0.840\,/\,{\color{gray}\tiny 0.710}
        % & 0.999\,/\,{\color{gray}\tiny 0.999} & 0.999\,/\,{\color{gray}\tiny 0.999} & 0.999\,/\,{\color{gray}\tiny 0.999} & 0.999\,/\,{\color{gray}\tiny 0.999}
        % \\
        {DEiT}    & 0.912\,/\,{\color{gray}\tiny 0.783} & 0.925\,/\,{\color{gray}\tiny 0.793} & 0.925\,/\,{\color{gray}\tiny 0.790} & 0.920\,/\,{\color{gray}\tiny 0.781}
        & 0.999\,/\,{\color{gray}\tiny 0.999} & 0.999\,/\,{\color{gray}\tiny 0.999} & 0.999\,/\,{\color{gray}\tiny 0.999} & 0.999\,/\,{\color{gray}\tiny 0.999} 
        \\
        {Siglip2}  & 0.682\,/\,{\color{gray}\tiny 0.511} & 0.674\,/\,{\color{gray}\tiny 0.482} & 0.674\,/\,{\color{gray}\tiny 0.463} & 0.680\,/\,{\color{gray}\tiny 0.455}
        & 0.999\,/\,{\color{gray}\tiny 0.999} & 0.999\,/\,{\color{gray}\tiny 0.999} & 0.999\,/\,{\color{gray}\tiny 0.999} & 0.999\,/\,{\color{gray}\tiny 0.999}
        \\
        BeiT      & 0.999\,/\,{\color{gray}\tiny 0.999} & 0.999\,/\,{\color{gray}\tiny 0.999} & 0.999\,/\,{\color{gray}\tiny 0.999} & 0.999\,/\,{\color{gray}\tiny 0.999}
        & 0.999\,/\,{\color{gray}\tiny 0.999} & 0.999\,/\,{\color{gray}\tiny 0.999} & 0.999\,/\,{\color{gray}\tiny 0.999} & 0.999\,/\,{\color{gray}\tiny 0.999}
        \\
        Data2Vec  & 0.984\,/\,{\color{gray}\tiny 0.929} & 0.973\,/\,{\color{gray}\tiny 0.894} & 0.966\,/\,{\color{gray}\tiny 0.866} & 0.959\,/\,{\color{gray}\tiny 0.815}
        & 0.999\,/\,{\color{gray}\tiny 0.999} & 0.999\,/\,{\color{gray}\tiny 0.999} & 0.999\,/\,{\color{gray}\tiny 0.999} & 0.999\,/\,{\color{gray}\tiny 0.999}
        \\
        MLCD  & 0.734\,/\,{\color{gray}\tiny 0.551} & 0.719\,/\,{\color{gray}\tiny 0.521} & 0.722\,/\,{\color{gray}\tiny 0.500} & 0.726\,/\,{\color{gray}\tiny 0.477}
        & 0.769\,/\,{\color{gray}\tiny 0.638}  & 0.761\,/\,{\color{gray}\tiny 0.609} & 0.759\,/\,{\color{gray}\tiny 0.594} & 0.759\,/\,{\color{gray}\tiny 0.580}
        \\
        VIT  & 0.875\,/\,{\color{gray}\tiny 0.710} & 0.875\,/\,{\color{gray}\tiny 0.631} & 0.862\,/\,{\color{gray}\tiny 0.624} & 0.891\,/\,{\color{gray}\tiny 0.580}
        & 0.999\,/\,{\color{gray}\tiny 0.999} & 0.999\,/\,{\color{gray}\tiny 0.999} & 0.999\,/\,{\color{gray}\tiny 0.999} & 0.999\,/\,{\color{gray}\tiny 0.999}
        \\
        SAM  & 0.994\,/\,{\color{gray}\tiny 0.987} & 0.992\,/\,{\color{gray}\tiny 0.984} & 0.990\,/\,{\color{gray}\tiny 0.984} & 0.989\,/\,{\color{gray}\tiny 0.976}
        & 0.998\,/\,{\color{gray}\tiny 0.996} & 0.996\,/\,{\color{gray}\tiny 0.995} & 0.995\,/\,{\color{gray}\tiny 0.993} & 0.994\,/\,{\color{gray}\tiny 0.989} 
        \\
        CLIP & 0.817\,/\,{\color{gray}\tiny 0.647} & 0.812\,/\,{\color{gray}\tiny 0.639} & 0.822\,/\,{\color{gray}\tiny 0.646} & 0.820\,/\,{\color{gray}\tiny 0.642}
        & 0.999\,/\,{\color{gray}\tiny 0.999} & 0.999\,/\,{\color{gray}\tiny 0.999} & 0.999\,/\,{\color{gray}\tiny 0.999} & 0.999\,/\,{\color{gray}\tiny 0.999} 
        \\
        Swin   & 0.754\,/\,{\color{gray}\tiny 0.528} & 0.740\,/\,{\color{gray}\tiny 0.511} & 0.755\,/\,{\color{gray}\tiny 0.453} & 0.882\,/\,{\color{gray}\tiny 0.552}
        & 0.912\,/\,{\color{gray}\tiny 0.818}  & 0.909\,/\,{\color{gray}\tiny 0.830} & 0.909\,/\,{\color{gray}\tiny 0.832} & 0.926\,/\,{\color{gray}\tiny 0.862} 
        \\
        SwinV2 & 0.779\,/\,{\color{gray}\tiny 0.651} & 0.760\,/\,{\color{gray}\tiny 0.587} & 0.765\,/\,{\color{gray}\tiny 0.584} & 0.809\,/\,{\color{gray}\tiny 0.645}
        & 0.903\,/\,{\color{gray}\tiny 0.842} & 0.895\,/\,{\color{gray}\tiny 0.804} & 0.899\,/\,{\color{gray}\tiny 0.796} & 0.919\,/\,{\color{gray}\tiny 0.834} 
        \\
        \bottomrule
    \end{tabular}
    \label{tab:tokenoffset}
\end{table}

\begin{table*}[t]
\centering
\small
\setlength{\tabcolsep}{3pt}
 \begin{subtable}[t]{0.48\textwidth}
\begin{tabular}{l|cHcHcH|cHcHcH}
\toprule
\multirow{2}{*}{Model} &
\multicolumn{6}{c|}{\textbf{Bilinear}} &
\multicolumn{6}{c}{\textbf{AnyUp}} \\
\cmidrule(lr){2-7} \cmidrule(lr){8-13}
& EPE↓ & D1↓ & R@1↑ & R@2↑ & R@5↑ & EC-SIM↑
& EPE↓ & D1↓ & R@1↑ & R@2↑ & R@5↑ & EC-SIM↑ \\
\midrule
DINO       & 11.40 & 0.843 & 0.146 & 0.320 & 0.560 & 0.989
           & 13.02 & 0.859 & 0.285 & 0.421 & 0.610 & 0.958 \\
DINOv2     & 12.64 & 0.810 & 0.131 & 0.296 & 0.515 & 0.890
           & 13.82 & 0.857 & 0.240 & 0.368 & 0.555 & 0.899 \\
DINOv3     & 11.66 & 0.828 & 0.129 & 0.267 & 0.474 & 0.803
           & 12.96 & 0.861 & 0.178 & 0.291 & 0.477 & 0.821 \\
I-JEPA     &  9.90 & 0.787 & 0.151 & 0.322 & 0.561 & 0.940
           & 10.98 & 0.819 & 0.259 & 0.392 & 0.587 & 0.938 \\
% MetaCLIP-B16-FullCC2.5B & 31.61 & 0.923 & 0.140 & 0.311 & 0.534 & 0.910 & 31.863 & 0.925 & 0.253 & 0.381 & 0.565 & 0.903 \\
DEiT       & 14.69 & 0.809 & 0.152 & 0.320 & 0.553 & 0.908
           & 16.63 & 0.842 & 0.266 & 0.397 & 0.586 & 0.906 \\
SigLIP2    & 11.04 & 0.656 & 0.141 & 0.299 & 0.519 & 0.760
           & 11.33 & 0.740 & 0.221 & 0.347 & 0.544 & 0.770 \\
BeiT       & 28.98 & 0.901 & 0.143 & 0.298 & 0.531 & 0.939
           & 29.46 & 0.912 & 0.211 & 0.337 & 0.545 & 0.942 \\
Data2Vec   & 17.55 & 0.919 & 0.153 & 0.324 & 0.564 & 0.986
           & 24.56 & 0.917 & 0.256 & 0.389 & 0.588 & 0.976 \\
MLCD       & 23.25 & 0.811 & 0.112 & 0.237 & 0.416 & 0.745
           & 22.19 & 0.844 & 0.150 & 0.250 & 0.422 & 0.768 \\
ViT        & 22.13 & 0.825 & 0.142 & 0.312 & 0.545 & 0.925
           & 23.63 & 0.857 & 0.261 & 0.390 & 0.576 & 0.917\\
SAM        & 19.67 & 0.877 & 0.153 & 0.330 & 0.569 & 0.961
           & 20.94 & 0.927 & 0.289 & 0.438 & 0.634 & 0.982 \\
CLIP       & 21.39 & 0.793 & 0.148 & 0.310 & 0.531 & 0.847
           & 21.40 & 0.829 & 0.232 & 0.355 & 0.538 & 0.841 \\
Swin       & 30.87 & 0.900 & 0.087 & 0.182 & 0.331 & 0.734
           & 32.82 & 0.913 & 0.082 & 0.145 & 0.274 & 0.716 \\
SwinV2     & 27.53 & 0.903 & 0.095 & 0.198 & 0.360 & 0.811
           & 30.17 & 0.916 & 0.096 & 0.166 & 0.305 & 0.798 \\
\bottomrule
\end{tabular}
\caption{w/ PE}
\end{subtable}
 \begin{subtable}[t]{0.48\textwidth}
\begin{tabular}{l|cHcHcH|cHcHcH}
\toprule
\multirow{2}{*}{Model} &
\multicolumn{6}{c|}{\textbf{Bilinear}} &
\multicolumn{6}{c}{\textbf{AnyUp}} \\
\cmidrule(lr){2-7} \cmidrule(lr){8-13}
& EPE↓ & D1↓ & R@1↑ & R@2↑ & R@5↑ & EC-SIM↑
& EPE↓ & D1↓ & R@1↑ & R@2↑ & R@5↑ & EC-SIM↑ \\
\midrule
DINO       & 24.21 & 0.861 & 0.126 & 0.265 & 0.477 & 0.846
           & 26.68 & 0.890 & 0.174 & 0.284 & 0.471 & 0.857 \\
DINOv2     & 38.08 & 0.908 & 0.087 & 0.179 & 0.315 & 0.692
           & 38.00 & 0.925 & 0.099 & 0.167 & 0.296 & 0.705 \\
DINOv3     & 51.89 & 0.934 & 0.080 & 0.166 & 0.293 & 0.605
           & 51.32 & 0.941 & 0.088 & 0.148 & 0.264 & 0.619 \\
I-JEPA     & 44.97 & 0.927 & 0.088 & 0.183 & 0.321 & 0.782
           & 45.99 & 0.940 & 0.096 & 0.166 & 0.301 & 0.805 \\
% MetaCLIP & 23.74 & 0.824 & 0.131 & 0.270 & 0.470 & 0.665
         % & 24.75 & 0.863 & 0.159 & 0.263 & 0.437 & 0.687 \\
DEiT       & 26.22 & 0.861 & 0.122 & 0.254 & 0.436 & 0.640
           & 27.53 & 0.889 & 0.149 & 0.247 & 0.411 & 0.680 \\
SigLIP2    & 38.99 & 0.909 & 0.088 & 0.183 & 0.312 & 0.678
           & 39.70 & 0.924 & 0.099 & 0.170 & 0.300 & 0.741 \\
BeiT       & 46.31 & 0.929 & 0.115 & 0.243 & 0.441 & 0.872
           & 48.13 & 0.942 & 0.145 & 0.242 & 0.415 & 0.882 \\
Data2Vec   & 39.64 & 0.907 & 0.112 & 0.239 & 0.424 & 0.786
           & 39.67 & 0.923 & 0.141 & 0.234 & 0.394 & 0.781 \\
MLCD       & 38.36 & 0.859 & 0.086 & 0.182 & 0.314 & 0.566
           & 38.43 & 0.889 & 0.096 & 0.163 & 0.289 & 0.580 \\
ViT        & 28.36 & 0.875 & 0.110 & 0.234 & 0.414 & 0.745
           & 29.29 & 0.899 & 0.136 & 0.227 & 0.387 & 0.760 \\
SAM        & 47.07 & 0.894 & 0.156 & 0.332 & 0.574 & 0.964
           & 45.73 & 0.917 & 0.292 & 0.438 & 0.644 & 0.975 \\
CLIP       & 53.80 & 0.941 & 0.081 & 0.166 & 0.290 & 0.541
           & 52.79 & 0.948 & 0.087 & 0.149 & 0.268 & 0.588 \\
Swin       & 34.89 & 0.906 & 0.088 & 0.182 & 0.323 & 0.661
           & 37.70 & 0.923 & 0.075 & 0.133 & 0.251 & 0.639 \\
SwinV2     & 33.99 & 0.917 & 0.088 & 0.186 & 0.329 & 0.792 
           & 35.14 & 0.925 & 0.084 & 0.147 & 0.275 & 0.778 \\
\bottomrule
\end{tabular}
\caption{w/o PE}
\end{subtable}
\caption{\textbf{Comparison of different upsampling method for the stereo probing results at 448× resolution}. We can see that the use of AnyUp will not affect EPE much (apart from Data2Vec), but can significantly improve the Recall performances.
}
\label{tab:stereo_upsample_comparison}
\end{table*}

\begin{figure*}
    \centering
    \setlength{\tabcolsep}{0pt}
    
    \begin{tabular}{cc cc || cc cc}

         \multicolumn{2}{c}{\textbf{DINO} w/ PE} & \multicolumn{2}{c||}{\textbf{DINO} w/o PE}
         & \multicolumn{2}{c}{\textbf{DINOv2} w/ PE} & \multicolumn{2}{c}{\textbf{DINOv2} w/o PE}\\
         \includegraphics[width=0.12\linewidth,trim={12.6cm 0 12.6cm 1.2cm},clip]{misc/disp/epipolar_response_pos_anyup_pca_dino-vitb16.png} &
         \includegraphics[width=0.12\linewidth,trim={25.2cm 0 0 1.2cm},clip]{misc/disp/epipolar_response_pos_anyup_pca_dino-vitb16.png} &
         \includegraphics[width=0.12\linewidth,trim={12.6cm 0 12.6cm 1.2cm},clip]{misc/disp/epipolar_response_no_pos_anyup_pca_dino-vitb16.png} &
         \includegraphics[width=0.12\linewidth,trim={25.2cm 0 0 1.2cm},clip]{misc/disp/epipolar_response_no_pos_anyup_pca_dino-vitb16.png}
         &
         \includegraphics[width=0.12\linewidth,trim={12.6cm 0 12.6cm 1.2cm},clip]{misc/disp/epipolar_response_pos_anyup_pca_dinov2-large.png} &
         \includegraphics[width=0.12\linewidth,trim={25.2cm 0 0 1.2cm},clip]{misc/disp/epipolar_response_pos_anyup_pca_dinov2-large.png} &
         \includegraphics[width=0.12\linewidth,trim={12.6cm 0 12.6cm 1.2cm},clip]{misc/disp/epipolar_response_no_pos_anyup_pca_dinov2-large.png} &
         \includegraphics[width=0.12\linewidth,trim={25.2cm 0 0 1.2cm},clip]{misc/disp/epipolar_response_no_pos_anyup_pca_dinov2-large.png} \\
         
         \multicolumn{2}{c}{\textbf{DINOv3} w/ PE} & \multicolumn{2}{c||}{\textbf{DINOv3} w/o PE} 
         & \multicolumn{2}{c}{\textbf{I-JEPA} w/ PE} & \multicolumn{2}{c}{\textbf{I-JEPA} w/o PE} \\
         \includegraphics[width=0.12\linewidth,trim={12.6cm 0 12.6cm 1.2cm},clip]{misc/disp/epipolar_response_pos_anyup_pca_dinov3-vit7b16-pretrain-lvd1689m.png} &
         \includegraphics[width=0.12\linewidth,trim={25.2cm 0 0 1.2cm},clip]{misc/disp/epipolar_response_pos_anyup_pca_dinov3-vit7b16-pretrain-lvd1689m.png} &
         \includegraphics[width=0.12\linewidth,trim={12.6cm 0 12.6cm 1.2cm},clip]{misc/disp/epipolar_response_no_pos_anyup_pca_dinov3-vit7b16-pretrain-lvd1689m.png} &
         \includegraphics[width=0.12\linewidth,trim={25.2cm 0 0 1.2cm},clip]{misc/disp/epipolar_response_no_pos_anyup_pca_dinov3-vit7b16-pretrain-lvd1689m.png}
         &
         \includegraphics[width=0.12\linewidth,trim={12.6cm 0 12.6cm 1.2cm},clip]{misc/disp/epipolar_response_pos_anyup_pca_ijepa_vitg16_22k.png} &
         \includegraphics[width=0.12\linewidth,trim={25.2cm 0 0 1.2cm},clip]{misc/disp/epipolar_response_pos_anyup_pca_ijepa_vitg16_22k.png} &
         \includegraphics[width=0.12\linewidth,trim={12.6cm 0 12.6cm 1.2cm},clip]{misc/disp/epipolar_response_no_pos_anyup_pca_ijepa_vitg16_22k.png} &
         \includegraphics[width=0.12\linewidth,trim={25.2cm 0 0 1.2cm},clip]{misc/disp/epipolar_response_no_pos_anyup_pca_ijepa_vitg16_22k.png} \\
         
         \multicolumn{2}{c}{\textbf{Swin} w/ PE} & \multicolumn{2}{c||}{\textbf{Swin} w/o PE} 
         & \multicolumn{2}{c}{\textbf{SwinV2} w/ PE} & \multicolumn{2}{c}{\textbf{SwinV2} w/o PE} \\
         \includegraphics[width=0.12\linewidth,trim={12.6cm 0 12.6cm 1.2cm},clip]{misc/disp/epipolar_response_pos_anyup_pca_swin-base-patch4-window7-224.png} &
         \includegraphics[width=0.12\linewidth,trim={25.2cm 0 0 1.2cm},clip]{misc/disp/epipolar_response_pos_anyup_pca_swin-base-patch4-window7-224.png} &
         \includegraphics[width=0.12\linewidth,trim={12.6cm 0 12.6cm 1.2cm},clip]{misc/disp/epipolar_response_no_pos_anyup_pca_swin-base-patch4-window7-224.png} &
         \includegraphics[width=0.12\linewidth,trim={25.2cm 0 0 1.2cm},clip]{misc/disp/epipolar_response_no_pos_anyup_pca_swin-base-patch4-window7-224.png}
         &
         \includegraphics[width=0.12\linewidth,trim={12.6cm 0 12.6cm 1.2cm},clip]{misc/disp/epipolar_response_pos_anyup_pca_swinv2-base-patch4-window16-256.png} &
         \includegraphics[width=0.12\linewidth,trim={25.2cm 0 0 1.2cm},clip]{misc/disp/epipolar_response_pos_anyup_pca_swinv2-base-patch4-window16-256.png} &
         \includegraphics[width=0.12\linewidth,trim={12.6cm 0 12.6cm 1.2cm},clip]{misc/disp/epipolar_response_no_pos_anyup_pca_swinv2-base-patch4-window16-256.png} &
         \includegraphics[width=0.12\linewidth,trim={25.2cm 0 0 1.2cm},clip]{misc/disp/epipolar_response_no_pos_anyup_pca_swinv2-base-patch4-window16-256.png}
         \\
         \multicolumn{2}{c}{\textbf{MLCD} w/ PE} & \multicolumn{2}{c||}{\textbf{MLCD} w/o PE}
         & \multicolumn{2}{c}{\textbf{Data2Vec} w/ PE} & \multicolumn{2}{c}{\textbf{Data2Vec} w/o PE}  \\
         \includegraphics[width=0.12\linewidth,trim={12.6cm 0 12.6cm 1.2cm},clip]{misc/disp/epipolar_response_pos_anyup_pca_mlcd-vit-bigG-patch14-224.png} &
         \includegraphics[width=0.12\linewidth,trim={25.2cm 0 0 1.2cm},clip]{misc/disp/epipolar_response_pos_anyup_pca_mlcd-vit-bigG-patch14-224.png} &
         \includegraphics[width=0.12\linewidth,trim={12.6cm 0 12.6cm 1.2cm},clip]{misc/disp/epipolar_response_no_pos_anyup_pca_mlcd-vit-bigG-patch14-224.png} &
         \includegraphics[width=0.12\linewidth,trim={25.2cm 0 0 1.2cm},clip]{misc/disp/epipolar_response_no_pos_anyup_pca_mlcd-vit-bigG-patch14-224.png}
         &
         \includegraphics[width=0.12\linewidth,trim={12.6cm 0 12.6cm 1.2cm},clip]{misc/disp/epipolar_response_pos_anyup_pca_data2vec-vision-large.png} &
         \includegraphics[width=0.12\linewidth,trim={25.2cm 0 0 1.2cm},clip]{misc/disp/epipolar_response_pos_anyup_pca_data2vec-vision-large.png} &
         \includegraphics[width=0.12\linewidth,trim={12.6cm 0 12.6cm 1.2cm},clip]{misc/disp/epipolar_response_no_pos_anyup_pca_data2vec-vision-large.png} &
         \includegraphics[width=0.12\linewidth,trim={25.2cm 0 0 1.2cm},clip]{misc/disp/epipolar_response_no_pos_anyup_pca_data2vec-vision-large.png}
         \\
         \multicolumn{2}{c}{\textbf{SAM} w/ PE} & \multicolumn{2}{c||}{\textbf{SAM} w/o PE} 
         & \multicolumn{2}{c}{\textbf{CLIP} w/ PE} & \multicolumn{2}{c}{\textbf{CLIP} w/o PE} \\
         \includegraphics[width=0.12\linewidth,trim={12.6cm 0 12.6cm 1.2cm},clip]{misc/disp/epipolar_response_pos_anyup_pca_sam-vit-base.png} &
         \includegraphics[width=0.12\linewidth,trim={25.2cm 0 0 1.2cm},clip]{misc/disp/epipolar_response_pos_anyup_pca_sam-vit-base.png} &
         \includegraphics[width=0.12\linewidth,trim={12.6cm 0 12.6cm 1.2cm},clip]{misc/disp/epipolar_response_no_pos_anyup_pca_sam-vit-base.png} &
         \includegraphics[width=0.12\linewidth,trim={25.2cm 0 0 1.2cm},clip]{misc/disp/epipolar_response_no_pos_anyup_pca_sam-vit-base.png}
         &
         \includegraphics[width=0.12\linewidth,trim={12.6cm 0 12.6cm 1.2cm},clip]{misc/disp/epipolar_response_pos_anyup_pca_clip-vit-large-patch14.png} &
         \includegraphics[width=0.12\linewidth,trim={25.2cm 0 0 1.2cm},clip]{misc/disp/epipolar_response_pos_anyup_pca_clip-vit-large-patch14.png} &
         \includegraphics[width=0.12\linewidth,trim={12.6cm 0 12.6cm 1.2cm},clip]{misc/disp/epipolar_response_no_pos_anyup_pca_clip-vit-large-patch14.png} &
         \includegraphics[width=0.12\linewidth,trim={25.2cm 0 0 1.2cm},clip]{misc/disp/epipolar_response_no_pos_anyup_pca_clip-vit-large-patch14.png}  \\
         \multicolumn{2}{c}{\textbf{SigLip2} w/ PE} & \multicolumn{2}{c||}{\textbf{SigLip2} w/o PE} 
         & \multicolumn{2}{c}{\textbf{BeiT} w/ PE} & \multicolumn{2}{c}{\textbf{BeiT} w/o PE} \\
         \includegraphics[width=0.12\linewidth,trim={12.6cm 0 12.6cm 1.2cm},clip]{misc/disp/epipolar_response_pos_anyup_pca_siglip2-so400m-patch14-224.png} &
         \includegraphics[width=0.12\linewidth,trim={25.2cm 0 0 1.2cm},clip]{misc/disp/epipolar_response_pos_anyup_pca_siglip2-so400m-patch14-224.png} &
         \includegraphics[width=0.12\linewidth,trim={12.6cm 0 12.6cm 1.2cm},clip]{misc/disp/epipolar_response_no_pos_anyup_pca_siglip2-so400m-patch14-224.png} &
         \includegraphics[width=0.12\linewidth,trim={25.2cm 0 0 1.2cm},clip]{misc/disp/epipolar_response_no_pos_anyup_pca_siglip2-so400m-patch14-224.png}
         &
         \includegraphics[width=0.12\linewidth,trim={12.6cm 0 12.6cm 1.2cm},clip]{misc/disp/epipolar_response_pos_anyup_pca_beit-base-patch16-224-pt22k.png} &
         \includegraphics[width=0.12\linewidth,trim={25.2cm 0 0 1.2cm},clip]{misc/disp/epipolar_response_pos_anyup_pca_beit-base-patch16-224-pt22k.png} &
         \includegraphics[width=0.12\linewidth,trim={12.6cm 0 12.6cm 1.2cm},clip]{misc/disp/epipolar_response_pos_anyup_pca_beit-base-patch16-224-pt22k.png} &
         \includegraphics[width=0.12\linewidth,trim={25.2cm 0 0 1.2cm},clip]{misc/disp/epipolar_response_pos_anyup_pca_beit-base-patch16-224-pt22k.png} \\
         \multicolumn{2}{c}{\textbf{ViT} w/ PE} & \multicolumn{2}{c||}{\textbf{ViT} w/o PE} 
         & \multicolumn{2}{c}{\textbf{DEiT} w/ PE} & \multicolumn{2}{c}{\textbf{DEiT} w/o PE} \\
         \includegraphics[width=0.12\linewidth,trim={12.6cm 0 12.6cm 1.2cm},clip]{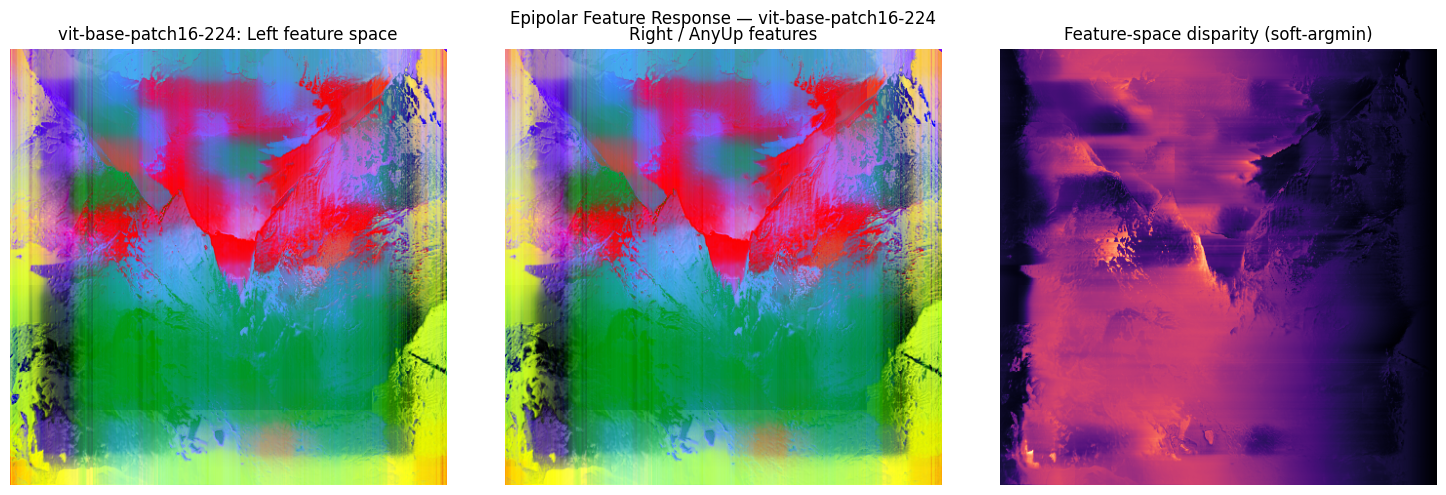} &
         \includegraphics[width=0.12\linewidth,trim={25.2cm 0 0 1.2cm},clip]{misc/disp/epipolar_response_pos_anyup_pca_vit-base-patch16-224.png} &
         \includegraphics[width=0.12\linewidth,trim={12.6cm 0 12.6cm 1.2cm},clip]{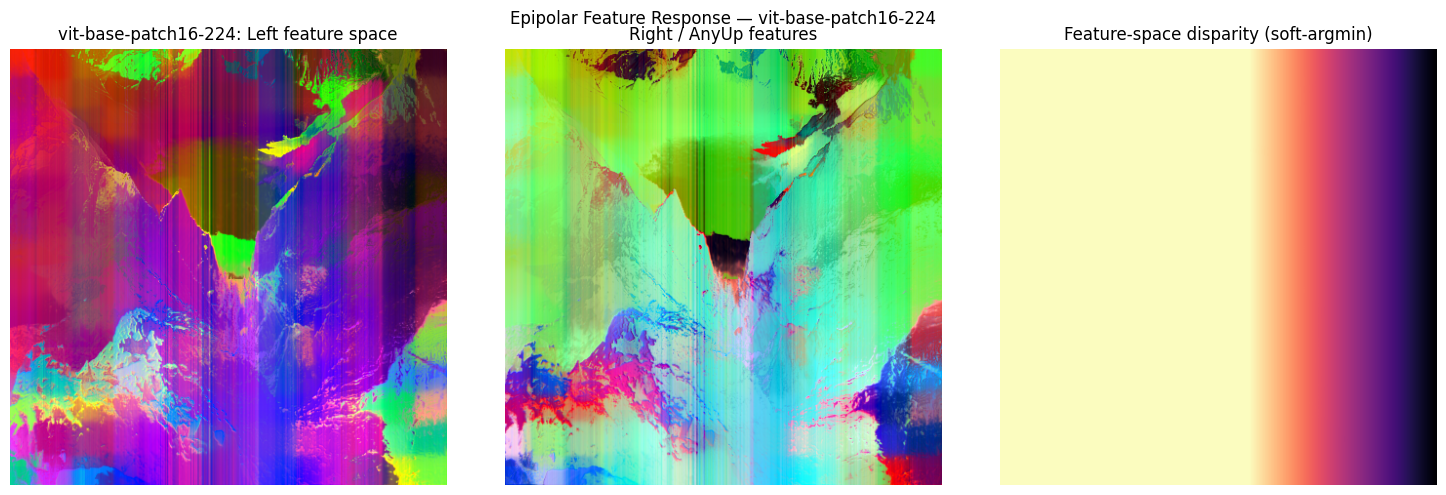} &
         \includegraphics[width=0.12\linewidth,trim={25.2cm 0 0 1.2cm},clip]{misc/disp/epipolar_response_no_pos_anyup_pca_vit-base-patch16-224.png}
         &
         \includegraphics[width=0.12\linewidth,trim={12.6cm 0 12.6cm 1.2cm},clip]{misc/disp/epipolar_response_pos_anyup_pca_deit-base-patch16-224.png} &
         \includegraphics[width=0.12\linewidth,trim={25.2cm 0 0 1.2cm},clip]{misc/disp/epipolar_response_pos_anyup_pca_deit-base-patch16-224.png} &
         \includegraphics[width=0.12\linewidth,trim={12.6cm 0 12.6cm 1.2cm},clip]{misc/disp/epipolar_response_no_pos_anyup_pca_deit-base-patch16-224.png} &
         \includegraphics[width=0.12\linewidth,trim={25.2cm 0 0 1.2cm},clip]{misc/disp/epipolar_response_no_pos_anyup_pca_deit-base-patch16-224.png} \\
    \end{tabular}
    \caption{\textbf{Visualization of the AnyUp interpolation}
    of feature activations and their epipolar responses (left and right of each pair, respectively), shown with and without PE.
    With PE, models exhibit coherent spatial structure and well-localized epipolar responses.
    Removing PE consistently collapses both feature geometry and epipolar localization, manifesting as noisy patterns (DINO, Swin), stripe artifacts (Swin/SAM), or nearly blank activations (Data2Vec, MetaClip, SigLIP2).}
    \label{fig:epipolar_vis_all}
\end{figure*}

% \begin{figure*}
%     \centering
%     \setlength{\tabcolsep}{0pt}
%     \input{figures/epipolar_vis_2}
%     \caption{\textbf{Continued Visualization of the AnyUp interpolation}. The general trends are preserved.}
%     \label{fig:epipolar_vis_all2}
% \end{figure*}

\subsection{Token-Level Epipolar Consistency}

We present further evaluation results for the epipolar probing experiment as in~\Cref{subsec:remove_pe_locality}. For each encoder, we extract low-resolution dense feature maps and upsample them to input resolution using two different strategies:
(1) a standard \textbf{Bilinear} interpolation baseline, and
(2) the adaptive \textbf{AnyUp} method~\cite{wimmer2025anyup}, which learns content-aware upsampling filters that better preserve high-frequency structure and contextual cues.
All models are evaluated at $448\times$ resolution using our feature-level epipolar probe, with results summarized in~\cref{tab:stereo_upsample_comparison}. More visualizations are presented in~\cref{fig:epipolar_vis_all}.

\paragraph{Results.}
With PE, models produce structured and spatially coherent feature maps. Their epipolar responses are sharply localized, indicating that positional information strongly constrains the spatial neighborhood over which tokens interact. This behavior is especially pronounced in DINO-series models, where the presence of PE yields clean, directional epipolar peaks that align with the underlying stereo geometry.

Removing PE leads to a dramatic collapse in geometric structure for nearly all models. Notably, aside of the epipolar response, removing PE can also significantly affect the feature space in DINO and SAM models. The MLCD model is the model unaffected models that preserves local features even without PE. In all other cases, the epipolar response becomes flat, diffuse, or completely uninformative, confirming that the geometry collapses when positional cues are removed.

\begin{figure}[h]
    \centering
\setlength{\tabcolsep}{0pt}
    \begin{tabular}{cc cc}
         \multicolumn{2}{c}{\textbf{DINOv2} w/ PE} & \multicolumn{2}{c}{\textbf{DINOv2} w/o PE}\\
         \includegraphics[width=0.24\linewidth,trim={12.6cm 0 12.6cm 1.2cm},clip]{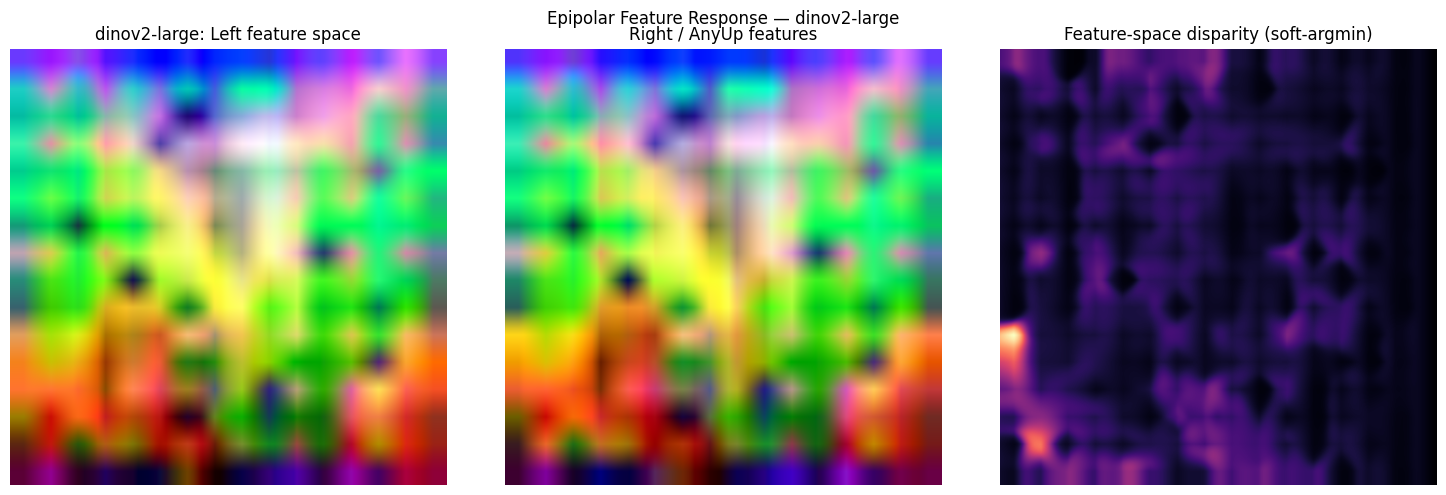} &
         \includegraphics[width=0.24\linewidth,trim={25.2cm 0 0 1.2cm},clip]{misc/disp/epipolar_response_pos_bilinear_pca_dinov2-large.png} &
         \includegraphics[width=0.24\linewidth,trim={12.6cm 0 12.6cm 1.2cm},clip]{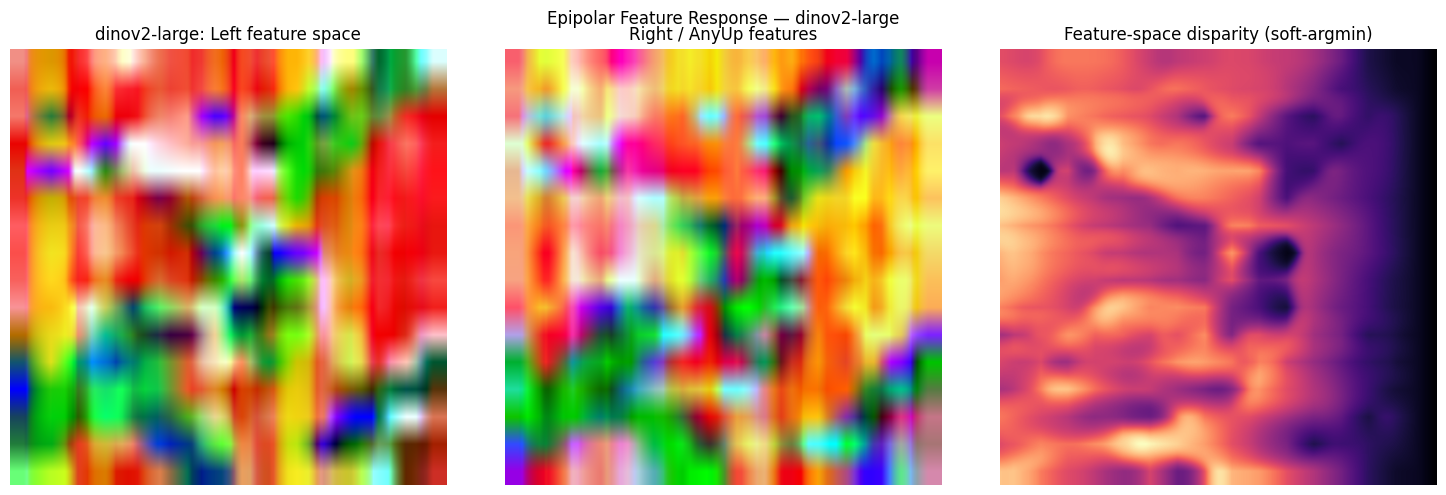} &
         \includegraphics[width=0.24\linewidth,trim={25.2cm 0 0 1.2cm},clip]{misc/disp/epipolar_response_no_pos_bilinear_pca_dinov2-large.png}
    \end{tabular}
    \caption{\textbf{Visualization of the direct bilinear interpolation} of feature activations and their epipolar responses (left and right of each pair, respectively), shown with and without PE. Visualizations are not semantically meaningful.}
    \label{fig:epipolar_vis_all_bilinear}
\end{figure}

\paragraph{AnyUp vs. Bilinear.}
Apart from Data2Vec, AnyUp does not affect the EPE for most models.
{AnyUp} retains the underlying feature structure and enriches local semantics, resulting in higher recall (R@1 and R@5) across most models.
These improvements reflect that AnyUp can be used as a visualization tool that can preserve both the semantic distinctiveness without harming too much of the epipolar alignment of deep features. \Cref{fig:epipolar_vis_all_bilinear} presents the  direct bilinear interpolation of feature activations and their epipolar responses.

\begin{figure*}[t]
    \centering
    \scriptsize
    \setlength{\tabcolsep}{0pt}
    
    \renewcommand{\arraystretch}{0.25}
    \begin{tabular}{@{}cc cc cc cc@{}}

        \multicolumn{3}{c}{\hspace{1em} \textbf{ Input Left Image \hspace{3em} Input Right Image}}
        & \multicolumn{3}{l}{\hspace{1em} \textbf{SAM Feature Visualization (Left only)}}
        & \multicolumn{2}{c}{\hspace{1em} \textbf{Epipolar Peak Response}} \\
        \cmidrule(lr){1-3} \cmidrule(lr){4-6} \cmidrule(lr){7-8}
        \multicolumn{3}{c}{
            \setlength{\tabcolsep}{1pt} 
            \begin{tabular}{@{}cc@{}}
                 \includegraphics[width=0.166\linewidth]{misc/disp/temp_L.png}
                 &
                 \includegraphics[width=0.166\linewidth]{misc/disp/temp_R.png} \\
            \end{tabular}
        }
        &
        \multicolumn{3}{c!{\vrule width 1pt}}{
            \setlength{\tabcolsep}{0pt} 
            \begin{tabular}{@{}cc@{}}
                {shuffled PE} & {pair shuffled PE} \\
                 \includegraphics[width=0.167\linewidth,trim={12.6cm 0 12.6cm 1.2cm},clip]{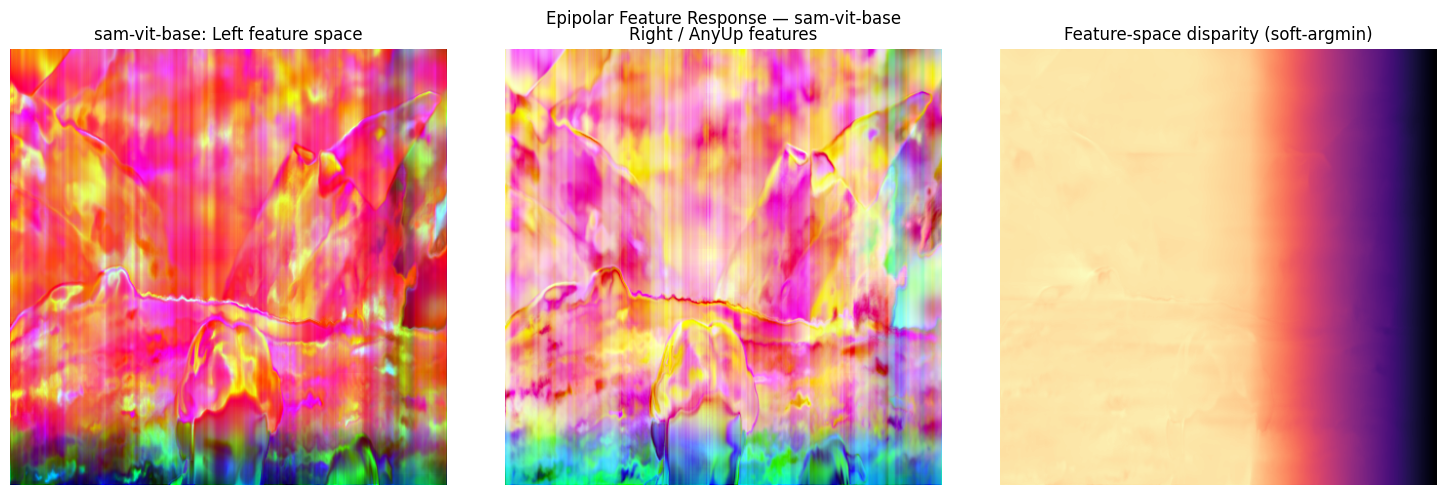}
                 &
                 \includegraphics[width=0.167\linewidth,trim={12.6cm 0 12.6cm 1.2cm},clip]{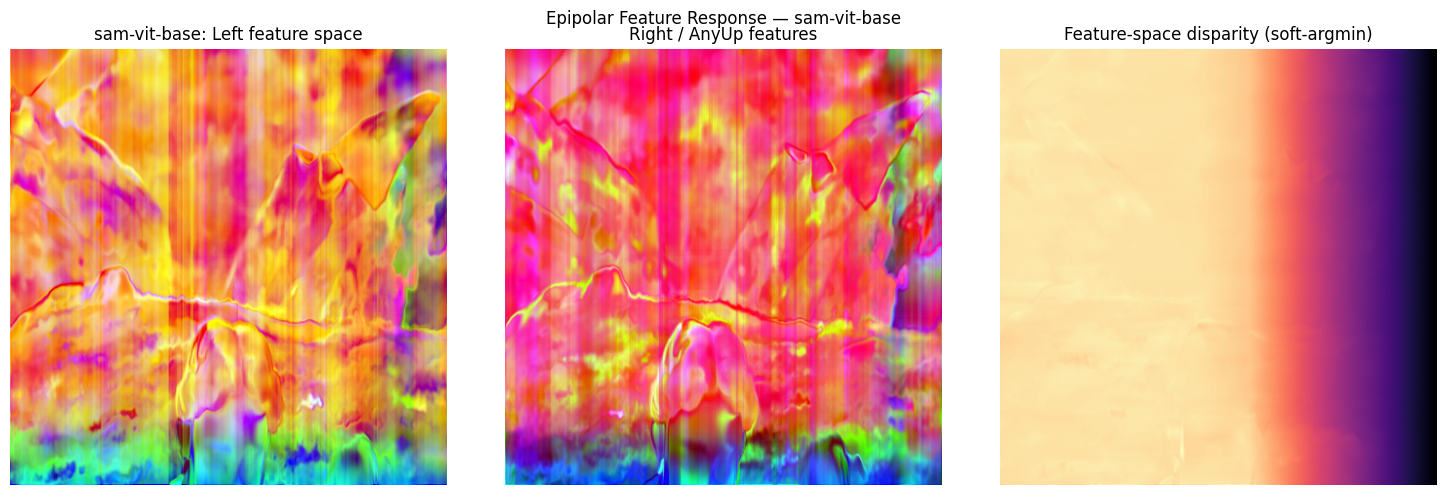} \\
            \end{tabular}
        }
        &
        \multicolumn{2}{c}{
            \setlength{\tabcolsep}{0pt} 
            \begin{tabular}{cc}
                \multicolumn{2}{l}{\hspace{1em} \textbf{SAM}}  \\
                \cmidrule(lr){1-1}
                {shuffled PE} & {pair shuffled PE} \\
                 \includegraphics[width=0.12\linewidth,trim={25.2cm 0 0 1.2cm},clip]{misc/epipolar_response_plots_no_reshuffle/epipolar_response_pos_anyup_pca_sam-vit-base.png}
                 &
                 \includegraphics[width=0.12\linewidth,trim={25.2cm 0 0 1.2cm},clip]{misc/epipolar_response_plots_no_reshuffle/epipolar_response_pos_anyup_pca_sam-vit-base.png} \\
            \end{tabular}
        }

         % \multicolumn{2}{c}{\textbf{SAM} shuffled PE} & \multicolumn{2}{c}{\textbf{SAM} pair shuffled PE}\\
         % \includegraphics[width=0.12\linewidth,trim={12.6cm 0 12.6cm 1.2cm},clip]{misc/epipolar_response_plots_no_reshuffle/epipolar_response_pos_anyup_pca_sam-vit-base.png} &
         % \includegraphics[width=0.12\linewidth,trim={25.2cm 0 0 1.2cm},clip]{misc/epipolar_response_plots_no_reshuffle/epipolar_response_pos_anyup_pca_sam-vit-base.png} &
         % \includegraphics[width=0.12\linewidth,trim={12.6cm 0 12.6cm 1.2cm},clip]{misc/epipolar_response_plots_no_reshuffle/epipolar_response_pos_anyup_pca_sam-vit-base.png} &
         % \includegraphics[width=0.12\linewidth,trim={25.2cm 0 0 1.2cm},clip]{misc/epipolar_response_plots_no_reshuffle/epipolar_response_pos_anyup_pca_sam-vit-base.png}
         \\
         \cmidrule[1pt](lr){1-6}

        \multicolumn{2}{l}{\hspace{1em} \textbf{DINO}}
        & \multicolumn{2}{l}{\hspace{1em} \textbf{DINOv2}} 
        & \multicolumn{2}{l}{\hspace{1em} \textbf{DINOv3}} 
        & \multicolumn{2}{l}{\hspace{1em} \textbf{I-JEPA}}  \\
        \cmidrule(lr){1-1} \cmidrule(lr){3-3} \cmidrule(lr){5-5} \cmidrule(lr){7-7}

         {\scriptsize shuffled PE} & {\scriptsize pair shuffled PE}
         & {\scriptsize shuffled PE} & {\scriptsize pair shuffled PE}
         & {\scriptsize shuffled PE} & {\scriptsize pair shuffled PE} 
         & {\scriptsize shuffled PE} & {\scriptsize pair shuffled PE} \\

         % {\textbf{DINO} shuffled PE} & {\textbf{DINO} pair shuffled PE}
         % & {\textbf{DINOv2} shuffled PE} & {\textbf{DINOv2} pair shuffled PE}
         % & {\textbf{DINOv3} shuffled PE} & {\textbf{DINOv3} pair shuffled PE} 
         % & {\textbf{I-JEPA} pair shuffled PE} & {\textbf{I-JEPA} pair shuffled PE} \\
         
         \includegraphics[width=0.12\linewidth,trim={25.2cm 0 0 1.2cm},clip]{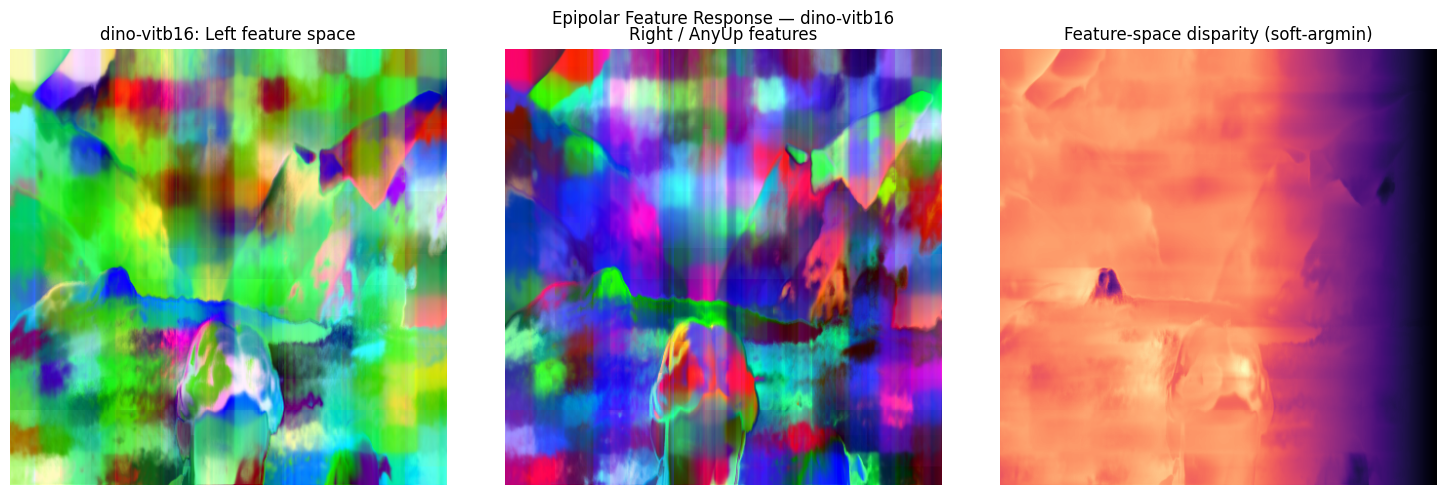} &
         \includegraphics[width=0.12\linewidth,trim={25.2cm 0 0 1.2cm},clip]{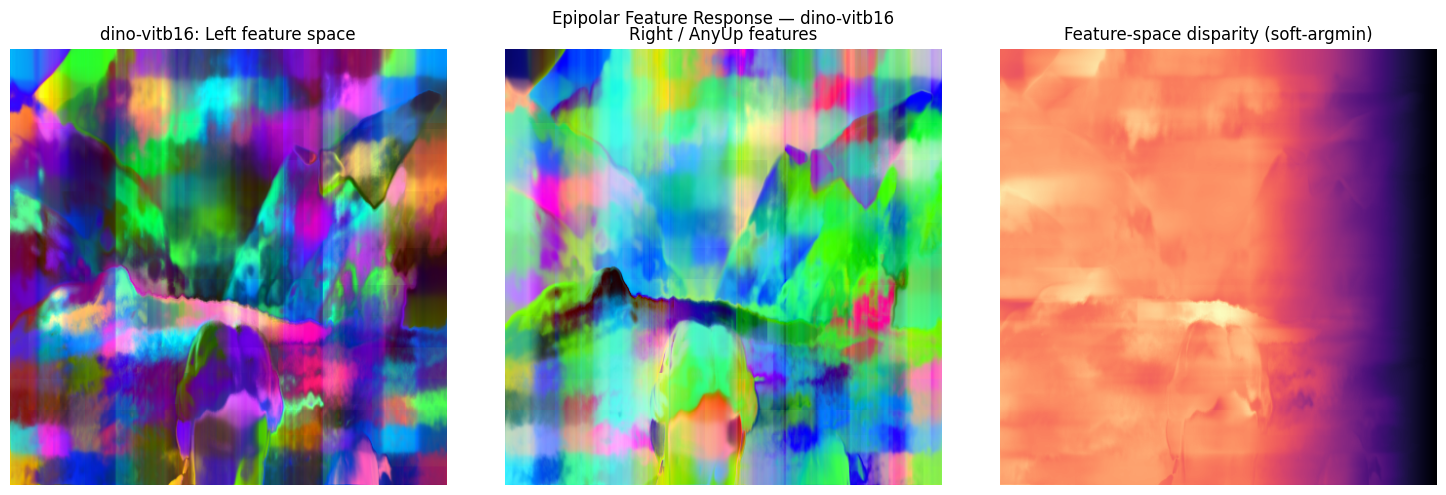}
         &
         \includegraphics[width=0.12\linewidth,trim={25.2cm 0 0 1.2cm},clip]{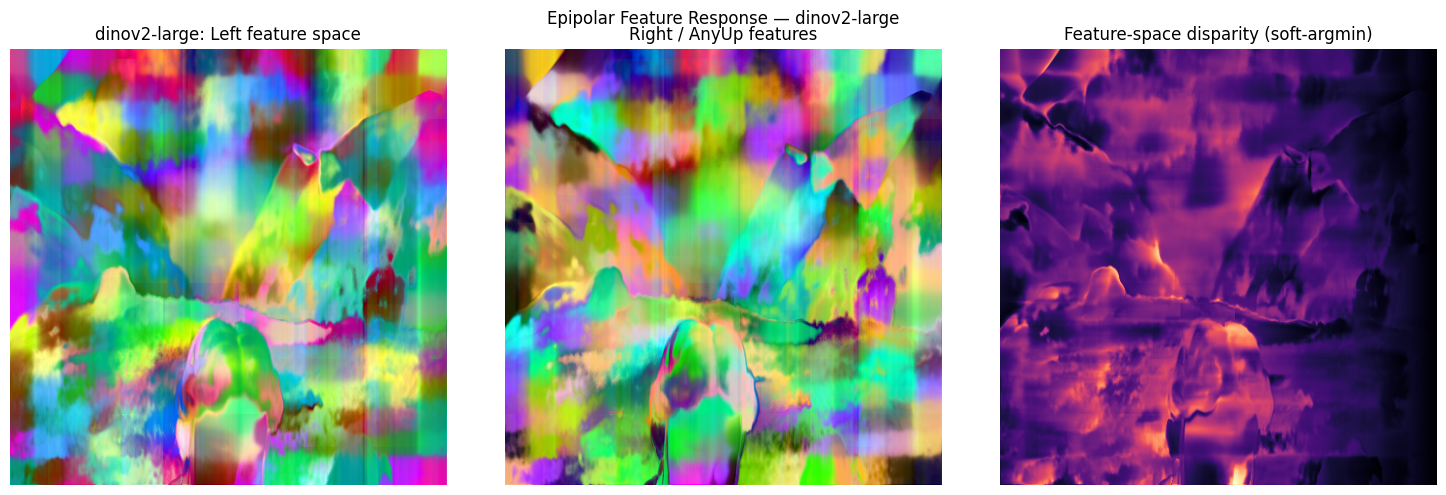} &
         \includegraphics[width=0.12\linewidth,trim={25.2cm 0 0 1.2cm},clip]{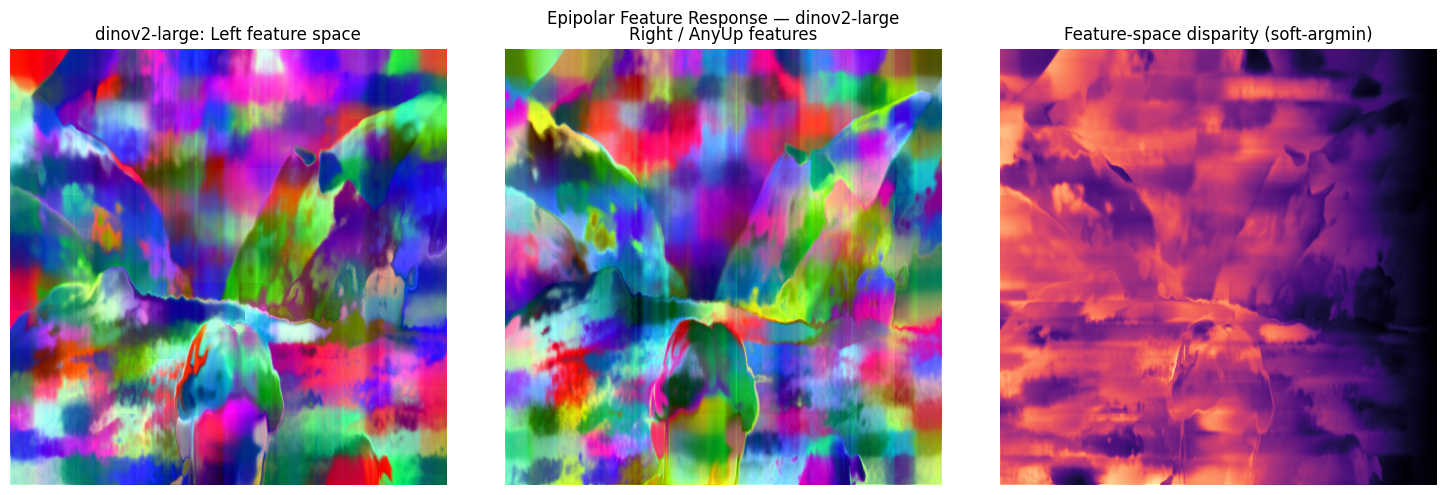} 
         &
         \includegraphics[width=0.12\linewidth,trim={25.2cm 0 0 1.2cm},clip]{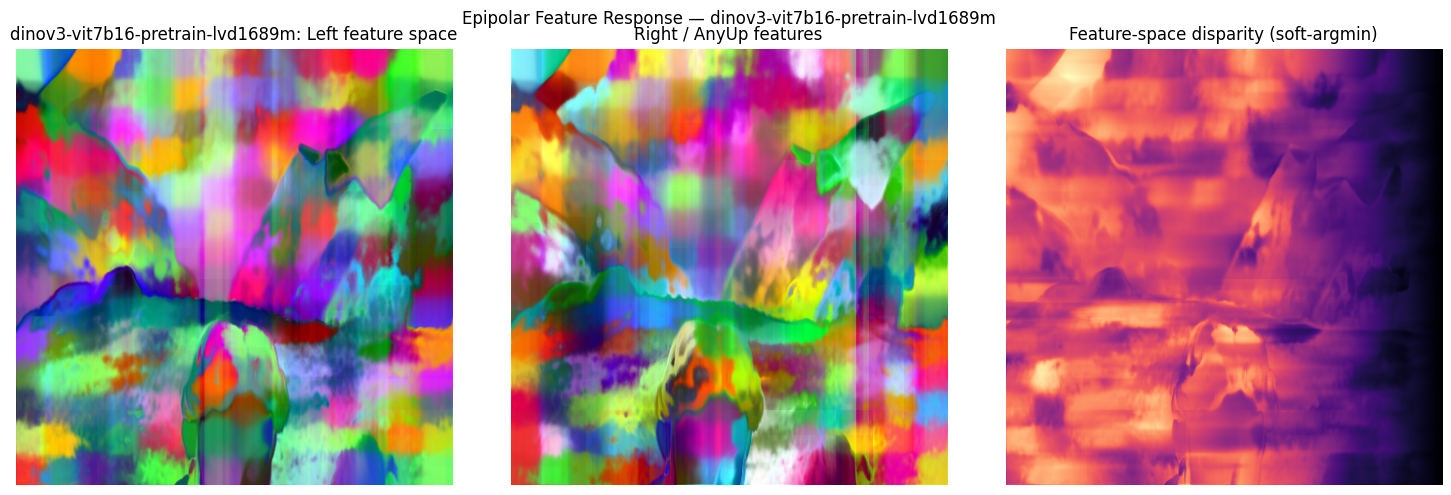} &
         \includegraphics[width=0.12\linewidth,trim={25.2cm 0 0 1.2cm},clip]{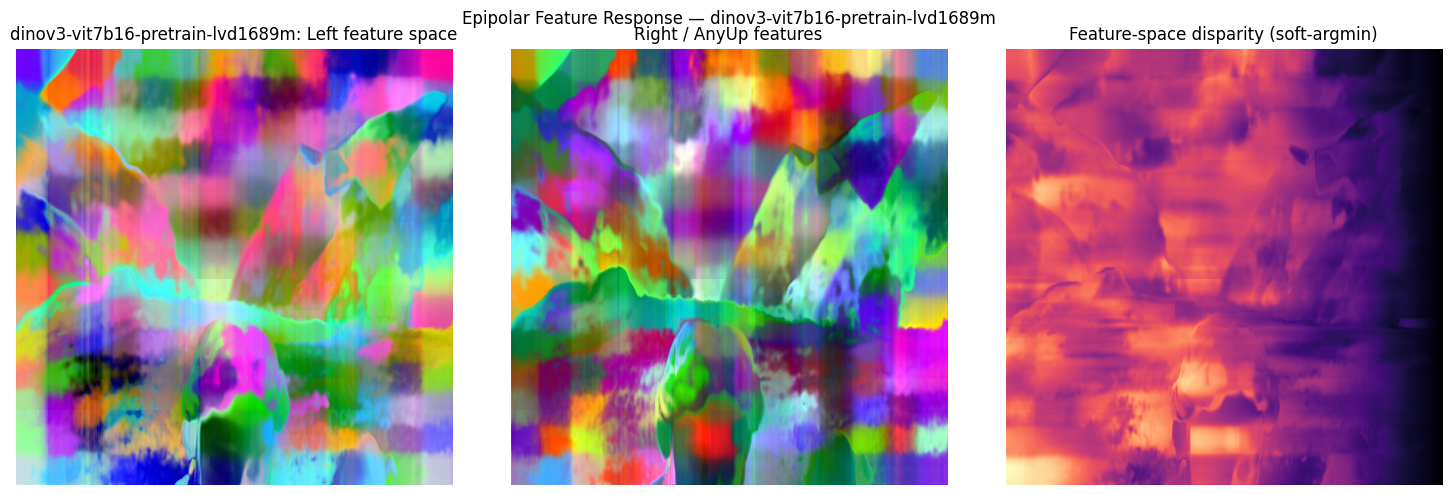}
         &
         \includegraphics[width=0.12\linewidth,trim={25.2cm 0 0 1.2cm},clip]{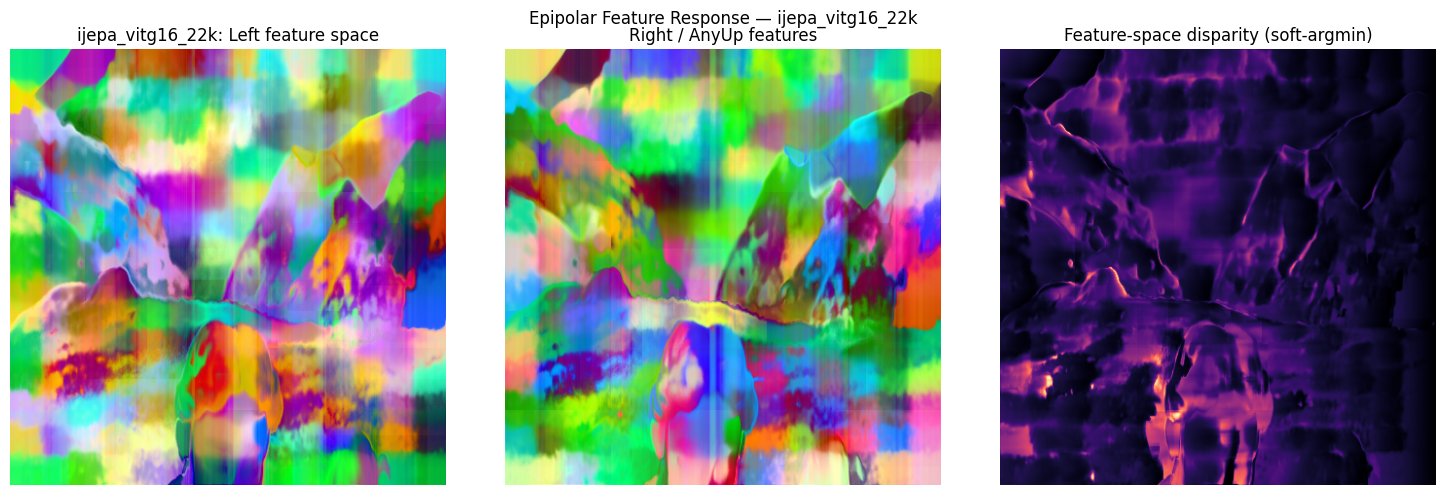} &
         \includegraphics[width=0.12\linewidth,trim={25.2cm 0 0 1.2cm},clip]{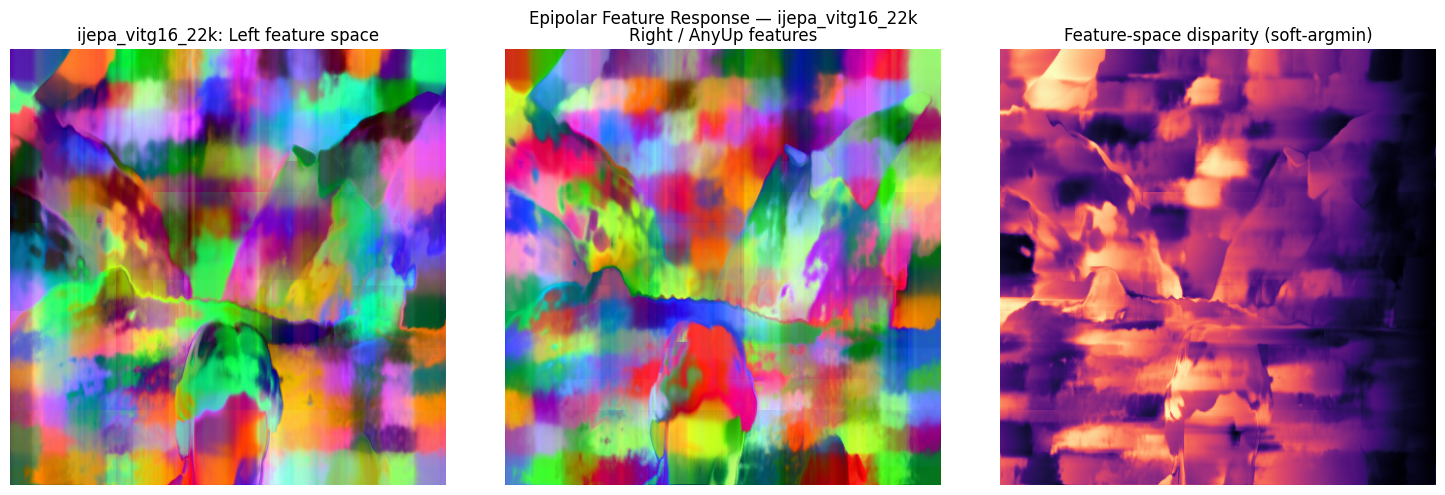} \\

         % {\textbf{MetaClip}} & {}
         % & {\textbf{DEiT}} & {}
         % & {\textbf{SigLip2}} & {} 
         % & {\textbf{BeiT}} & {} \\
        \multicolumn{2}{l}{\hspace{1em} \textbf{CLIP}}
        & \multicolumn{2}{l}{\hspace{1em} \textbf{DEiT}} 
        & \multicolumn{2}{l}{\hspace{1em} \textbf{SigLip2}} 
        & \multicolumn{2}{l}{\hspace{1em} \textbf{BeiT}}  \\
        \cmidrule(lr){1-1} \cmidrule(lr){3-3} \cmidrule(lr){5-5} \cmidrule(lr){7-7}

         {shuffled PE} & {pair shuffled PE}
         & {shuffled PE} & {pair shuffled PE}
         & {shuffled PE} & {pair shuffled PE} 
         & {shuffled PE} & {pair shuffled PE} \\
         \includegraphics[width=0.12\linewidth,trim={25.2cm 0 0 1.2cm},clip]{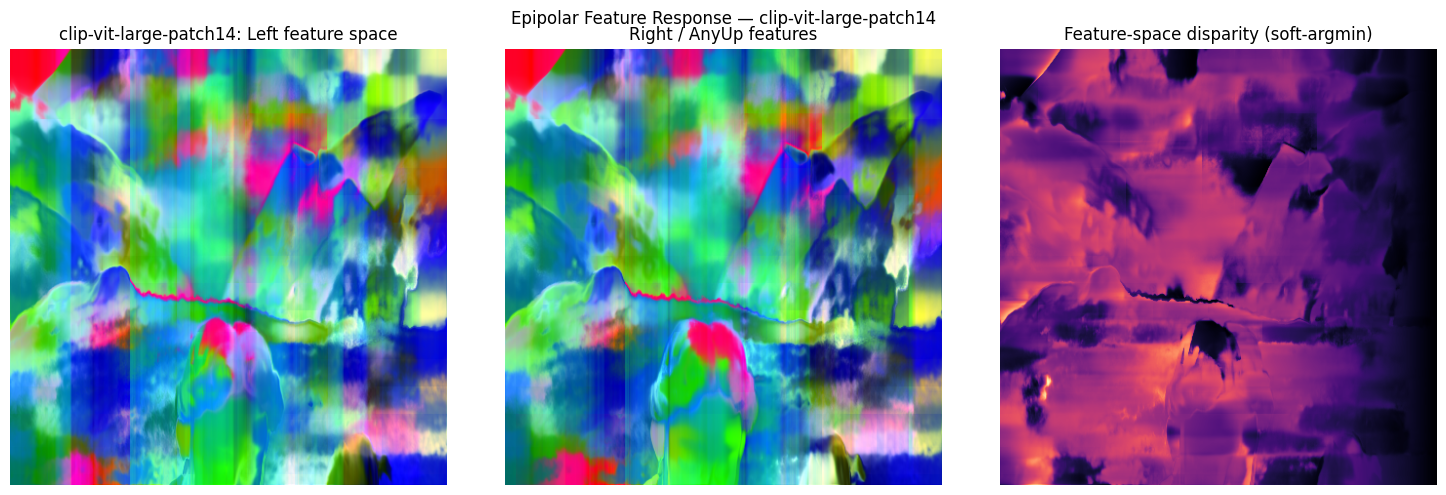} &
         \includegraphics[width=0.12\linewidth,trim={25.2cm 0 0 1.2cm},clip]{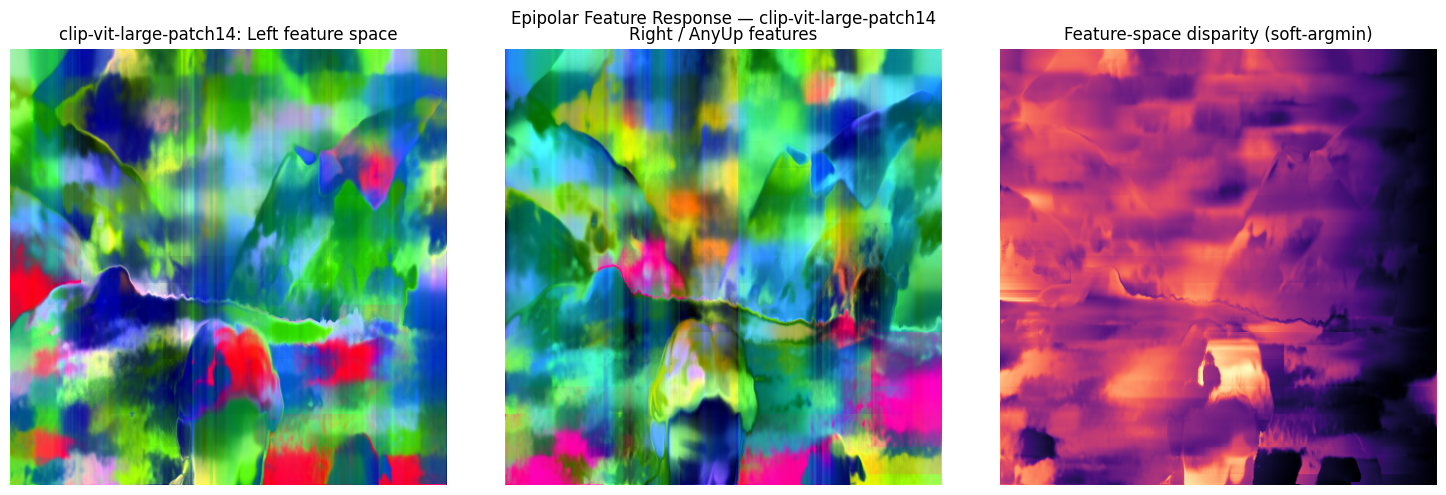}
         &
         \includegraphics[width=0.12\linewidth,trim={25.2cm 0 0 1.2cm},clip]{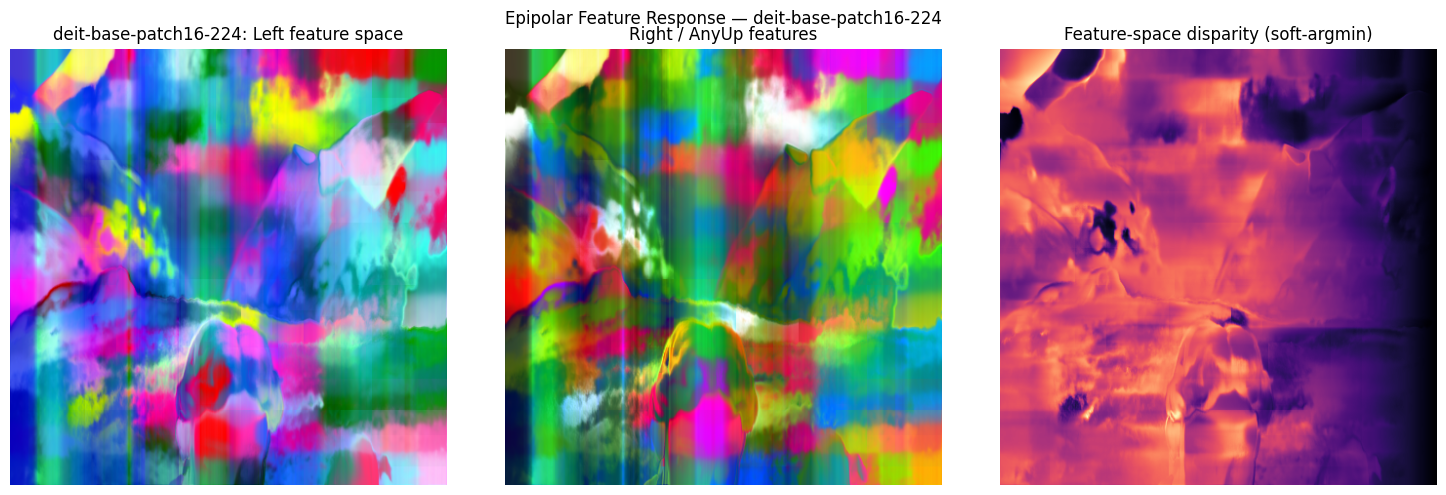} &
         \includegraphics[width=0.12\linewidth,trim={25.2cm 0 0 1.2cm},clip]{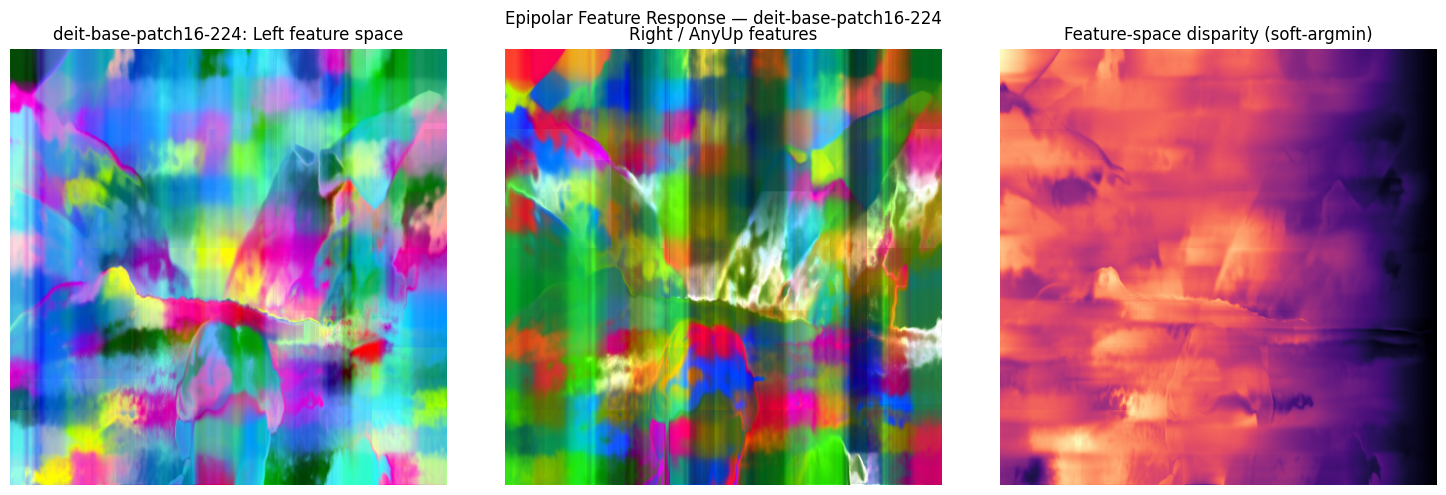} 
         &
         \includegraphics[width=0.12\linewidth,trim={25.2cm 0 0 1.2cm},clip]{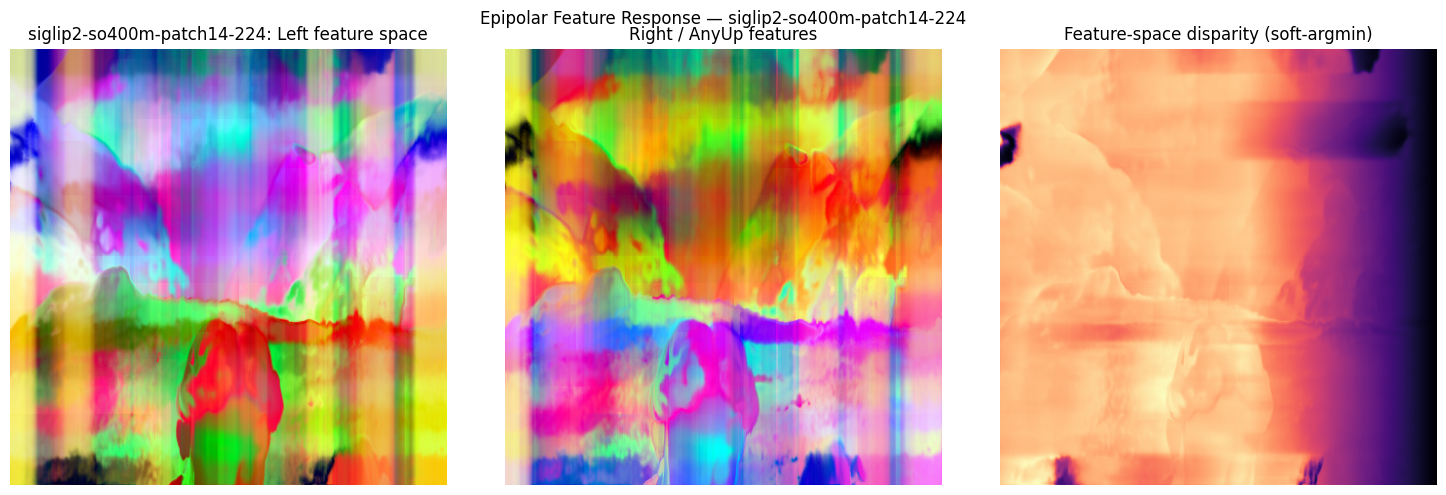} &
         \includegraphics[width=0.12\linewidth,trim={25.2cm 0 0 1.2cm},clip]{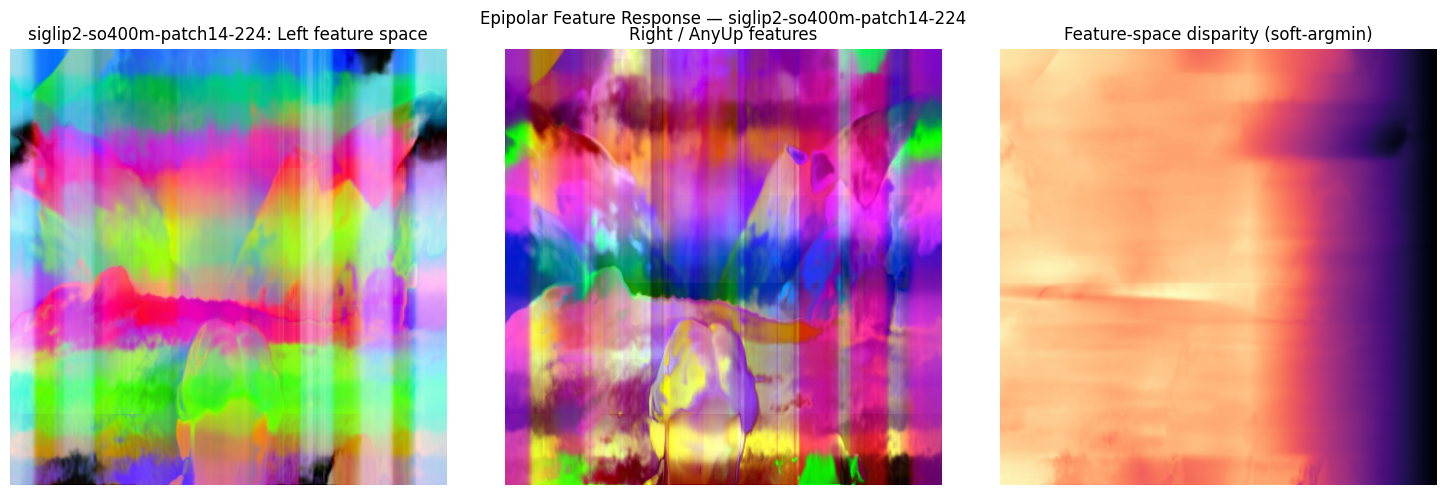}
         &
         \includegraphics[width=0.12\linewidth,trim={25.2cm 0 0 1.2cm},clip]{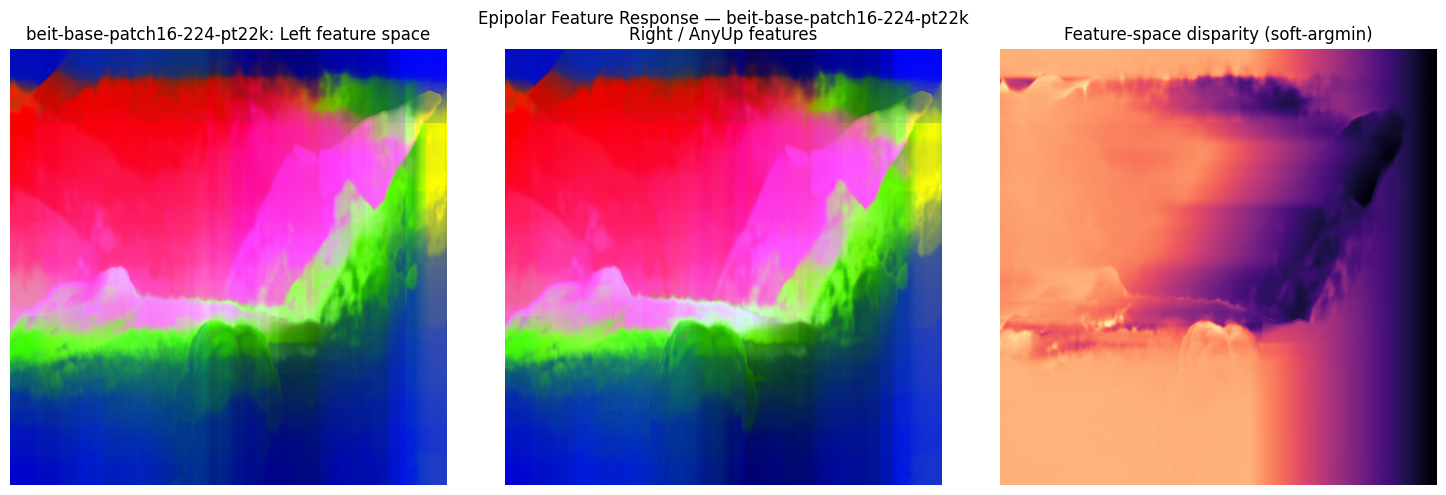} &
         \includegraphics[width=0.12\linewidth,trim={25.2cm 0 0 1.2cm},clip]{misc/epipolar_response_plots_no_reshuffle/epipolar_response_pos_anyup_pca_beit-base-patch16-224-pt22k.png} \\
         
        \multicolumn{2}{l}{\hspace{1em} \textbf{Swin}}
        & \multicolumn{2}{l}{\hspace{1em} \textbf{SwinV2}} 
        & \multicolumn{2}{l}{\hspace{1em} \textbf{MLCD}} 
        & \multicolumn{2}{l}{\hspace{1em} \textbf{Data2Vec}}  \\
        \cmidrule(lr){1-1} \cmidrule(lr){3-3} \cmidrule(lr){5-5} \cmidrule(lr){7-7}

         {shuffled PE} & {pair shuffled PE}
         & {shuffled PE} & {pair shuffled PE}
         & {shuffled PE} & {pair shuffled PE} 
         & {shuffled PE} & {pair shuffled PE} \\
         \includegraphics[width=0.12\linewidth,trim={25.2cm 0 0 1.2cm},clip]{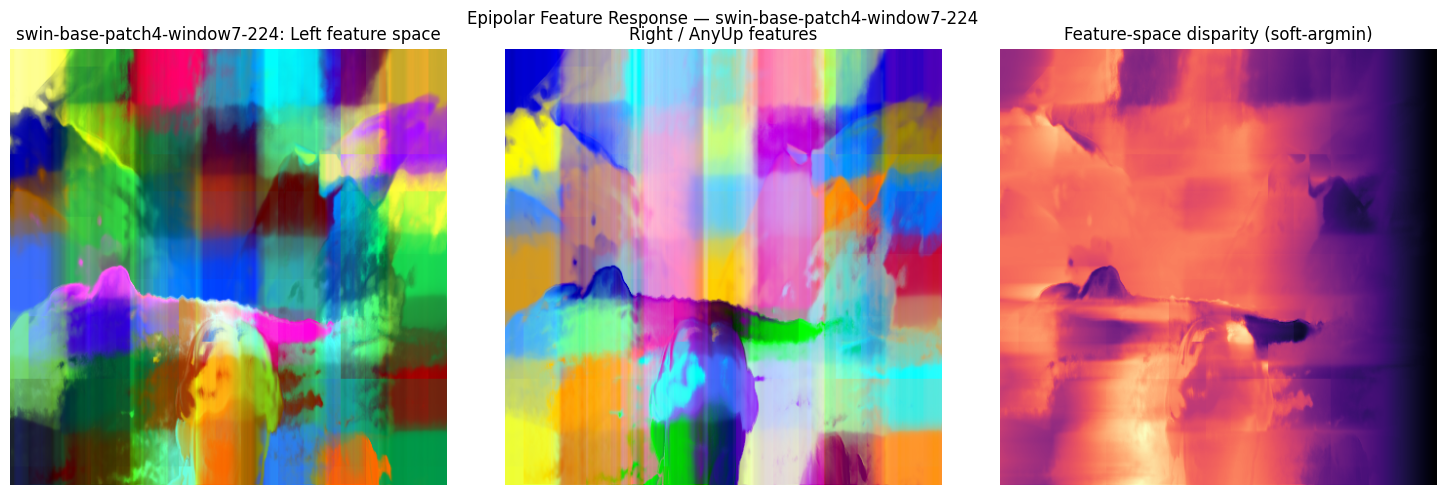} &
         \includegraphics[width=0.12\linewidth,trim={25.2cm 0 0 1.2cm},clip]{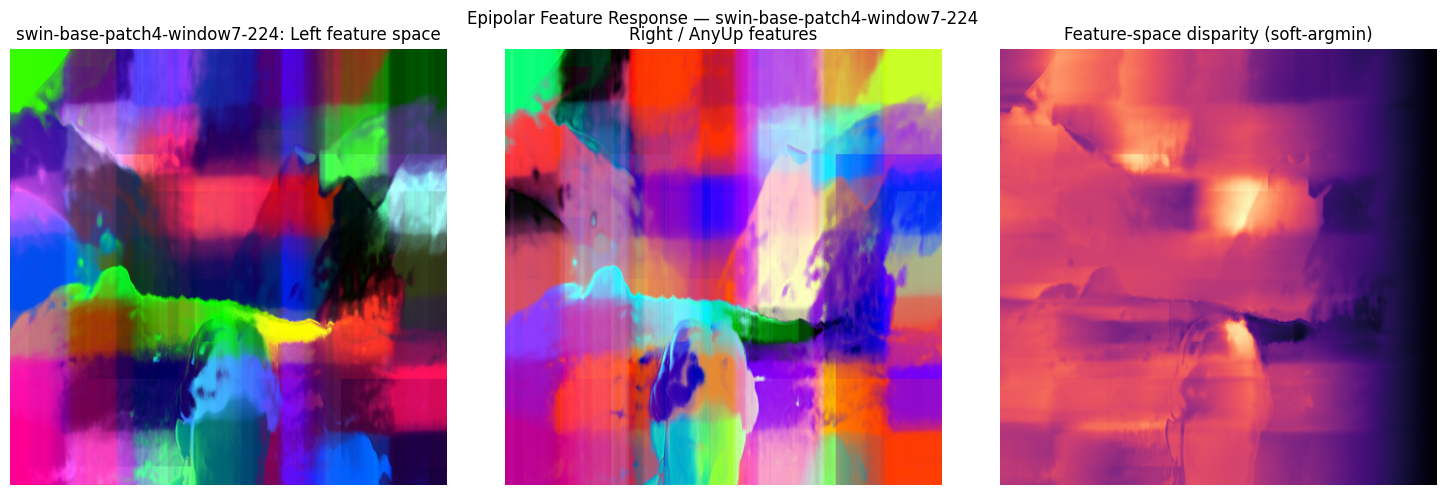}
         &
         \includegraphics[width=0.12\linewidth,trim={25.2cm 0 0 1.2cm},clip]{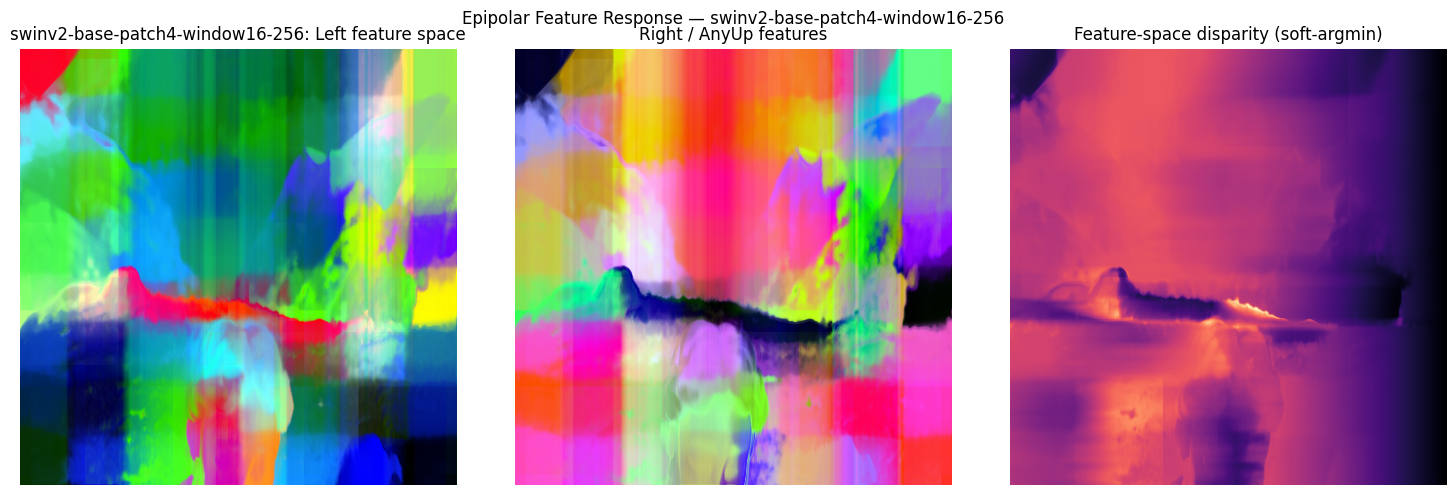} &
         \includegraphics[width=0.12\linewidth,trim={25.2cm 0 0 1.2cm},clip]{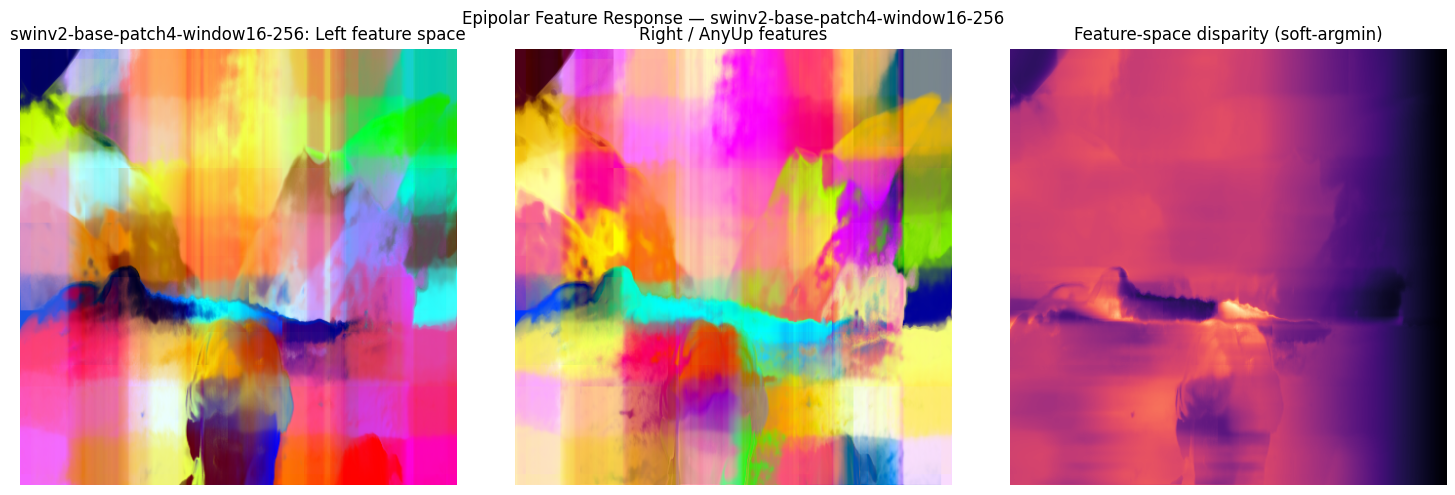} 
         &
         \includegraphics[width=0.12\linewidth,trim={25.2cm 0 0 1.2cm},clip]{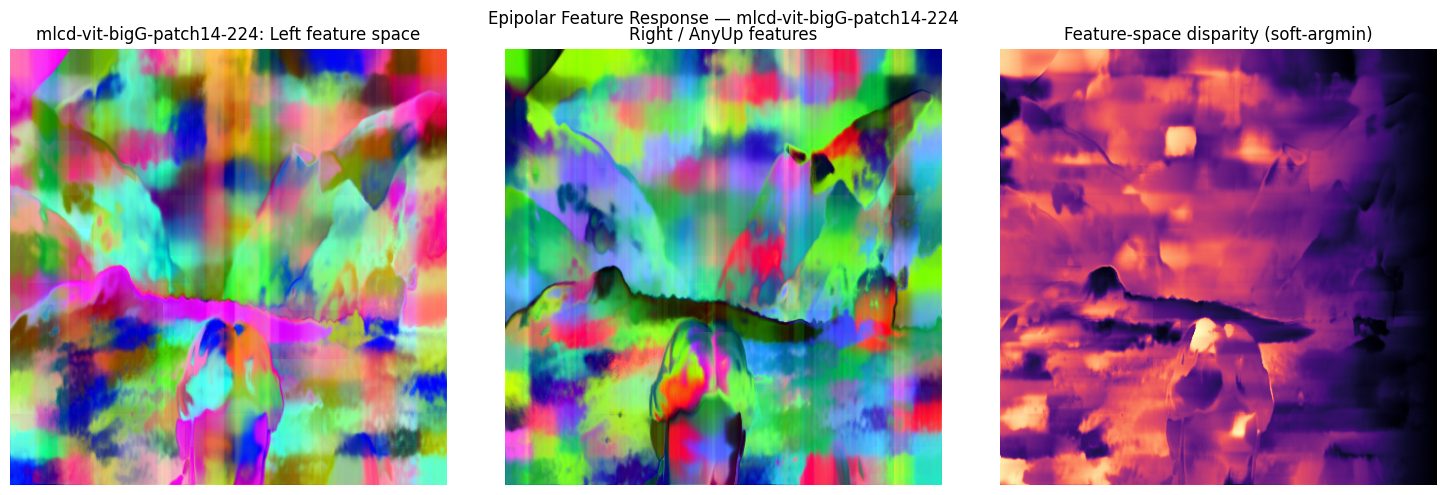} &
         \includegraphics[width=0.12\linewidth,trim={25.2cm 0 0 1.2cm},clip]{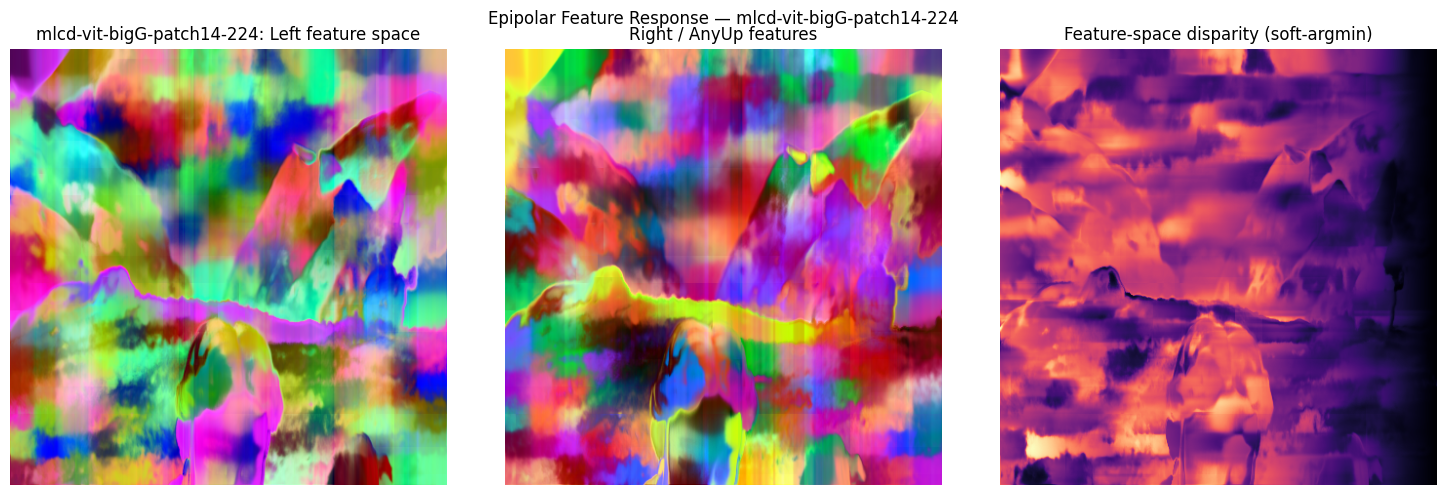}
         &
         \includegraphics[width=0.12\linewidth,trim={25.2cm 0 0 1.2cm},clip]{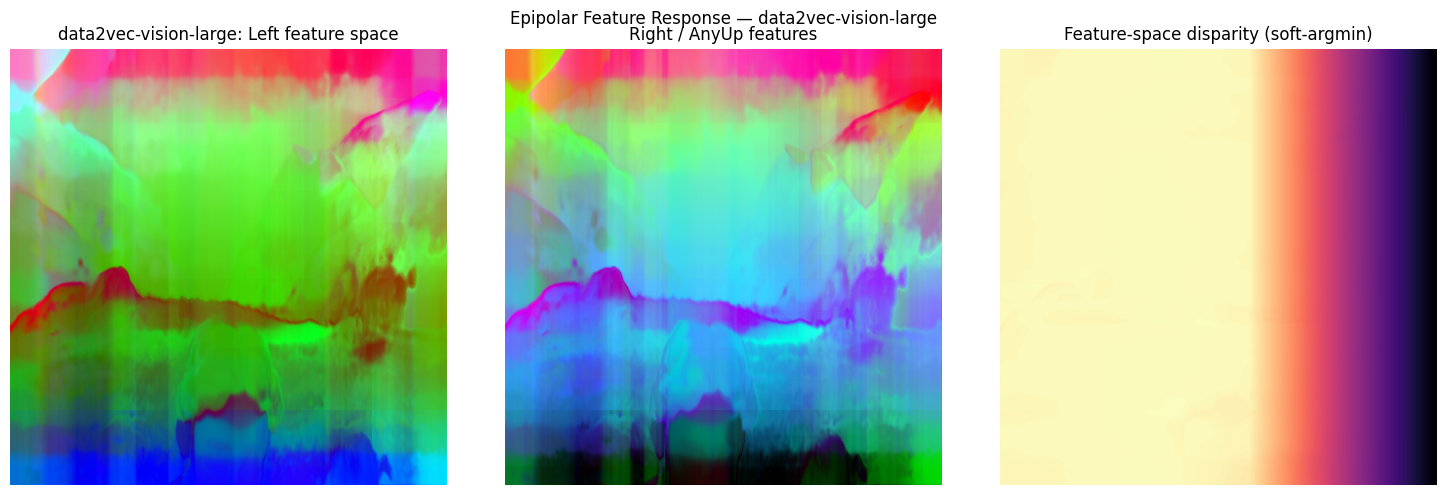} &
         \includegraphics[width=0.12\linewidth,trim={25.2cm 0 0 1.2cm},clip]{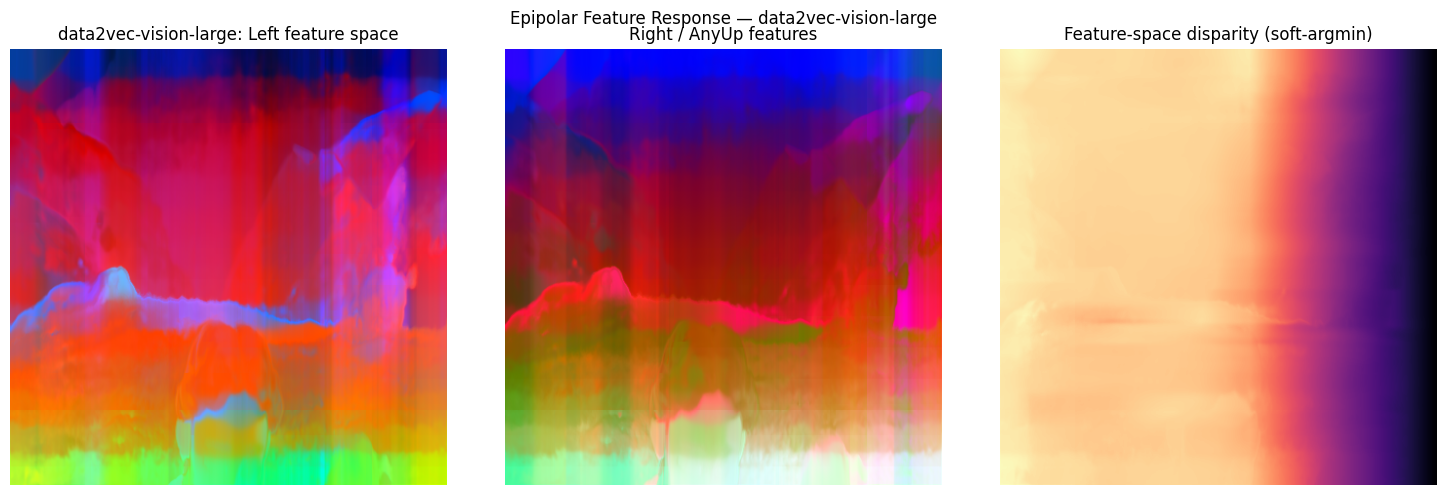}
         \\
    \end{tabular}
    \caption{\textbf{Epipolar peak response visualization for shuffled PEs.} 
    \textbf{Top-left}: an example stereo pair and the corresponding SAM feature maps with and without paired PE shuffling. Shuffled PEs result in collaposed/uninterpretable peak correspondence in general.
    }
    \label{fig:epipolar_plot_shuffled}
\end{figure*}

\textbf{Shuffled PEs.} In addition, we provide qualitative results of epipolar responses under random and pairwise PE shuffling in~\Cref{fig:epipolar_plot_shuffled}. Combining with the quantitative results in our main paper, though pairwise PE for absolute encoding schemes preserve geometry correspondence, the visualizations present in a chaotic manner that are not visually interpretable. 

\paragraph{The BEiT exception.}
BEiT stands out as an outlier, failing to exhibit any geometric correspondence even without any PE altering.
This aligns with prior observations that BEiT produces smooth, low-frequency representations with limited spatial specificity~\cite{bao2021beit,he2022masked,oquab2023dinov2,el2024probing}.
Interestingly, Data2Vec maintains stronger correspondence, despite employing the same positional encoding strategy as BEiT.
Since BEiT predicts discrete visual tokens via a dVAE, while Data2Vec predicts continuous latent representations, this suggests that discretization or quantization may suppress fine-grained geometry.
We emphasize, however, that disentangling the precise roles of discretization and quantization in shaping geometric fidelity remains an open question.

\paragraph{Residual Positional Cues.}
Interestingly, several architectures (\eg, Swin, MLCD) preserve a degree of spatial structure even when PEs are removed.
This residual locality is likely induced by the architectural design (\eg, local windows, multi-scale hierarchies) that implicitly encode spatial structure. 
Understanding these implicit sources of spatial structure and how they interact with or complement explicit positional encodings also poses an interesting direction for future architecture design.

\end{document}